%% file: main.tex
\documentclass[manuscript,screen,nonacm]{acmart}
\AtBeginDocument{%
  }

\copyrightyear{2026}
\acmYear{2026}

\usepackage{amsmath}
\usepackage[linesnumbered,ruled,vlined]{algorithm2e}
\usepackage{booktabs}
\usepackage{multirow}
\usepackage{placeins}
\usepackage{subcaption}
\usepackage{amsmath,amsthm,mathtools}
\usepackage{enumitem}
\input{macro}

\begin{document}

\title{One-Shot Clustering for Federated Learning Under Clustering-Agnostic Assumption}

\author{Maciej Krzysztof Zuziak}
\affiliation{%
  \institution{KDD Lab - National Research Council of Italy}
  \city{Pisa}
  \state{Tuscany}
  \country{Italy}}
\affiliation{%
  \institution{European University Institute}
  \city{Florence}
  \state{Tuscany}
  \country{Italy}}
\email{maciejkrzysztof.zuziak@isti.cnr.it}
\email{maciej.zuziak@eui.eu}

\author{Roberto Pellungrini}
\affiliation{%
  \institution{KDD Lab - Scuola Normale di Pisa}
 \city{Pisa}
 \state{Tuscany}
 \country{Italy}}
\email{roberto.pellungrini@sns.it}

\author{Salvatore Rinzivillo}
\affiliation{%
  \institution{KDD Lab - National Research Council of Italy}
 \city{Pisa}
 \state{Tuscany}
 \country{Italy}}
\email{rinzivillo@isti.cnr.it}

\renewcommand{\shortauthors}{Zuziak et al.}

\begin{abstract}
    \fl is a widespread and well-adopted paradigm of decentralised learning that allows training one model from multiple sources without the need to transfer data between participating clients directly. Since its inception in 2016, it has been divided into numerous subfields that deal with application-specific issues, such as data heterogeneity or resource allocation. One such sub-field, \cfl, is dealing with the problem of clustering the population of clients into separate cohorts to deliver personalised models. Although a few remarkable works have been published in this domain, the problem remains largely unexplored, as its basic assumptions and settings differ slightly from those of standard \fl. In this work, we present \ocfl,  a clustering-agnostic algorithm that can automatically detect the earliest suitable moment for clustering. Our algorithm is based on the computation of the cosine distance between gradients of the clients and a temperature measure that detects when the federated model starts to converge.
    We empirically evaluate our methodology by testing various one-shot clustering algorithms for over forty different tasks on five benchmark datasets. Our experiments showcase the good performance of our approach when used to perform \cfl in an automated manner without the need to adjust hyperparameters. We also revisit the practical feasibility of \cfl algorithms based on the gradients or weights of the clients, providing firm evidence of the high efficiency of density-based clustering methods when used to differentiate between the loss surfaces of neural networks trained on different distributions. Moreover, by inspecting the feasibility of local explanations generated with the help of GradCAM, we can provide more insights into the relationship between personalisation and the explainability of local predictions.
\end{abstract}

\ccsdesc[300]{Computing methodologies~Cluster analysis}
\ccsdesc[500]{Computing methodologies~Distributed algorithms}
\ccsdesc[500]{Computing methodologies~Neural networks}
\ccsdesc[500]{Computing methodologies~Distributed artificial intelligence}

\keywords{Federated Learning, Clustered Federated Learning, Client Clustering, Personalisation, Model Explainability, Cosine Distance, One-Shot Clustering}

\maketitle

\section{Introduction}
\label{sec:introduction}
The recent decade has witnessed exponential growth in the number of available data sources, primarily due to the expansion of \iot products and the popularisation of extensive data storing facilities that allow for the storage of unprecedented amounts of information. To leverage such an infrastructure, new paradigms of learning emerge, allowing data engineers and scientists to train models distributively without the need to transfer the data to a central location, which would often be unrealizable due to the sheer volume of stored information.

One such modern approach to decentralised \ml is \fl, a paradigm first introduced by \cite{b22} where the local models are trained locally, and the final (global) model is aggregated from the parameters of the local equivalents. Since the baseline \fl algorithm has been introduced under the name of \fedavg, the field has undergone recent developments with algorithms such as \fedopt (designed by \cite{b23}), SCAFFOLD (designed by \cite{b24}), and FedProx (designed by \cite{b40}) addressing the specific challenges that may arise in a federated environment.

The task of clustering clients into several conglomerations to train separate (personalised) models is one of such specific challenges that may arise, especially in the cross-silo setting, where the number of participants is limited to dozens rather than thousands and the institutional nature of the participants may require delivery of personalised solutions, at least at the granularity of the regional level. One example of such a situation is training a global risk model for insurance companies, where the significant differences in the local market are so vast that the final (global) model is inoperable for any of the interested parties. In the majority of such situations, companies may not know \textbf{(a)} that there exists some form of incongruence and \textbf{(b)} what grouping would be required to deliver models of satisfactory quality. The subfield of \cfl addresses such issues by applying clustering to the population of clients, delivering several personalised models that still retain some level of generalisability.

In this work, we approach the issue of fast and reliable \fl clustering that is automatically performed in the first few rounds of the training. Since many works in the field of \cfl employ clustering algorithms on the matrices of pairwise cosine distance (or cosine similarity), our main interest was focused on the generalisation of the presented procedure with an improvement of the clustering efficiency, especially clustering correctness. We, therefore, investigate whether it is possible to automatically detect the suitable clustering moment and perform efficient one-shot clustering at an early stage of the process without specifying any hyperparameters. Moreover, we turned our attention to the scenario of medium heterogeneity, where the clients differ only slightly in data-generating distributions.

We propose a clustering-agnostic algorithm that is free of any hyperparameters and clusters at the first sign of the (global) model's convergence. Our approach, called \ocfl, can be used to perform a reliable clustering in the first rounds of the training without the need to adjust either the number of clusters or the precise clustering criterion. As evidenced by the experimental section, when combined with density-based clustering (e.g., HDBSCAN, designed by \cite{b19}, or Mean-Shift, designed by \cite{b20}), it achieves better results than state-of-the-art \cfl algorithms in terms of clustering performance and personalisation, while retaining comparable levels of generalisability.

The paper makes four contributions:
\begin{enumerate}
    \item We introduce OCFL, a one-shot and clustering-algorithm-agnostic CFL
    procedure that automatically proposes an early clustering round without
    requiring knowledge of the number of DGPs (Data Generating Processes) or a dataset-specific trigger threshold.
    \item We establish a fixed-round recovery guarantee for the observed
    cosine-distance matrix under angular block separation. We then evaluate a
    ground-truth separation margin at the temperature-triggered round, clearly
    separating the theoretical certificate from the empirical trigger.
    \item We evaluate seven training/clustering configurations over five
    datasets, four split types, and two client populations, for 40 scenarios and
    280 runs. The study includes subtle overlapping and imbalanced splits rather
    than relying only on easily separable label swaps.
    \item We systematically analyse how cluster assignment changes local
    explanation quality through qualitative GradCAM maps, insertion/deletion
    AUC, and critical-difference analysis.
\end{enumerate}
All the experiments and their visual representation can be re-created using code deposited at the GitHub Repository shared with this article. The simulation code base is available at \url{https://github.com/MKZuziak/OCFLSuite}, and all numerical results together with the plotting code at \url{https://github.com/MKZuziak/OCFL_Archive}.
The remainder of the paper is organised as follows. Section~\ref{sec:RelatedWork} reviews related work; Section~\ref{sec:methodology} presents OCFL and its recovery analysis; Section~\ref{sec:experiments} reports the clustering, predictive, temperature, and margin experiments; Section~\ref{sec:xai_experiments} studies cluster-level explainability; Section~\ref{sec:considerations} states the scope and limitations; and Section~\ref{sec:conclusions} concludes. Complete scenario-level tables, auxiliary plots, proofs, and implementation details are provided in the appendix.

\section{Related Work}\label{sec:RelatedWork}
\subsection{Federated Learning} \label{RelatedWork:FL}
\fl was first introduced in the work of \cite{b22} as a decentralised optimisation method, where the training is distributed across several clients and each client shares only the parameters of the model trained on the local data, not disclosing its local (personal) dataset directly. The basic method of solving this problem, \fedavg (presented in \cite{b22}) assumes averaging values of the local parameters to arrive at the general global model. In the following years, different additional solutions to that problem were presented. \fedopt (introduced by \cite{b23}) generalises the \fedavg algorithm, allowing one to introduce adaptive optimisation into the process. Among others, SCAFFOLD (designed by \cite{b24}) and FedProx (designed by \cite{b40}) tackle the issue of data heterogeneity, limiting the gradient variability and thus stabilising the convergence process. As \fl became an important area of research, several surveys have tried to summarise the advancements made in the last decade of research, such as those presented by \cite{b25} or \cite{b26}. In this paper, we assume a horizontal learning scenario. In contrast to the vertical setting (surveyed, for example, by \cite{b27}), the horizontal one assumes that the sample space is disjoint, while the feature space is shared among all the participants. Implementing our clustering algorithms, we employ a \fedopt baseline. Let $P=\{1,\ldots,n\}$ be the client set, let $D_i$ contain $N_i$ examples,
and define the local empirical risk by $
F_i(\theta)
=
\frac{1}{N_i}
\sum_{(x,y)\in D_i}
\ell\!\left(h_\theta(x),y\right).
$ Horizontal FL seeks a shared parameter vector by solving $\theta^\star
\in
\operatorname*{arg\,min}_{\theta}
F(\theta),
\ 
F(\theta)
=
\sum_{i=1}^{n}
\frac{N_i}{\sum_{j=1}^{n}N_j}
F_i(\theta).$ FedOpt approximates this objective by alternating local optimisation with
server-side aggregation. In order to do that, we employ the \fedopt algorithm presented in \cite{b23}, which is a well-accepted baseline.

\subsection{Clustering in FL} \label{RelatedWork:FCL}
Clustering is a common personalisation approach in \fl: it is listed alongside model and data interpolation in \cite{b9} and placed under similarity-based methods in \cite{b53}. We view \cfl as split into two families: \textit{hypothesis-based clustering}, which groups clients by similarity of their hypothesis or objective (e.g. \cite{b3}, \cite{b9}), and \textit{parameter-based clustering}, which groups by similarity of local model parameters because those reflect the local data generating process (e.g. \cite{b1}, \cite{b7}, \cite{b2}, \cite{b4}, \cite{b10}, \cite{b8}). This paper focuses on the latter and proposes a natural extension of methods that compare local parameter sets.
Parameter-based methods typically differ in when clustering is performed and which algorithm is used. \cite{b1} uses hierarchical clustering on model parameters, fixing the clustering round and linkage hyperparameters a priori. \cite{b7} uses cosine similarity to bipartition clients and treats clustering as a postprocessing step after the global model converges. Other approaches use variants of distance or similarity such as EDC in \cite{b2}, combinations of distance and client-side adaptation in \cite{b5}, pairwise similarity to measure dataset heterogeneity in \cite{b10}, and cosine distance with client-side correction for dropouts in \cite{b8}.
Our method is closest to \cite{b7} and \cite{b1}, so we compare to them directly. By testing various splits, with emphasis on heterogeneity in class supports and label proportions, we extend their approaches to harder clustering scenarios and aim to identify the earliest moment for reliable one-shot clustering. Because we remain in the parameter-based paradigm, our method can augment techniques from \cite{b2}, \cite{b5}, \cite{b8}, and \cite{b10}.
We compare full network parameterisations rather than low-dimensional embeddings as in \cite{b2}, since our goal is to remain clustering algorithm agnostic. Experiments show that density-based methods such as HDBSCAN \cite{b19} or Mean-Shift \cite{b20} handle complex parameter spaces surprisingly well. Unlike \cite{b46} and \cite{b47}, we treat the problem as \textbf{hard clustering}, seeking a disjoint and well-defined cluster structure; these assumptions define our baseline comparisons to \cite{b7} and \cite{b1}.

\subsection{XAI and FL Intersection}
\label{RelatedWork:XAIFL}
There are many prior works on the intersection of \fl and explanation methods (explainable AI, XAI). The scoping review by \cite{b56} establishes three major forms of FL-XAI interplay: (a) interpretable models trainable in \fl; (b) explanation methods affected by \fl training; and (c) XAI techniques integrated into \fl schemas. The first category incorporates methods to train an explainer in federation, such as a federated training of fuzzy rule systems \cite{b57} or gradient-boosted trees \cite{b58}. The second category assesses the utility of the explanation methods in the \fl scenario \cite{b59, b60}. Finally, the third category integrates the explanation techniques into the existing federated algorithms (such as \fedopt) -- for example, the work of \cite{b61} uses explanations to assess the client relevance, while the work of \cite{b62} presents a privacy-preserving method for assessing the feature relevance.

Not much work exists on the intersection of personalisation in \fl and its impact on the quality of explanations. The work of \cite{b54} uses explanation methods to guide client selection in \cfl, similarly to the work of \cite{b61} deployed in the medical domain. Moreover, the work of \cite{b55} applies explainability techniques to \cfl in the domain of energy forecasting. Given the available research on the intersection of \fl and XAI, this paper takes a different research angle: we are mainly interested in the relationship between the personalisation as delivered by attributing certain clients to appropriate clusters and the quality of locally generated explanations, as based upon the examination of the saliency maps. To this end, we are also interested in more rigorous study of this relationship -- starting from the conventional qualitative analysis of the generated saliency maps, to the Critical-Difference analysis. Given the soaring interest of the \fl community in the explainability methods, we believe that a fundamental study on how the personalisation affects the locally generated explanation is a timely and important research avenue.

\section{Methodology}
\label{sec:methodology}
\subsection{Data Generating Processes} \label{methodology:data_process}

Simulating a divergence in data-generating processes that resembles the naturally occurring differences in data distributions is a challenge that is approached in various ways. \cite{b8} and \cite{b5} adopt label distribution and dataset distribution skew, with \cite{b8} additionally experimenting with differences in feature distribution. Both \cite{b7} and \cite{b1} employ label swaps to simulate incongruence in data distribution. \cite{b6} follows the methodology presented by \cite{b11} using the Dirichlet distribution to simulate a heterogeneous environment. Additionally, FEMNIST (employed by \cite{b1} and \cite{b10}) and Fed-Goodreads (employed by \cite{b10}) are used in some papers.

Let $(\Omega,\mathcal F,\mathbb P)$ be a probability space and let
$(\mathcal X,\mathcal A)$ be the measurable example space, where
$\mathcal X=\mathbb R^d\times\mathcal Y$ and $\mathcal Y$ is finite.  A
data-generating process (DGP) is a measurable map
$\zeta_c:\Omega\to\mathcal X$ with induced distribution

\begin{equation}
Q_c(A)
=
\mathbb P\!\left(\zeta_c^{-1}(A)\right),
\qquad
A\in\mathcal A.
\label{eq:patch-pushforward}
\end{equation}

Let $\mathcal C=\{c_1,\ldots,c_G\}$ denote the set of represented DGP
clusters.  An injective map
$\mathcal I:\mathcal C\to\mathfrak Z$ assigns a distinct DGP to each
cluster, and a surjective map $Z:P\to\mathcal C$ assigns every client to a
represented cluster.  Writing $Z(p_i)=c_{z_i}$, client $i$ receives a local
dataset sampled from $Q_{z_i}$.  The true client partition is therefore

\begin{equation}
\mathcal C^\star
=
\left\{
C_c^\star
\right\}_{c=1}^{G},
\qquad
C_c^\star
=
\left\{
i\in P:
z_i=c
\right\}.
\label{eq:patch-true-partition}
\end{equation}

Surjectivity ensures that every represented DGP has at least one client.

In practice, one would preferably opt for a standard benchmarking dataset as a point of validation for the tested algorithms. To simulate various levels of divergence, while still being consistent with the methodology presented above, we define two different split scenarios: \textbf{(i) overlapping} and \textbf{(ii) non-overlapping} splits. \textbf{Non-overlapping} simulates a behaviour encountered in data silos when the set of classes is completely disjoint from the set of classes available in another silo. \textbf{Overlapping} is more challenging clustering-wise, as it implies that there is some overlap between those two sets. It may be caused, for example, by the fact that both distributions are part of the same parametric family, but the respective parameters are not identical.

For each DGP cluster $c$, let $\mathcal Y_c\subseteq\mathcal Y$ be its available label set.  A split is \emph{non-overlapping} when $\mathcal Y_c\cap\mathcal Y_d = \varnothing \
\forall c\neq d,$ and is \emph{overlapping} when $\mathcal Y_c\cap\mathcal Y_d \neq \varnothing \ \exists \ c\neq d.$
For balanced splits, labels are sampled uniformly from $\mathcal Y_c$.
For imbalanced splits, each cluster draws $v_c\sim\operatorname{Dirichlet}(\alpha\mathbf 1)$ and then samples $Y\sim\operatorname{Categorical}(v_c)$. Conditional features are drawn from the dataset distribution $Q_c(X\mid Y)$. In all splits this class-conditional feature distribution is that of the underlying dataset and is shared across clusters, so the clusters differ only in their label supports and label proportions.  We consider all four combinations of overlapping/non-overlapping and balanced/imbalanced splits.

Compared to \cite{b25}, our methodology covers \textit{\textbf{label distribution skew}} ($\varphi_i[y]$ varies across clients, with $\varphi_i[\boldsymbol{x}|y] = \varphi_j[\boldsymbol{x}|y] \quad \forall i, j \in P$) and non-independent samples. We omit \textit{feature distribution skew}, \textit{quality skew}, \textit{\textbf{concept drift}} and \textit{\textbf{concept shift}}. Subtle heterogeneity in class supports and label proportions is harder to detect than extreme heterogeneity, which heavily impacts learning but is easily detected \cite{b1, b3, b7}.

\subsection{Hypothesis Function Similarity} \label{methodology:function_similarity}

Accessing the information about the local distribution is impossible, as it would nullify the basic assumption underlying the \fl. However, some information about the status of the learning can be derived from the model's parameters that reflect a local hypothesis function. For each cluster $C_i \in C$, there exists a hypothesis function $h_{C_i}$ that minimises the empirical risk function on a dataset sampled from the corresponding data-generating distribution, i.e. $h_{C_i}^* = \operatorname*{arg\,min}_{h \in \mathcal{H}} R_{ERF}(h, D_i) $, where $D_i \sim \varphi_i$. Every client belonging to a certain cluster is training its own hypothesis function $h_{p_i}$ that minimises the local risk function. The hypothesis function $h_{C_i}^*$ is not guaranteed to outperform the local hypothesis function $h_{p_i}$ in every case since it depends on the data stored on the local client. However, given the environment, $h_{C_i}^*$ will be the function that minimises the empirical risk for the whole cluster. One illustrative example may be connected to a small sample size of the local dataset. If the client is in possession of only a few samples, an overfitted local hypothesis function may outperform the best cluster hypothesis function. However, as the client will procure new samples, the best cluster hypothesis function will perform better than the overfitted local function. Those notions are formalised in the Definition \ref{def:Cluster_Hypothesis_Function}. 
\begin{definition}[Cluster hypothesis function]
\label{def:Cluster_Hypothesis_Function}
For cluster $c$, let $C_c^\star=\{i:Z(p_i)=c\}$. A cluster-optimal
parameter vector is any

\begin{equation}
\theta_c^\star
\in
\operatorname*{arg\,min}_{\theta}
\sum_{i\in C_c^\star}
\omega_{i\mid c}F_i(\theta),
\qquad
\omega_{i\mid c}\geq 0,
\qquad
\sum_{i\in C_c^\star}\omega_{i\mid c}=1.
\label{eq:patch-cluster-risk}
\end{equation}
The corresponding cluster hypothesis is $h_{\theta_c^\star}$.  We use
$\omega_{i\mid c}=N_i/\sum_{j\in C_c^\star}N_j$ when weighting clients by
their local sample sizes.
\end{definition}

The similarities of local loss functions are used by \cite{b3} or \cite{b9} to cluster the population of clients into a predefined number of clusters. This approach suffers from a high computational cost (as each client must test out each possible hypothesis function to find the one that minimises the local risk) and the necessity to know \textit{a priori} the structure of the population, \ie the number of total clusters. However, one could note that the hypothesis function $h_i$ is parameterised by the n-dimensional tensor $\boldsymbol{\Theta}$.\footnote{The dimensions of the tensor will depend on the used model. In the case of conventional $n$ layered feedforward neural networks, the weights are often represented in $n-1$ separate matrices of size $a \times b$, where $b$ is the number of neurons in the layer $j$ and $a$ is the corresponding number of neurons in the $j + 1$ layer. This can be represented as a three-dimensional tensor $\Theta \in \mathbb{R}^{(n - 1) \times a \times b}$. Without the loss of generality, we assume that $\Theta \in \mathbb{R}^n$ or $\Theta \in \mathbb{R}^n \times \mathbb{R}^m$ where $n, m \in \mathbb{Z}$.} The flattened one-dimensional version of this tensor, i.e. $\boldsymbol{\hat{\Theta}} = Vec(\Theta)$, can be compared with other parameterised hypothesis functions without the need to directly assess the hypothesis function performance on the local dataset - the method used in \cite{b1}, \cite{b2}, \cite{b7} or \cite{b8}. 

The current state of the art does not provide answers to when the clustering could be initialised at the earliest. To develop an intuitive understanding of the problem, let $C_s(\boldsymbol{\hat{\Theta}}_i,\boldsymbol{\hat{\Theta}}_j)$ denote \textbf{cosine similarity} between vectors of parameters $\hat{\Theta}_i$ and $\hat{\Theta}_j$ and $C_d(\boldsymbol{\hat{\Theta}_i}, \boldsymbol{\hat{\Theta}_j}) = 1 - C_s(\boldsymbol{\hat{\Theta}_i},\boldsymbol{\hat{\Theta}_j})$ denote a \textbf{cosine distance}. Those are placed in Definitions \ref{def:Cosine_Similarity} and \ref{def:Cosine_Distance} for reference.

\begin{definition}[Cosine Similarity]\label{def:Cosine_Similarity}
    \begin{equation}
        C_s(\boldsymbol{\hat{\Theta}}_i,\boldsymbol{\hat{\Theta}}_j) = \frac{\boldsymbol{\hat{\Theta}_i}^{\top} \boldsymbol{\hat{\Theta}_j}}{||\boldsymbol{\hat{\Theta}_i}|| ||\boldsymbol{\hat{\Theta}_j}||} = \frac{\sum_k^n \hat{\theta}_{k(i)}\hat{\theta}_{k(j)}}{\sqrt{\sum_k^n \hat{\theta}^2_{k(i)}} \sqrt{\sum_k^n \hat{\theta}^2_{k(j)}}}
    \end{equation}
    and $C_s(\boldsymbol{\hat{\Theta}}_i,\boldsymbol{\hat{\Theta}}_j) \in [-1, 1]$.
\end{definition}

\begin{definition}[Cosine Distance]\label{def:Cosine_Distance}
    \begin{equation}
        C_d(\boldsymbol{\hat{\Theta}_i}, \boldsymbol{\hat{\Theta}_j}) = 1 - C_s(\boldsymbol{\hat{\Theta}_i},\boldsymbol{\hat{\Theta}_j}) = 1 - \frac{\boldsymbol{\hat{\Theta}_i}^{\top} \boldsymbol{\hat{\Theta}_j}}{||\boldsymbol{\hat{\Theta}_i}|| ||\boldsymbol{\hat{\Theta}_j}||}
    \end{equation}
    and $C_d(\boldsymbol{\hat{\Theta}}_i,\boldsymbol{\hat{\Theta}}_j) \in [0, 2]$.
\end{definition}

Subsequently, define $\boldsymbol{\Gamma} \in \mathbb{R}^{n \times n}$ symmetric matrix that captures the similarity between the parametrisation of a hypothesis function provided by all the sampled clients. The $(i,j)$ entry of this matrix will be defined as $\gamma_{(i, j)} = \gamma_{(j, i)} = C_d(\boldsymbol{\hat{\Theta}}_i,\boldsymbol{\hat{\Theta}_j})$ and the whole matrix will be called Divergence Matrix as in Definition \ref{def:Divergence_Matrix}.

\begin{definition}[Divergence Matrix]
\label{def:Divergence_Matrix}
    Let $\Theta = \{\hat{\Theta}_i\}_{i=0}^n$ denote the set of individual parametrisations dispatched by clients. Then, the \textbf{Divergence Matrix} $\Gamma \in \mathbb{R}^{n \times n}$ is a symmetric matrix s.t.:
    \begin{equation}
        \Gamma_{(i, j)} = \Gamma_{(j, i)} = C_d(\boldsymbol{\hat{\Theta}}_i,\boldsymbol{\hat{\Theta}_j}) = 1 - C_s(\boldsymbol{\hat{\Theta}_i},\boldsymbol{\hat{\Theta}_j}) = 1 - \frac{\boldsymbol{\hat{\Theta}_i}^{\top} \boldsymbol{\hat{\Theta}_j}}{||\boldsymbol{\hat{\Theta}_i}|| ||\boldsymbol{\hat{\Theta}_j}||}
    \end{equation}
\end{definition}

Definition~\ref{def:Divergence_Matrix} gives the generic cosine-distance construction for
nonzero client-dispatched parameter vectors.  OCFL instantiates the
dispatched vector with the local parameter displacement

\begin{equation}
\widehat{\Theta}_i^{\,t}
:=
\Delta_i^t
=
\theta_i^{(t,K)}
-
\theta_{C(i)}^t,
\qquad
\boldsymbol{\Gamma}_t(i,j)
=
C_d\!\left(\Delta_i^t,\Delta_j^t\right),
\label{eq:patch-divergence-instantiation}
\end{equation}

where $\theta_{C(i)}^t$ is the round-$t$ model of the cluster containing
client $i$; before the one-shot split, it is the common population model. Cosine distance is evaluated only for nonzero inputs.

Matrix $\boldsymbol{\Gamma}$ captures the learning divergence between the sampled nodes. To obtain a single-valued description of the current convergence progress, we define the \textbf{Clustering Temperature Function} in Definition \ref{def:Temperature_Function}.

\begin{definition}[Temperature Function with Norm $p$]\label{def:Temperature_Function}
    Let $\boldsymbol{\Gamma} \in \mathbb{R}^{n \times n}$ be a Divergence Matrix (Definition \ref{def:Divergence_Matrix}). Then, the \textbf{Clustering Temperature Function with Norm $p$} is defined as:
    \begin{equation}
        T(\boldsymbol{\Gamma}) = \frac{||\boldsymbol{\Gamma}||_p}{\lambda}
    \end{equation}
    where additionally $\lambda > 0$ and $p$ is a chosen norm s.t.:
    \begin{equation}
        ||\boldsymbol{\Gamma}||_p = ||vec(\boldsymbol{\Gamma})||_p = (\sum_{i=1}^n \sum_{j=1}^n |\gamma_{i,j}|^p)^{\frac{1}{p}}
    \end{equation}
\end{definition}

We use the entry-wise matrix $p$-norm:

\begin{equation}
\lVert\boldsymbol{\Gamma}\rVert_p
=
\left(
\sum_{i=1}^{n}
\sum_{j=1}^{n}
\left|
\Gamma(i,j)
\right|^p
\right)^{1/p}.
\label{eq:patch-entrywise-norm}
\end{equation}

Because $\Gamma(i,i)=0$ and every cosine distance lies in $[0,2]$, its
largest possible value is $2[n(n-1)]^{1/p}$. We therefore set

\begin{equation}
\lambda_p
=
2[n(n-1)]^{1/p},
\qquad
T(\boldsymbol{\Gamma})
=
\frac{\lVert\boldsymbol{\Gamma}\rVert_p}{\lambda_p},
\qquad
0\leq T(\boldsymbol{\Gamma})\leq 1
\label{eq:patch-temperature-normalisation}
\end{equation}

All reported experiments use $p=2$, so
$\lVert\boldsymbol{\Gamma}\rVert_2$ is the Frobenius norm.  Temperature is
an aggregate measure of pairwise update divergence; it is not, by
definition alone, a convergence measure.

\subsection{Clustering Procedure}\label{methodology:clustering_procedure}
The proposed algorithm is based on the empirical observations of the \textbf{Temperature Function} from Definition \ref{def:Temperature_Function}. The preliminary hypothesis assumed that the function $T$ may express two different types of behaviours: \textbf{(i)} either the value may decrease at the beginning of the training (initial convergence) and then rise after reaching a local minimum due to incongruence of the data splits, or \textbf{(ii)} it may rise from the very beginning without reaching a local minimum at all due to aforementioned incongruence of the data splits. Those preliminary assumptions were verified with the detailed and repeated observation of the \textbf{Temperature Function} behaviour over multiple datasets and multiple data splits (this is evidenced and discussed in the Subsection \ref{experiment: temperature_function}).

Using this observation, we present a \textit{\textbf{One-Shot Clustered Federated Learning (OCFL)}} in Algorithm \ref{algo:ocfl}. Following this procedure, the first initial cluster is equal to the whole population.\footnote{The presented algorithm assumes sampling the whole population each round. However, a dynamic environment (where clients appear and disappear on an ongoing basis) can be taken into consideration in several different ways. This is explored in additional research directions at the end of this paper.} The first $T_{-1}$ value is set to $\infty$ (line 2). As mentioned earlier, two types of behaviours are expected: we either \textbf{(i)} observe a descent and wait for the first sign of a rise or \textbf{(ii)} await the rise of the gradients from the very beginning of the training. Since we are considering one-shot clustering, the flag $\mathcal{F}$ is initially set to false (line 2).  Lines 3-10 describe the beginning of the standard \fl process, with computation of local stochastic gradient (line 8), performing local update (line 9) and computing local model delta (line 10), all done in parallel. After the gradients are aggregated, the next step is to check whether the clustering was already performed (line 11).  If not, the matrix $\Gamma$ capturing the cosine distance between the gradients is reconstructed, and the \textbf{Clustering Temperature} (Definition \ref{def:Temperature_Function}) is calculated (line 12-13). If the newly registered temperature exceeds the one previously registered (line 14), the algorithm performs an actual clustering and switches the flag $\mathcal{F}$ to $True$. Otherwise, it simply returns the initial clustering structure. The in-cluster models are reconstructed from gradients of clients attributed to a specific cluster (lines 19-21). Please note that for clarity, a symbol of $\boldsymbol{\Theta}$ was replaced with a symbol $\theta$ in the description of the algorithm.

\begin{algorithm}[t]
\caption{One-Shot Clustered Federated Learning (OCFL)}
\label{algo:ocfl}
\KwIn{client set $P$; initial model $\theta^0$; local steps $K$;
client and server optimisers; learning rates $\eta_c,\eta_s$;
clustering algorithm $\mathcal A$; number of rounds $R$}
$\mathcal C\leftarrow\{P\}$;
$\theta_P^0\leftarrow\theta^0$\;
$\mathcal F\leftarrow\mathrm{False}$;
$T_{-1}\leftarrow+\infty$\;
\For{$t=0,\ldots,R-1$}{
  \ForEach{$C\in\mathcal C$}{
    \ForEach{client $i\in C$ \textbf{in parallel}}{
      $\theta_i^{(t,0)}\leftarrow\theta_C^t$\;
      \For{$k=0,\ldots,K-1$}{
        $g_i(\theta_i^{(t,k)})\leftarrow
        \nabla_\theta\mathcal L(
        h(D_i,\theta_i^{(t,k)}),y)$\;
        $\theta_i^{(t,k+1)}\leftarrow
        \mathrm{ClientOpt}(
        \theta_i^{(t,k)},
        g_i(\theta_i^{(t,k)}),\eta_c)$\;
      }
      $\Delta_i^t\leftarrow
      \theta_i^{(t,K)}-\theta_C^t$\;
    }
  }
  \If{$\mathcal F=\mathrm{False}$}{
    $\boldsymbol{\Gamma}_t(i,j)\leftarrow
    C_d(\Delta_i^t,\Delta_j^t)$
    for all $i,j\in P$\;
    $T_t\leftarrow
    \lVert\boldsymbol{\Gamma}_t\rVert_p/\lambda_p$\;
    \If{$T_t>T_{t-1}$\nllabel{l:trigger}}{
      $\mathcal C_{\mathrm{new}}\leftarrow
      \mathcal A(\boldsymbol{\Gamma}_t)$\;
      $\theta_C^t\leftarrow\theta_P^t$
      for every $C\in\mathcal C_{\mathrm{new}}$\;
      $\mathcal C\leftarrow\mathcal C_{\mathrm{new}}$\;
      $\mathcal F\leftarrow\mathrm{True}$\;
    }
  }
  \ForEach{$C\in\mathcal C$}{
    $\overline{\Delta}_C^t\leftarrow
    |C|^{-1}\sum_{i\in C}\Delta_i^t$\;
    $\theta_C^{t+1}\leftarrow
    \mathrm{ServerOpt}(
    \theta_C^t,\overline{\Delta}_C^t,\eta_s)$\;
  }
}
\end{algorithm}

\subsection{Temperature as a proxy of true clustering separability}

The Temperature Function is an inexpensive empirical signal for deciding
\emph{when} to cluster, but it carries no formal guarantee on its own. This subsection studies when the Divergence Matrix is separable enough for any reasonable clustering algorithm to recover the true
partition; Section~\ref{experiments:margin:temperature} then verifies empirically that the temperature predominantly fires at such a round. We first characterise when the observed cosine-distance matrix is cluster-recoverable at a fixed round.  We then give a structural condition under which temperature tracks the corresponding population separation.
Accordingly, temperature is treated as a computationally inexpensive proposal signal; its first observed rise is not, by itself, a certificate of the earliest recoverable round. Let $\mathcal{C}=\{c_1,\ldots,c_G\}$, with $G\geq 2$, be the true DGP partition and define $z_i\in\{1,\ldots,G\}$ by $Z(p_i)=c_{z_i}$, using the client--cluster map $Z$ of
Section~\ref{methodology:data_process}.  At round $t$, client $i$ starts from $\theta^t$ and, after $K$ local optimiser steps, returns $\Delta_i^t=\theta_i^{t,K}-\theta^t$.  Thus, $G$ denotes the number of true clusters and $K$ retains its algorithmic meaning as the number of local steps.  The quantities used by OCFL are:

\begin{equation}
\boldsymbol{\Gamma}_t(i,j)
=
d_{\cos}\!\left(\Delta_i^t,\Delta_j^t\right),
\qquad
T_t
=
\frac{\lVert\boldsymbol{\Gamma}_t\rVert_p}{\lambda_p}.
\label{eq:minimal-observed-objects}
\end{equation}

\paragraph{Clustered-update assumption.}
At a fixed round $t$, suppose that every local update admits the decomposition

\begin{equation}
\Delta_i^t
=
\mu_{z_i}^t+\xi_i^t,
\qquad
\max_{1\leq i\leq n}\lVert\xi_i^t\rVert_2
\leq
\varepsilon_t,
\qquad
\min_{1\leq c\leq G}\lVert\mu_c^t\rVert_2
\geq
r_t>0.
\label{eq:clustered-update-minimal}
\end{equation}

Here, $\mu_c^t$ is the population update direction of DGP cluster $c$, while $\xi_i^t$ collects within-DGP heterogeneity, finite-sample stochasticity, and local-optimisation drift.  Define:

\begin{equation}
\rho_t
=
\frac{\varepsilon_t}{r_t},
\qquad
B_t
=
\min_{c\neq d}
d_{\cos}\!\left(\mu_c^t,\mu_d^t\right).
\label{eq:minimal-rho-B}
\end{equation}

\begin{lemma}[Perturbation of normalised directions]
\label{lemma:perturb}
Let $a,x\in\mathbb{R}^d$ with $a\neq 0$ and $\lVert x-a\rVert \leq \varepsilon \leq \lVert a\rVert/2$. Then
\begin{equation}
\left\lVert
\frac{x}{\lVert x\rVert}
- \frac{a}{\lVert a\rVert}
\right\rVert
\leq 2\frac{\varepsilon}{\lVert a\rVert}.
\end{equation}
\end{lemma}
The proof of Lemma \ref{lemma:perturb} is reported in Appendix~\ref{appendix:lemmaperturb}.

\begin{theorem}[Cosine-distance recovery under angular separation] 
\label{th:cosine-distance}
Suppose the clustered-update conditions in Equation~\eqref{eq:clustered-update-minimal} hold at round $t$ and $\rho_t \leq 1/2$. Then:

\begin{enumerate}[label=(\roman*)]
\item If $z_i=z_j$, then $ d_{\cos}(\Delta_i^t,\Delta_j^t) \leq 8\rho_t^2$.

\item If $z_i\neq z_j$, then $ d_{\cos}(\Delta_i^t,\Delta_j^t) \geq B_t-4\rho_t$.
\end{enumerate}

In particular, thresholding the observed dissimilarities at any value $h_t\in(8\rho_t^2,B_t-4\rho_t)$ yields a graph whose connected components are the true DGP clusters, up to label permutation. A different clustering algorithm inherits exact recovery only when it is consistent under this strict block-separation condition.
\end{theorem}
The full proof of Theorem~\ref{th:cosine-distance} is reported in Appendix \ref{appendix:cosine-distance}.

Theorem~\ref{th:cosine-distance} is a fixed-round result.  OCFL selects the data-dependent round:

\begin{equation}
\tau_T
=
\inf
\left\{
t\geq 1:
T_t>T_{t-1}
\right\}.
\label{eq:patch-temperature-trigger}
\end{equation}

The theorem does not assert that $\tau_T$ is the first recoverable round, or even that it is recoverable in every run. We therefore connect the trigger to the theorem through the observed separation margin evaluated at $\tau_T$ in the controlled experiments.

\subsection{Observed DGP-separation margin at the triggered round}

In the simulated experiments, the DGP label $z_i$ of every client is known for evaluation.  For any round $t$, define the observed within-DGP diameter, between-DGP separation, and strict separation margin by

\begin{equation}
W_t^\star
=
\max_{\substack{i,j\in P\\z_i=z_j}}
\boldsymbol{\Gamma}_t(i,j),
\qquad
B_t^\star
=
\min_{\substack{i,j\in P\\z_i\neq z_j}}
\boldsymbol{\Gamma}_t(i,j),
\qquad
M_t^\star
=
B_t^\star-W_t^\star.
\label{eq:patch-true-margin}
\end{equation}

The star distinguishes these observed, label-based quantities from the latent population separation $B_t$ in Equation~\eqref{eq:minimal-rho-B}. Theorem~\ref{th:cosine-distance} implies
$M_t^\star
\geq
B_t-4\rho_t-8\rho_t^2.$
Consequently, $M_t^\star>0$ means that every observed between-DGP dissimilarity is larger than every observed within-DGP dissimilarity. It is therefore a direct empirical check that the divergence matrix is strictly block-separated according to the true simulated partition at round $t$. We evaluate this quantity at $t=\tau_T$. 

\section{Experiments}
\label{sec:experiments}
In this section, we present  a number of experiments evaluating the performance of \ocfl algorithm (Subsection \ref{experiments:clustering_correctness}), the impact of the clustering on the personalisation (performance on the local test set) and generalisation (performance on the hold-out test set) (Subsection \ref{experiments:clustering_performance}) and the behaviour of the \textit{Clustering Temperature} Function (Subsection \ref{experiment: temperature_function}). The data on which the tests were evaluated is reported at the beginning of this section (Subsection \ref{experiments:datasets}), and the experimental setting is reported in Subsection \ref{experiments:set-up}. The experimental work related to cluster-level explainability is presented in a separate section (Section \ref{sec:xai_experiments}).

\subsection{Datasets} \label{experiments:datasets}
We experiment on five different image classification datasets with increasing levels of difficulty: MNIST \cite{b15}, FMNIST \cite{b16}, CIFAR10 \cite{b17}, PathMNIST and BloodMNIST \cite{b48}. We use the methodology described in subsection \ref{methodology:data_process} to simulate four different types of data splits: \textbf{(i)} overlapping balanced, \textbf{(ii)} non-overlapping balanced, \textbf{(iii)} overlapping imbalanced and \textbf{(iv)} non-overlapping imbalanced for \textbf{(a)} 15 and \textbf{(b)} 30 clients. The combination of five datasets, four different splits, and two client cardinalities yields a total of 40 different scenarios. Given seven different algorithms to test, this implies 280 separate runs. The theoretical assumptions regarding the data splits were presented in Subsection \ref{methodology:data_process}. Details for the data splits can be found in Appendix \ref{appendix:datasplits}. All four split types vary only the label support and the label proportions across clusters, while the class-conditional feature distribution is that of the underlying dataset for every client; a controlled feature-skew evaluation remains outside the current empirical scope.

\subsection{Experimental Set-up} \label{experiments:set-up}
Given that our proposal is agnostic w.r.t. the actual clustering strategy, we pair it with density-based methods HDBSCAN \cite{b19} and Mean-Shift \cite{b20} and with the parameter free Affinity Propagation \cite{b21}. We use four baselines to evaluate generality: \textbf{Baseline No Clustering} BNC, the method of \cite{b7} (SCL), the hierarchical update clustering of \cite{b1} (BCL), and an idealised \ocfl with KMeans where the number of data generating distributions is known a priori (OCFL-KM). These are compared to our OCFL combined with HDBSCAN (OCFL-HDBS), Mean-Shift (OCFL-MS) and Affinity Propagation (OCFL-AFF). The next section describes the experiments and comparisons.
For MNIST and FMNIST we use a deep convolutional network inspired by Tiny VGG \cite{b28}. For CIFAR10, PathMNIST and BloodMNIST we use ResNet34 as in \cite{b18}. In centralised experiments we obtained about 99\% and 94\% test accuracy on MNIST and FMNIST, and roughly 82\% on CIFAR10. On PathMNIST and BloodMNIST ResNet34 matched the results reported by \cite{b48}. To avoid excessive hyperparameter tuning affecting clustering, we trained with SGD and a shared local learning rate of 0.01 for MNIST, FMNIST and CIFAR10. Batch sizes were 32 for MNIST and FMNIST and 64 for CIFAR10. For PathMNIST we used the Adam optimiser introduced by \cite{b52} with a learning rate of $10^{-7}$, batch size 128 and weight decay 0.01. For BloodMNIST Adam was used with a learning rate of $10^{-6}$, batch size 128 and weight decay 0.01. A summary of these settings appears in Table~\ref{tab:general_hyperparameters} in the Appendix.
We adopt a \fedopt baseline for the federated training as described in Section \ref{RelatedWork:FL}. We reimplemented \cite{b1} and \cite{b7} using their pseudocode and public code for \cite{b7}.\footnote{The code used for reimplementation of the work of \cite{b7} can be found under the following link: \url{https://github.com/felisat/clustered-federated-learning}.} The original implementation of \cite{b7} includes a minimal number of rounds after which clustering can occur, so we kept that hyperparameter. The authors recommend setting epsilon values via the maximal gradient magnitude, which would require running all splits to measure gradients and then re-running experiments, an impractical workflow for our study. Instead we derive proxy values from centralised training, $\Delta\hat{\theta}^t = |SGD(\theta^t, D) - SGD(\theta^{t-1}, D)|$, and use the convergence round to pick a cool-down period. We then follow \cite{b7} to choose $\epsilon_1$ and $\epsilon_2$, resorting to a small heuristic search if needed. This procedure yielded $\epsilon_1 = 0.35$ and $\epsilon_2 = 1.00$ for MNIST and FMNIST but not for CIFAR10, PathMNIST or BloodMNIST. For the method of \cite{b1} we tuned a distance threshold in the [0.05, 0.4] range using cosine similarity and average linkage, retaining the best average results.
Each dataset split was trained for 50 global rounds for MNIST and FMNIST, 75 for PathMNIST and BloodMNIST, and 80 for CIFAR10, with three local epochs per round. Because HDBSCAN requires a minimal cluster size and we aim to keep clustering hyperparameter free, we set the minimal cluster size to 20\% of the sample size. All experiments use $p=2$ in Definition~\ref{def:Temperature_Function}; hence temperature is the normalised Frobenius norm of the divergence matrix.

\subsection{Clustering Evaluation} \label{experiments:clustering_correctness}
We evaluate how accurately each method recovers the true DGP partition. For every run, the inferred partition is compared with the ground-truth partition using the adjusted Rand index (ARI), adjusted mutual information (AMI), and homogeneity (HOM; every inferred cluster contains clients of a single true cluster). Scores are averaged over the first 50 rounds of each run, which rewards methods that identify the correct structure early; ARI and AMI penalise both premature incorrect splits and persistent over-clustering.
Aggregated results for all five datasets across every possible split are reported in Table \ref{tab.Personalization_via_Clustering_Aggregated_Clustering}. In aggregate, OCFL-HDBS and OCFL-MS are the strongest non-oracle variants: ARI values are approximately $0.95$ on MNIST, $0.96$ on FMNIST, about $0.80$ on CIFAR10, and generally between $0.80$ and $0.98$ on PathMNIST and BloodMNIST. Overlapping-balanced splits are the most difficult. SCL frequently does not split on CIFAR10 and PathMNIST and BloodMNIST, whereas BCL and OCFL-AFF often produce overly granular partitions.
 
\begin{table}
\centering
\resizebox{0.75\columnwidth}{!}{%
\begin{tabular}{cccccccc|cccccc}
\tiny
&&\multicolumn{6}{c}{Non Overlapping} & \multicolumn{6}{c}{Overlapping} \\
&&\multicolumn{3}{c}{Balanced} & \multicolumn{3}{c}{Imbalanced} & \multicolumn{3}{c}{Balanced} & \multicolumn{3}{c}{Imbalanced} \\
&& ARI & AMI & HOM & ARI & AMI & HOM & ARI & AMI & HOM & ARI & AMI & HOM \\

\toprule
\multirow{6}{*}{MNIST 15} & SCL & 0.44 & 0.46 & 0.57 & 0.57 & 0.57 & 0.57 & 0.21 & 0.31 & 0.57 & 0.29 & 0.38 & 0.57\\
& BCL & -- & -- & -- & -- & -- & -- & -- & -- & -- & -- & -- & -- \\
& OCFL-KM & \textbf{0.96} & \textbf{0.96 }& \textbf{0.96} & \textbf{0.96} & \textbf{0.96} & \textbf{0.96} & \textbf{0.96} & \textbf{0.96} & \textbf{0.96} & \textbf{0.96} & \textbf{0.96} & \textbf{0.96} \\
& OCFL-AFF & 0.23 & 0.27 & 0.26 & 0.13 & 0.06 & 0.12 & 0.23 & 0.27 & 0.26 & 0.30 & 0.30 & 0.30 \\
& OCFL-MS & 0.71 & 0.67 & \textbf{0.96} & \textbf{0.96} & \textbf{0.96} & \textbf{0.96} & 0.87 & 0.87 & \textbf{0.96} & \textbf{0.96} & \textbf{0.96} & \textbf{0.96} \\
& OCFL-HDBS & \textbf{0.96} & \textbf{0.96 }& \textbf{0.96} & \textbf{0.96} & \textbf{0.96} & \textbf{0.96} & \textbf{0.96} & \textbf{0.96} & \textbf{0.96} & \textbf{0.96} & \textbf{0.96} & \textbf{0.96}  \\
\cmidrule(lr{0.150em}){3-5}\cmidrule(lr{0.150em}){6-8}\cmidrule(lr{0.150em}){9-11}\cmidrule(lr{0.150em}){12-14}

\multirow{6}{*}{MNIST 30} & SCL & 0.45 & 0.49 & 0.57 & 0.57 & 0.58 & 0.57 & 0.19 & 0.30 & 0.57 & 0.26 & 0.33 & 0.57 \\
& BCL & -- & -- & -- & -0.04 & -0.07 & 0.03 & -- & -- & -- & -- & -- & -- \\
& OCFL-KM & \textbf{0.96} & \textbf{0.96} & \textbf{0.96} & 0.96 & 0.96 & 0.96 & 0.90 & 0.90 & 0.90 & \textbf{0.94} & \textbf{0.94} & \textbf{0.94} \\
& OCFL-AFF & 0.38 & 0.44 & 0.37 & 0.66 & 0.63 & 0.65 & 0.36 & 0.42 & 0.35 & 0.50 & 0.49 & 0.42 \\
& OCFL-MS & 0.85 & 0.84 & 0.94 & 0.94 & 0.94 & 0.94 & 0.87 & 0.87 & 0.92 & \textbf{0.94} & \textbf{0.94} & \textbf{0.94} \\
& OCFL-HDBS & 0.92 & 0.92 & 0.92 & \textbf{0.98} & \textbf{0.98} & \textbf{0.98} & \textbf{0.94} & \textbf{0.94} & \textbf{0.94} & \textbf{0.94} & \textbf{0.94} & \textbf{0.94}  \\
\cmidrule(lr{0.150em}){3-5}\cmidrule(lr{0.150em}){6-8}\cmidrule(lr{0.150em}){9-11}\cmidrule(lr{0.150em}){12-14}

\multirow{6}{*}{FMNIST 15} & SCL & 0.46 & 0.48 & 0.57 & 0.55 & 0.53 & 0.57 & 0.19 & 0.28 & 0.57 & 0.17 & 0.27 & 0.57  \\
& BCL & -0.14 & -0.23 & 0.00 & -- & -- & -- & -- & -- & -- & -- & -- & -- \\
& OCFL-KM & \textbf{0.98} & \textbf{0.98} & \textbf{0.98} & \textbf{0.98} & \textbf{0.98} & \textbf{0.98} & \textbf{0.98} & \textbf{0.98} & \textbf{0.98} & \textbf{0.98} & \textbf{0.98} & \textbf{0.98} \\
& OCFL-AFF & 0.23 & 0.27 & 0.27 & 0.22 & 0.25 & 0.26 & 0.23 & 0.27 & 0.26 & 0.30 & 0.31 & 0.31 \\
& OCFL-MS & \textbf{0.98} & \textbf{0.98} & \textbf{0.98} & \textbf{0.98} & \textbf{0.98} & \textbf{0.98} & \textbf{0.98} & \textbf{0.98} & \textbf{0.98} & \textbf{0.98} & \textbf{0.98 }& \textbf{0.98} \\
& OCFL-HDBS & 0.96 & 0.96 & 0.96 & \textbf{0.98} & \textbf{0.98} & \textbf{0.98} & 0.96 & 0.96 & 0.96 & \textbf{0.98} & \textbf{0.98} & \textbf{0.98 } \\
\cmidrule(lr{0.150em}){3-5}\cmidrule(lr{0.150em}){6-8}\cmidrule(lr{0.150em}){9-11}\cmidrule(lr{0.150em}){12-14}

\multirow{6}{*}{FMNIST 30} & SCL & 0.57 & 0.57 & 0.57 & 0.57 & 0.58 & 0.57 & 0.42 & 0.43 & 0.57 & 0.53 & 0.52 & 0.57 \\
& BCL & -- & -- & -- & -- & -- & -- & -0.07 & -0.15 & 0.00 & -0.05 & -0.09 & 0.00 \\
& OCFL-KM & 0.96 & 0.96 & 0.96 & \textbf{0.98} & \textbf{0.98} & \textbf{0.98} & \textbf{0.96} & \textbf{0.96} & \textbf{0.96} & 0.96 & 0.96 & 0.96 \\
& OCFL-AFF & 0.39 & 0.45 & 0.37 & 0.66 & 0.64 & 0.67 & 0.37 & 0.43 & 0.36 & 0.51 & 0.50 & 0.43 \\
& OCFL-MS & 0.85 & 0.84 & \textbf{0.98} & 0.96 & 0.96 & 0.96 & 0.94 & 0.94 & 0.94 & \textbf{0.98} & \textbf{0.98} & \textbf{0.98} \\
& OCFL-HDBS & \textbf{0.98} & \textbf{0.98} & \textbf{0.98} & \textbf{0.98} & \textbf{0.98} & \textbf{0.98} & 0.94 & 0.94 & 0.94 & 0.96 & 0.96 & 0.96  \\
\cmidrule(lr{0.150em}){3-5}\cmidrule(lr{0.150em}){6-8}\cmidrule(lr{0.150em}){9-11}\cmidrule(lr{0.150em}){12-14}

\multirow{6}{*}{CIFAR 15} & SCL & -- & -- & -- & -- & -- & -- & -- & -- & -- & -- & -- & -- \\
& BCL & 0.00 & 0.00 & 0.60 & 0.00 & 0.00 & 0.60 & 0.00 & 0.00 & 0.60 & 0.00 & 0.00 & 0.60 \\
& OCFL-KM & 0.35 & 0.40 & 0.43 & 0.52 & 0.60 & 0.60 & 0.90 & 0.90 & 0.90 & 0.68 & 0.67 & 0.66 \\
& OCFL-AFF & -- & -- & -- & -- & -- & -- & 0.08 & 0.11 & 0.21 & 0.35 & 0.32 & 0.31 \\
& OCFL-MS & 0.23 & 0.26 & 0.57 & 0.16 & 0.18 & 0.58 & -- & -- & -- & 0.73 & 0.67 & 0.73 \\
& OCFL-HDBS & \textbf{0.73} & \textbf{0.75} & \textbf{0.77} & \textbf{0.84} & \textbf{0.84} & \textbf{0.84} & \textbf{0.59} & \textbf{0.66} & \textbf{0.69} & \textbf{0.77} & \textbf{0.74} & \textbf{0.74}  \\
\cmidrule(lr{0.150em}){3-5}\cmidrule(lr{0.150em}){6-8}\cmidrule(lr{0.150em}){9-11}\cmidrule(lr{0.150em}){12-14}

\multirow{6}{*}{CIFAR 30} & SCL & -- & -- & -- & -- & -- & -- & -- & -- & -- & -- & -- & -- \\
& BCL & 0.00 & 0.00 & 0.60 & 0.00 & 0.00 & 0.60 & 0.00 & 0.00 & 0.60 & 0.00 & 0.00 & 0.60 \\
& OCFL-KM & 0.37 & 0.54 & 0.51 & 0.42 & 0.56 & 0.56 & \textbf{0.90} &\textbf{ 0.90} & \textbf{0.90} & 0.59 & 0.55 & 0.58 \\
& OCFL-AFF & 0.35 & 0.41 & 0.34 & 0.46 & 0.45 & 0.39 & 0.23 & 0.22 & 0.20 & 0.32 & 0.30 & 0.26 \\
& OCFL-MS & 0.63 & 0.55 & 0.86 & 0.56 & 0.63 & 0.58 & -- & -- & -- & -- & -- & -- \\
& OCFL-HDBS & \textbf{0.77} &\textbf{0.77} & \textbf{0.77} & \textbf{0.88} & \textbf{0.88} & \textbf{0.88} & 0.64 & 0.69 & 0.69 & \textbf{0.83} & \textbf{0.81} & \textbf{0.81}  \\
\cmidrule(lr{0.150em}){3-5}\cmidrule(lr{0.150em}){6-8}\cmidrule(lr{0.150em}){9-11}\cmidrule(lr{0.150em}){12-14}

\multirow{6}{*}{PATHMNIST 15} & SCL & -- & -- & -- & -- & -- & -- & -- & -- & -- & -- & -- & -- \\
& BCL & 0.00 & 0.00 & 0.60 & 0.00 & 0.00 & 0.60 & 0.00 & 0.00 & 0.60 & 0.00 & 0.00 & 0.60 \\
& OCFL-KM & \textbf{0.96} & \textbf{0.96} & \textbf{0.96} & \textbf{0.96 }& \textbf{0.96} & \textbf{0.96} & 0.50 & 0.62 & 0.60 & 0.96 & 0.96 & 0.96 \\
& OCFL-AFF & 0.23 & 0.27 & 0.27 & 0.12 & 0.13 & 0.16 & 0.08 & 0.06 & 0.12 & 0.22 & 0.25 & 0.26 \\
& OCFL-MS & 0.78 & 0.80 & 0.98 & 0.38 & 0.58 & 0.46 & 0.45 & 0.55 & 0.62 & \textbf{0.96} & \textbf{0.96} & \textbf{0.96} \\
& OCFL-HDBS & 0.94 & 0.94 & 0.94 & \textbf{0.96} & \textbf{0.96} & \textbf{0.96} & \textbf{0.88 }& \textbf{0.89} & \textbf{0.98} & \textbf{0.96} & \textbf{0.96} & \textbf{0.96}  \\
\cmidrule(lr{0.150em}){3-5}\cmidrule(lr{0.150em}){6-8}\cmidrule(lr{0.150em}){9-11}\cmidrule(lr{0.150em}){12-14}

\multirow{6}{*}{PATHMNIST 30} & SCL & -- & -- & -- & -- & -- & -- & -- & -- & -- & -- & -- & -- \\
& BCL & 0.00 & 0.00 & 0.60 & 0.00 & 0.00 & 0.60 & 0.00 & 0.00 & 0.60 & 0.00 & 0.00 & 0.60 \\
& OCFL-KM & \textbf{0.98} & \textbf{0.98} & \textbf{0.98} & \textbf{0.96} & \textbf{0.96} & \textbf{0.96} & 0.51 & 0.58 & 0.60 & 0.98 & 0.98 & 0.98 \\
& OCFL-AFF & 0.39 & 0.45 & 0.37 & 0.29 & 0.36 & 0.29 & 0.39 & 0.45 & 0.37 & 0.32 & 0.30 & 0.35 \\
& OCFL-MS & 0.45 & 0.60 & 0.57 & 0.45 & 0.64 & 0.49 & 0.51 & 0.60 & 0.63 & \textbf{0.98 }& \textbf{0.98} & \textbf{0.98} \\
& OCFL-HDBS & \textbf{0.98} & \textbf{0.98} & \textbf{0.98} & \textbf{0.96} & \textbf{0.96} & \textbf{0.96} & \textbf{0.54} & \textbf{0.71} & \textbf{0.57} & \textbf{0.98} & \textbf{0.98} & \textbf{0.98}  \\
\cmidrule(lr{0.150em}){3-5}\cmidrule(lr{0.150em}){6-8}\cmidrule(lr{0.150em}){9-11}\cmidrule(lr{0.150em}){12-14}

\multirow{6}{*}{BLOODMNIST 15} & SCL & -- & -- & -- & -- & -- & -- & -- & -- & -- & -- & -- & -- \\
& BCL & 0.00 & 0.00 & 0.60 & 0.00 & 0.00 & 0.60 & 0.00 & 0.00 & 0.60 & 0.00 & 0.00 & 0.60 \\
& OCFL-KM & \textbf{0.98} & \textbf{0.98} & \textbf{0.98} & \textbf{0.92} & \textbf{0.92} & \textbf{0.92} & \textbf{0.98} & \textbf{0.98} & \textbf{0.98} & \textbf{0.98} & \textbf{0.98} & \textbf{0.98} \\
& OCFL-AFF & 0.23 & 0.27 & 0.27 & 0.12 & 0.13 & 0.16 & 0.23 & 0.27 & 0.27 & 0.31 & 0.31 & 0.31 \\
& OCFL-MS & 0.78 & 0.80 & \textbf{0.98} & \textbf{0.92} & \textbf{0.92} & \textbf{0.92} & 0.83 & 0.86 & 0.98 & 0.65 & 0.74 & \textbf{0.98} \\
& OCFL-HDBS & \textbf{0.98} & \textbf{0.98} & \textbf{0.98} & \textbf{0.92} &\textbf{ 0.92} & \textbf{0.92} & \textbf{0.98} & \textbf{0.98} & \textbf{0.98} & \textbf{0.98} & \textbf{0.98} & \textbf{0.98}  \\
\cmidrule(lr{0.150em}){3-5}\cmidrule(lr{0.150em}){6-8}\cmidrule(lr{0.150em}){9-11}\cmidrule(lr{0.150em}){12-14}

\multirow{6}{*}{BLOODMNIST 30} & SCL & -- & -- & -- & -- & -- & -- & -- & -- & -- & -- & -- & -- \\
& BCL & 0.00 & 0.00 & 0.60 & 0.00 & 0.00 & 0.60 & 0.00 & 0.00 & 0.60 & 0.00 & 0.00 & 0.60 \\
& OCFL-KM & \textbf{0.98} & \textbf{0.98} & \textbf{0.98} & \textbf{0.98} & \textbf{0.98} & \textbf{0.98} & \textbf{0.98} & \textbf{0.98} & \textbf{0.98} & \textbf{0.54} & \textbf{0.62} & \textbf{0.66} \\
& OCFL-AFF & 0.39 & 0.45 & 0.37 & 0.50 & 0.50 & 0.43 & 0.39 & 0.45 & 0.37 & 0.32 & 0.37 & 0.32 \\
& OCFL-MS & 0.93 & 0.93 & \textbf{0.98} & \textbf{0.98} & \textbf{0.98} & \textbf{0.98} & 0.53 & 0.67 & 0.59 & \textbf{0.54} & \textbf{0.62} & \textbf{0.66} \\
& OCFL-HDBS & \textbf{0.98} & \textbf{0.98} & \textbf{0.98} & \textbf{0.98} & \textbf{0.98} & \textbf{0.98} & \textbf{0.54} & \textbf{0.71} & \textbf{0.57} & 0.48 & 0.58 & 0.48  \\

\bottomrule
\end{tabular}}
\caption{Clustering performance measured in terms of Adjusted Rand Index (ARI), Adjusted Mutual Information Score (AMI) and Homogeneity Score (HOM) for all datasets (with a number of clients equal to 15 or 30), data splits, and across six different clustering algorithms: \cite{b7} (SCL), \cite{b1} (BCL), OCFL with K-Means Algorithm (OCFL-KM), OCFL with Affinity Propagation (OCFL-AFF), OCFL with Mean-Shift algorithm (OCFL-MS) and OCFL with HDBSCAN algorithm (OCFL-HDBS). The double hyphen (--) is placed if the algorithm performed no split (all clients remained in a single cluster) in the given scenario.}
\label{tab.Personalization_via_Clustering_Aggregated_Clustering}
\end{table}

\subsection{Local and Global Performance} \label{experiments:clustering_performance}
\begin{figure*}
    \centering
    \includegraphics[width=0.80\linewidth]{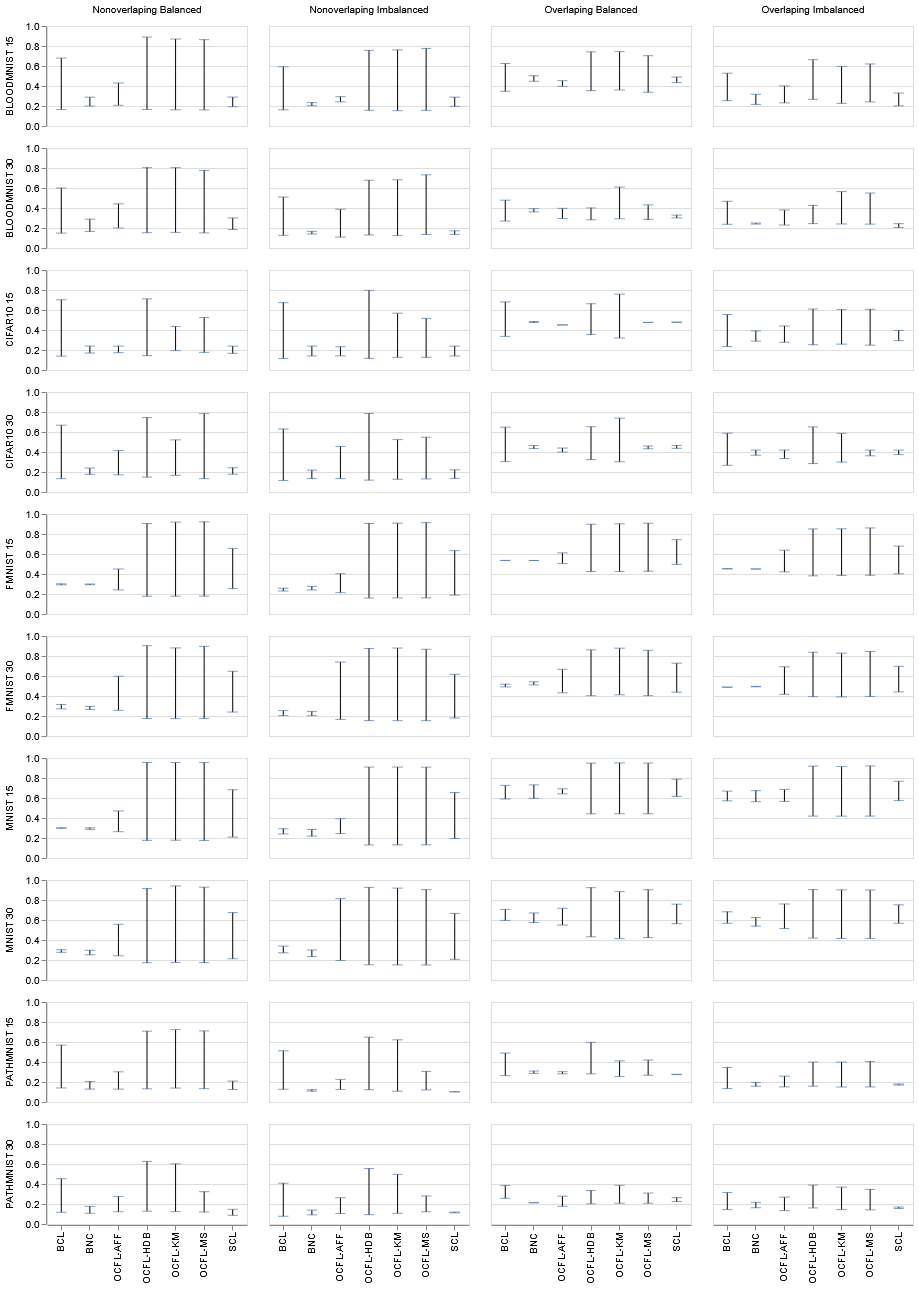}
    \Description{Segment plot of mean personalised (upper end) and generalised (lower end) F1 scores for each dataset and clustering method; segment length is the learning gap.}
    \caption{Average performance of models in terms of F1-score. In the plot, the lower part of each segment indicates generalised F1-score (GF1), while the upper part indicates personalised F1-score (PF1). The length of the segment represents the learning gap as defined in Equation~\eqref{eq:learning-gap}. The GF1 is calculated using a hold-out test set that is uniformly distributed across all the classes. The PF1 is calculated using a local test set - a hold-out sample taken from the local training set before the start of the training for each of the nodes.}
    \label{fig:experiment: aggregated_performance}
\end{figure*}

We define personalisation as the mean F1 score on client hold-out sets (PF1) and generalisation as the mean F1 score on the uniformly distributed orchestrator hold-out set (GF1). Their absolute difference is the learning gap
\begin{equation}
\operatorname{LG}(h,D_i,D_o)
=
\left|f(h,D_i)-f(h,D_o)\right|.
\label{eq:learning-gap}
\end{equation}
We opted for a uniformly distributed dataset across the labels for an orchestrator test set $D_{o}$. We utilise the fact that, as described in the Section \ref{experiments:datasets}, the orchestrator dataset consists of a held-out test dataset, which makes it a proper candidate for testing the generalisation capabilities of the model. A useful personalised model should increase PF1 without an excessive reduction in GF1. The results of the experiments are presented in Figure \ref{fig:experiment: aggregated_performance}. Correct OCFL partitions substantially increase PF1 while generally preserving GF1 relative to the alternative clustering methods. Conversely, singleton and near-singleton partitions perform poorly: local data alone are often insufficient, and clients benefit from sharing information with other clients generated by the same or a sufficiently similar DGP.

\subsection{Behaviour of Temperature Function}\label{experiment: temperature_function}
\begin{figure}
    \centering
    \includegraphics[width=1\linewidth]{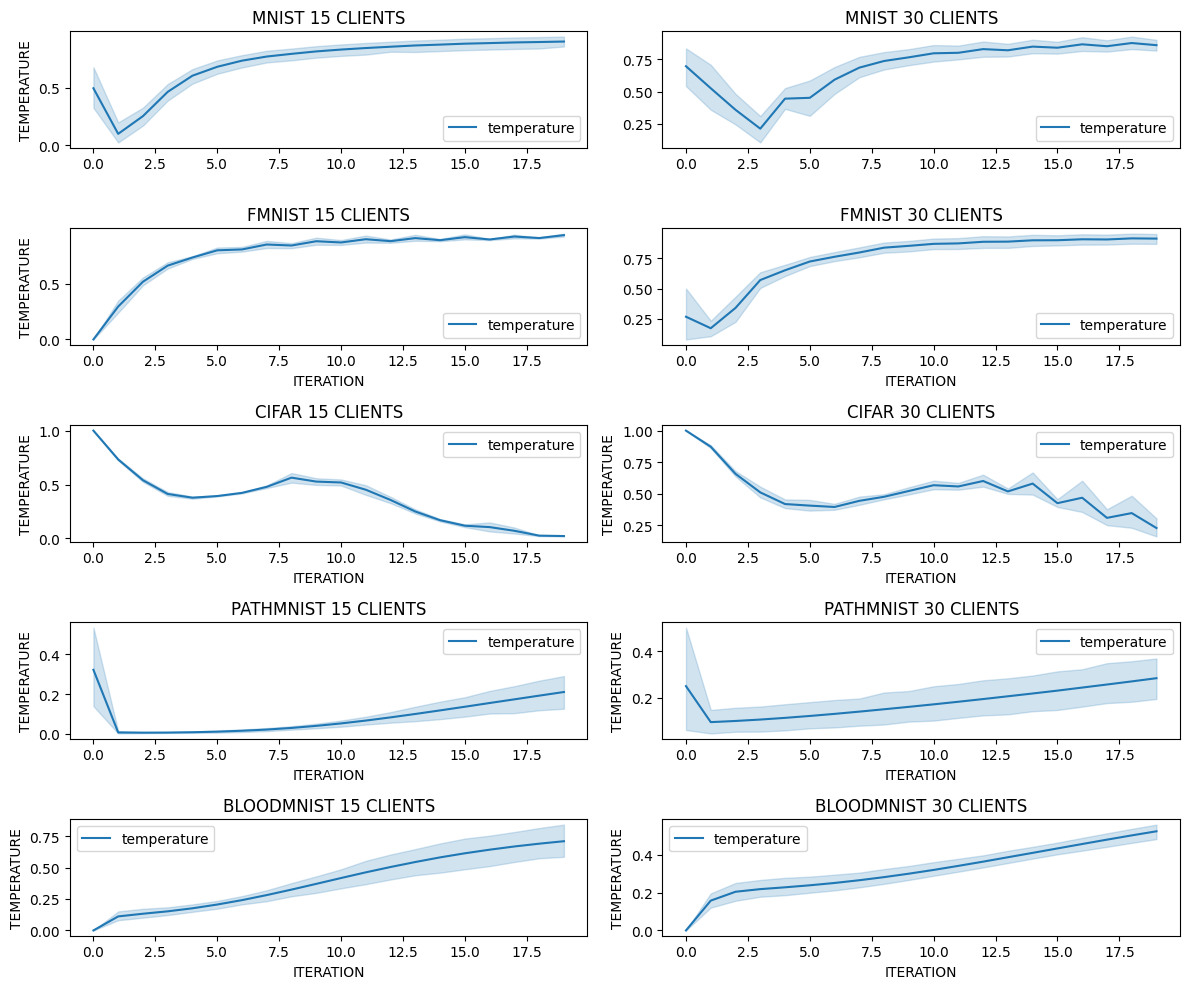}
    \Description{Line plots of the normalised Temperature Function over training rounds for each dataset and client count, showing the mean across data splits with a shaded confidence band.}
    \caption{Behaviour of the (normalised) Temperature Function as defined in the Definition \ref{def:Temperature_Function}. A solid blue line symbolises the mean across all the scenarios (overlapping and non-overlapping with balanced and imbalanced settings). Blue shadow represents confidence intervals established based on the variations between different scenarios.}
    \label{fig:stacked_by_split_temperature}
\end{figure}

In order to evaluate the behaviour of the Temperature Function (Definition \ref{def:Temperature_Function}) that is used as a method of detecting suitable clustering moment, we register the temperature of the baseline cluster each round for all the experiments performed on the five selected datasets. Visualisation of the experiments is reported in Figure \ref{fig:stacked_by_split_temperature}. Each plot presents a stacked representation of each task in a given number of clients. A solid blue line symbolises the mean across all the scenarios (overlapping and non-overlapping in balanced and imbalanced settings). Blue shadow represents confidence intervals established based on the variations between different scenarios. The temperature function is normalised in $[0,1]$ for comparison purposes. As explained in the Section \ref{methodology:function_similarity}, this can be done easily with a scaling constant $\lambda$. However, it does not have any practical effect on the efficiency of Algorithm \ref{algo:ocfl} and was done in order to render the visualisation more accessible.

Moreover, the figure in Appendix~\ref{app:temperature_function_additional} presents an averaged behaviour of the Temperature Function across all the datasets with corresponding confidence intervals. This graph should be approached with caution, as the data presented on it combines tasks from the same domain (classification in computer vision) but performed on different datasets. However, since it can also be interpreted as \textit{on-average} behaviour of the Temperature Function, it is presented together with individual averages displayed in Figure \ref{fig:stacked_by_split_temperature}. Individual graphs for each particular run are placed in Appendix~\ref{app:temperature_individual}.

\subsection{Separation margin at the triggered round}
\label{experiments:margin:temperature}

At the temperature-triggered round $\tau_T$ we evaluate the \textbf{separation margin}
$M^\star_{\tau_T}=B^\star_{\tau_T}-W^\star_{\tau_T}$ (Equation~\eqref{eq:patch-true-margin}), where
$B^\star$ is the smallest between-DGP and $W^\star$ the largest within-DGP cosine dissimilarity.
By Theorem~\ref{th:cosine-distance}, $M^\star_{\tau_T}\ge B_t-4\rho_t-8\rho_t^2$, so
$M^\star_{\tau_T}>0$ certifies that the divergence matrix is \emph{strictly block-separated}
according to the true simulated partition at $\tau_T$: the \textbf{partition is recoverable at the round
temperature fires.} Table~\ref{tab:margin} reports $M^\star_{\tau_T}$ across all forty scenarios.

Because $M^\star_{\tau_T}$ is the most extreme statistic -- determined by a single closest between-DGP
pair and a single widest within-DGP pair -- we complement it with the \emph{pair-order agreement}
$A^\star_{\tau_T}$, the fraction of (between, within) pair combinations that are correctly ordered,
i.e.\ the empirical probability that a between-DGP dissimilarity exceeds a within-DGP one
(the Mann-Whitney statistic of the entries of $\boldsymbol{\Gamma}_{\tau_T}$ under the true
partition). By construction $M^\star_{\tau_T}>0$ if and only if $A^\star_{\tau_T}=1$, so
$A^\star_{\tau_T}$ grades how far a non-separated matrix is from strict block order. The full explanation of how $A^\star_{\tau_T}$ was calculated is placed in Appendix \ref{app:margin-protocol}.

The margin is strictly positive in $25/40$ scenarios: in \emph{all} sixteen MNIST/FMNIST runs and
in every run where OCFL recovers the partition (a positive margin co-occurs with a median ARI of
$0.98$, cf.\ Table~\ref{tab.Personalization_via_Clustering_Aggregated_Clustering}). Where strict
separation fails, $A^\star_{\tau_T}$ distinguishes two regimes. On PathMNIST and BloodMNIST the
negative margins arise from isolated boundary pairs in a near-degenerate geometry
($|M^\star_{\tau_T}|\!\sim\!10^{-5}$) while the bulk of the matrix remains block-ordered
($A^\star_{\tau_T}\in[0.94,1.00]$ and $[0.85,1.00]$, respectively) -- consistent with the fact that
density-based algorithms still recover the partition there, strict separation being sufficient but
not necessary. On CIFAR10, by contrast, the bulk order itself degrades
($A^\star_{\tau_T}$ down to $0.40$, mean $0.69$), the regime in which we independently observe
gradient collapse and degraded recovery. The following experiments support the claims that temperature is a good \textbf{empirical proxy} for the round at which the partition becomes recoverable; per-scenario values of
$B^\star_{\tau_T},W^\star_{\tau_T},M^\star_{\tau_T},A^\star_{\tau_T}$ are placed in
Appendix~\ref{app:margin_section}.

\begin{table}[t]
  \centering
  \caption{Ground-truth separation margin $M^\star_{\tau_T}$ at the
  temperature-triggered round, per dataset and client count $|P|$. $M^\star_{\tau_T}>0$ (bold)
  certifies strict block separation of the divergence matrix under the true partition. For cells
  with $M^\star_{\tau_T}\le 0$ we report in parentheses the pair-order agreement
  $A^\star_{\tau_T}$; note that $M^\star_{\tau_T}>0$ implies $A^\star_{\tau_T}=1$. The full explanation of how $A^\star_{\tau_T}$ was calculated is placed in Appendix \ref{app:margin-protocol}.}
  \label{tab:margin}
  \small
  \begin{tabular}{ll cccc}
    \toprule
     &  & \multicolumn{2}{c}{Non-overlapping} & \multicolumn{2}{c}{Overlapping} \\
    \cmidrule(lr){3-4}\cmidrule(lr){5-6}
    Dataset & $|P|$ & Balanced & Imbalanced & Balanced & Imbalanced \\
    \midrule
\multirow{2}{*}{MNIST} & 15 & {\boldmath $+6.0\!\times\!10^{-1}$} & {\boldmath $+7.4\!\times\!10^{-1}$} & {\boldmath $+2.3\!\times\!10^{-1}$} & {\boldmath $+1.2\!\times\!10^{-1}$} \\
 & 30 & {\boldmath $+6.2\!\times\!10^{-1}$} & {\boldmath $+4.4\!\times\!10^{-1}$} & {\boldmath $+2.5\!\times\!10^{-1}$} & {\boldmath $+2.2\!\times\!10^{-1}$} \\
\cmidrule(lr){1-6}
\multirow{2}{*}{FMNIST} & 15 & {\boldmath $+7.1\!\times\!10^{-1}$} & {\boldmath $+4.1\!\times\!10^{-1}$} & {\boldmath $+2.6\!\times\!10^{-1}$} & {\boldmath $+3.6\!\times\!10^{-1}$} \\
 & 30 & {\boldmath $+6.7\!\times\!10^{-1}$} & {\boldmath $+5.4\!\times\!10^{-1}$} & {\boldmath $+2.2\!\times\!10^{-1}$} & {\boldmath $+1.9\!\times\!10^{-1}$} \\
\cmidrule(lr){1-6}
\multirow{2}{*}{CIFAR10} & 15 & $-3.2\!\times\!10^{-4}$ {\scriptsize(0.67)} & $-3.3\!\times\!10^{-4}$ {\scriptsize(0.45)} & $-5.8\!\times\!10^{-5}$ {\scriptsize(0.85)} & $-7.9\!\times\!10^{-5}$ {\scriptsize(0.93)} \\
 & 30 & $-8.7\!\times\!10^{-4}$ {\scriptsize(0.67)} & $-1.4\!\times\!10^{-3}$ {\scriptsize(0.40)} & $-3.0\!\times\!10^{-4}$ {\scriptsize(0.78)} & $-2.8\!\times\!10^{-4}$ {\scriptsize(0.80)} \\
\cmidrule(lr){1-6}
\multirow{2}{*}{PathMNIST} & 15 & $-6.9\!\times\!10^{-6}$ {\scriptsize(0.94)} & {\boldmath $+3.1\!\times\!10^{-6}$} & $-1.1\!\times\!10^{-6}$ {\scriptsize(0.97)} & {\boldmath $+5.1\!\times\!10^{-6}$} \\
 & 30 & $-1.7\!\times\!10^{-6}$ {\scriptsize(0.98)} & $-4.5\!\times\!10^{-6}$ {\scriptsize(0.998)} & $-8.2\!\times\!10^{-6}$ {\scriptsize(0.94)} & {\boldmath $+5.0\!\times\!10^{-6}$} \\
\cmidrule(lr){1-6}
\multirow{2}{*}{BloodMNIST} & 15 & {\boldmath $+5.6\!\times\!10^{-5}$} & {\boldmath $+4.7\!\times\!10^{-5}$} & {\boldmath $+2.4\!\times\!10^{-5}$} & {\boldmath $+2.4\!\times\!10^{-5}$} \\
 & 30 & {\boldmath $+5.5\!\times\!10^{-5}$} & {\boldmath $+4.9\!\times\!10^{-5}$} & $-1.7\!\times\!10^{-5}$ {\scriptsize(0.998)} & $-7.0\!\times\!10^{-4}$ {\scriptsize(0.85)} \\

    \bottomrule
  \end{tabular}
\end{table}

\subsection{Discussion} \label{experiments:discussion}

\subsubsection{Clustering Performance}
\label{experiments:discussion:performance}
An overview of the experimental data can indicate the high performance of the \ocfl algorithm, especially when it is combined with density-based methods such as Mean-Shift or HDBSCAN. In terms of clustering performance (section \ref{experiments:clustering_correctness}), OCFL-HDBS and OCFL-MS often outperform all other compared algorithms, with the ARI scores around 0.95 on all splits across the MNIST dataset, 0.96 across the FMNIST dataset and 0.80 across the CIFAR10 dataset (Table \ref{tab.Personalization_via_Clustering_Aggregated_Clustering}). Similar values (ranging from 0.80 to 0.98) hold on the PathMNIST and BloodMNIST datasets. From all the presented splits, Overlapping Balanced seems to be the most challenging, as even the OCFL-HDBS and OCFL-MS exhibit sometimes lower ARI values on that particular split (around 0.60 for OCFL-HDBS on CIFAR 15 and CIFAR 30 and around 0.5 for OCFL-MS for PathMNIST 15 and PathMNIST 30), while sometimes even failing altogether to distinguish more than one cluster (the case of OCFL-MS on CIFAR 15 and CIFAR 30). Theoretical study combined with empirical analysis of inter-cluster and intra-cluster distance (based on a ground-truth), can shed some further insights into why the algorithm fails in the given scenario. According to the Table \ref{tab:margin}, MNIST and FMNIST are perfectly block-separable at a time of clustering, allowing for a clean cluster retrieval. In the case of CIFAR10, we observe a real and measurable separability-collapse in case of all the data splits. The model collapse is additionally empirically validated by just observing the temperature function, as presented in Figure \ref{fig:stacked_by_split_temperature}.

It must also be highlighted that all four data splits generated for the experiments were challenging. Contrary to the label swaps or other similar methods often employed in studies of \cfl, the presented data splits assume more subtle differences in the label distribution as described in Subsection \ref{experiments:datasets}. Such a subtlety may confound many clustering algorithms. Hence, a weaker performance of the OCFL-KMeans or OCFL-Affinity, in line with the failures of algorithms presented in \cite{b1} and \cite{b7} to distinguish nodes and attribute them to more than one cluster. In this context, a high performance of density-based algorithms is an even more valuable insight.

Regarding all the other cases, the respective ARI and Adjusted Mutual Information Score correspond with the size of empirical margin calculated at the time of clustering. It implies that the temperature function serves as a cheap proxy that correctly triggers at the moment where the block separation is possible. Another key observation drawn from the data is the type of algorithm that can perform correct clustering upon the moment where block separation is possible. We observe general difficulty of the distance-based algorithms and it concerns state-of-the-art algorithms (such as those introduced by \cite{b1} and \cite{b7}) and OCFL-AFF alike. Usage of those algorithms requires a proper tuning of hyperparameters, which is especially challenging in the federated environment (detailed explanation on our different approaches to fine-tuning of \cite{b7} is placed in Appendix \ref{app:sattler}). Conversely, the density-based algorithms are performing exceptionally well, failing only in cases where the block separation is not possible.

\subsubsection{Model Performance}
\label{experiments:discussion:model_performance}The proper clustering should be reflected in the performance of personalised models, and this is the case as presented in Subsection \ref{experiments:clustering_performance}. The baseline model is characterised by an F1-score between $0.24$ (CIFAR10 dataset, as illustrated in Figure~\ref{fig:experiment: aggregated_performance}) and $0.6$ (for MNIST dataset). However, the local models delivered by the OCFL-MS and OCFL-HDBS achieve an F1-score between $0.96$ (MNIST) and $0.65$ (CIFAR10). Based on that figure, we could distinguish a \textbf{performance upper bound} that measures the ability of the model to solve the task locally and \textbf{performance lower bound} that measures the ability of the model to solve the task on a broader set of examples, possibly out-of-distribution.

What must be noted is the relatively high performance upper bound for all the OCFL algorithms. It means that they can deliver an accurate and higher than the compared state-of-the-art solutions for personalisation of local learning. On the other hand, the family of OCFL algorithms generally tends to have similar or only slightly lower \textit{performance lower bound}, which would imply that the loss of generalisation capabilities is not detrimental when compared to other clustering methods. Figure \ref{fig:experiment: aggregated_performance} can illustrate it quite well, as the generalised F1-score (lower part of each segment) is more or less comparable for all the presented methods, with the baseline serving as the most generalised case for obvious reasons. However, the personalised F1-score (higher part of each segment) is visibly better for the family of \ocfl methods, with particular emphasis on the high effectiveness of the OCFL-HDBS.

\subsubsection{Behaviour of the Temperature Function}\label{experiments:discussion:temperature}
A separate discussion concerns the behaviour of the Temperature Function (Definition \ref{def:Temperature_Function}), distinguishing \textbf{(i)} early-training behaviour and \textbf{(ii)} asymptotic behaviour.
Early on, assumptions from Subsection \ref{methodology:clustering_procedure} hold: either \textbf{(i)} similarity first increases (distance and Temperature Function decrease) then diverges after a local minimum, or \textbf{(ii)} similarity decreases from the first round without a minimum. Case (i) dominates on MNIST/CIFAR10; case (ii) on FMNIST/PathMNIST; BloodMNIST depends on the split. In both, \ocfl clusters successfully (after the minimum or by round two).
Asymptotically, either \textbf{(i)} similarities rise (models pull apart) or \textbf{(ii)} collapse makes gradients uninformative. Case (i) dominates across MNIST, FMNIST, PathMNIST, and BloodMNIST (all splits; 15/30 clients); case (ii) appears only on CIFAR10. Although a drop to $0$ initially suggested \fedavg/\fedopt convergence, this is unsupported given poor generalisability (Figure \ref{fig:experiment: aggregated_performance}). A more plausible explanation is model collapse: differing distributions yield uninformative gradients after $3$ local epochs. Importantly, asymptotics matter less for \ocfl than early behaviour; even if CIFAR10 converges, clustering still yields gains (Figure \ref{fig:experiment: aggregated_performance}).

The empirical evaluation of this function aims mainly to answer a question whether it can serve as an \textit{inexpensive proxy} for detecting the federated round at which the observed cosine-distance matrix is cluster-recoverable (block-separable). The experiments clearly show that - for the majority of cases - the cosine-distance matrix will be block-separable at a round indicated by the Temperature Function. The margin $M^\star_{\tau_T}$ will range from $+7.1 \times 10^{-1}$ to $+3.1 \times 10^{-6}$. Even if the matrix is not clearly separable, the negative values tend to be small in magnitude (from $-1.4 \times 10^{-3}$ to $-1.1 \times 10^{-6}$). The collapse of block-separability can be explained by analysing Lemma \ref{lemma:perturb}. The Equation~\eqref{eq:clustered-update-minimal} assumes that the update of client $i$ at round $t$ ($\Delta_i^t$) can be decomposed into the population update direction of DGP cluster $c$ ($\mu_{z_i}^t$) and the term that collects within-DGP heterogeneity, finite-sample stochasticity and local-optimisation drift ($\xi_i^t$). The magnitude of the second term can explain cases where the block-separability is impossible, e.g., the CIFAR10 case and some individual cases within the PathMNIST and BloodMNIST datasets. Separate discussion - not directly assessed in this work -- concerns methods and techniques that can control the magnitude of the term $\xi_i^t$.

\section{Cluster-Level Explainability}
\label{sec:xai_experiments}
We next examine whether cluster assignment changes the quality of explanations produced by the resulting local models. Evaluating saliency maps remains difficult because no agreed quantitative metric exists \cite{b43}. Following \cite{b44} (partly building on \cite{b45}) we use the insertion and deletion metrics: removal of the most important pixels should sharply reduce classifier confidence, so a low deletion AUC indicates a good explanation, while adding the most important pixels should sharply increase confidence, so a high insertion AUC indicates a good explanation. The add or remove process proceeds in $\lceil (W\times H)/s\rceil$ steps (with $W,H$ the image dimensions and $s$ the step size), and class probabilities are recorded at each step to compute the AUC.
We are less interested in comparing individual explainers than in the overall behaviour of explanations across clusters, scenarios and algorithms, so we built an aggregation framework. For each simulation we collect insertion and deletion AUCs per cluster, computing them across batches and averaging afterwards.\footnote{Identical to the implementation of \cite{b44}, available at \url{https://github.com/eclique/RISE/blob/master/evaluation.py}} We then summarise each cluster by mean, variance and quartiles, which yields a run-level summary independent of the number of clusters. Intuitively, clusters whose hypothesis functions are easy to interpret produce high average insertion AUC and low average deletion AUC, while even a single poorly interpretable cluster can degrade the overall score.
We show the procedure in Algorithm \ref{algo:INDE}: fix a scenario and an iteration (typically the last saved), form for each client an in sample distribution (test sets of clients in the same cluster), an out of sample distribution (test sets of clients outside the cluster) and an orchestrator distribution (samples from the orchestrator test set), sample examples, generate explanations, compute insertion and deletion scores, then aggregate for each cluster and visualise via box plots (Subsection \ref{xai:boxplots}) or perform Critical Difference analysis (Subsection \ref{xai:CD}). Sample sizes were 350 for MNIST, FMNIST, CIFAR10 and PathMNIST, and 180 for BloodMNIST.

\begin{algorithm}
    \caption{Insertion-Deletion Framework for \ocfl}
    \label{algo:INDE}
    \DontPrintSemicolon
    \SetKw{Input}{input}
    \SetKw{Return}{return}
    \Input{$\mathcal{E}(\cdot)$ (Explainer), $T \in \mathbb{N}$ (Evaluation Round), $\zeta \in \mathbb{R} \quad \zeta \leq 1$ (Sample Size), $\mathbf{I} \in \{InDistribution, OutOfDistribution, Orchestrator\}$  (Evaluation Mode)} \;
    $C_i^{INDE} \longleftarrow \{ \} \quad \forall C_i \in C$ \;
    \For{$C_i \in C$}{
        $\hat{h}_i = \frac{1}{|C_i|} \sum_{j \in C_i}\theta_j^{T,k}$ \;
        \If{$\mathbf{I} = \textit{InDistribution}$}{
            $\hat{D} = \{D^{test}_k \in D | k \in C_i\}$\;
        }
        \If{$\mathbf{I} = \textit{OutOfDistribution}$}{
            $\hat{D} = \{D^{test}_k \in D | k \notin C_i\}$\;
        }
        \If{$\mathbf{I} = \textit{Orchestrator}$}{
            $\hat{D} = D_{orchestrator}$\;
        }
        $\hat{D} \longleftarrow Sample(\hat{D}, \zeta)$ \;
        $e$ = AUC($\mathcal{E}(\hat{h}_i(x_i, y_i))$) \;
        $C_i^{INDE} = C_i^{INDE} \cup e$ \;
    }
    \Return{$C_1, C_2, \cdots C_n$}
\end{algorithm}

\subsection{Qualitative Analysis of Saliency Maps}
\label{xai:qualitative}

Visual inspection of the saliency maps generated by the local hypothesis functions on the dominant class (the class that the clustering should personalise towards) is the first step of the analysis. One of the undeniable advantages of the saliency-based explainability method (as presented by \cite{b49}) is the ability to plot and compare them easily against each other. This can serve as a preliminary check of the personalisation impact on the explainability.

For this experiment, we selected three clients for each dataset (clients 1, 7 and 12). The majority class is sampled from local test sets of each client, and an explanation is generated using hypothesis functions personalised by various clustering algorithms. Those results are then plotted for each sampled data point individually.
We provide examples of such a comparison in the Appendix Section~\ref{app:saliency_maps}.

The qualitative analysis of the saliency maps delivers a few valuable observations. Firstly, the saliency maps generated by models personalised by OCFL-KM, OCFL-MS or OCFL-HDBS tend to generate fewer \textit{artefacts} - unexplainable highlights that seem not to contain any substantial part of the image relevant for classification. They also tend to be more cohesive and focused on either the object itself or the direct surroundings of it, which may help the classifier to differentiate between the classes. The second observation concerns the relationship between the generated saliency map and a predicted class. There are examples of cases where the generated saliency maps were barely distinguishable from each other, yet the predicted class was different. In that regard, OCFL-KM, OCFL-MS and OCFL-HDBS tend to predict the correct class more frequently. As pointed out earlier, the usability of qualitative analysis of individual saliency maps is rather limited in its nature (as explained by \cite{b43}, \cite{b44} and \cite{b49}), and this display is made predominantly for demonstration purposes only. The next two subsections deal with a more formal analysis of the generated explanations.

\subsection{Inter-Cluster Average AUC for Insertion and Deletion}
\label{xai:boxplots}

\begin{figure}
    \centering
    \includegraphics[width=1\linewidth]{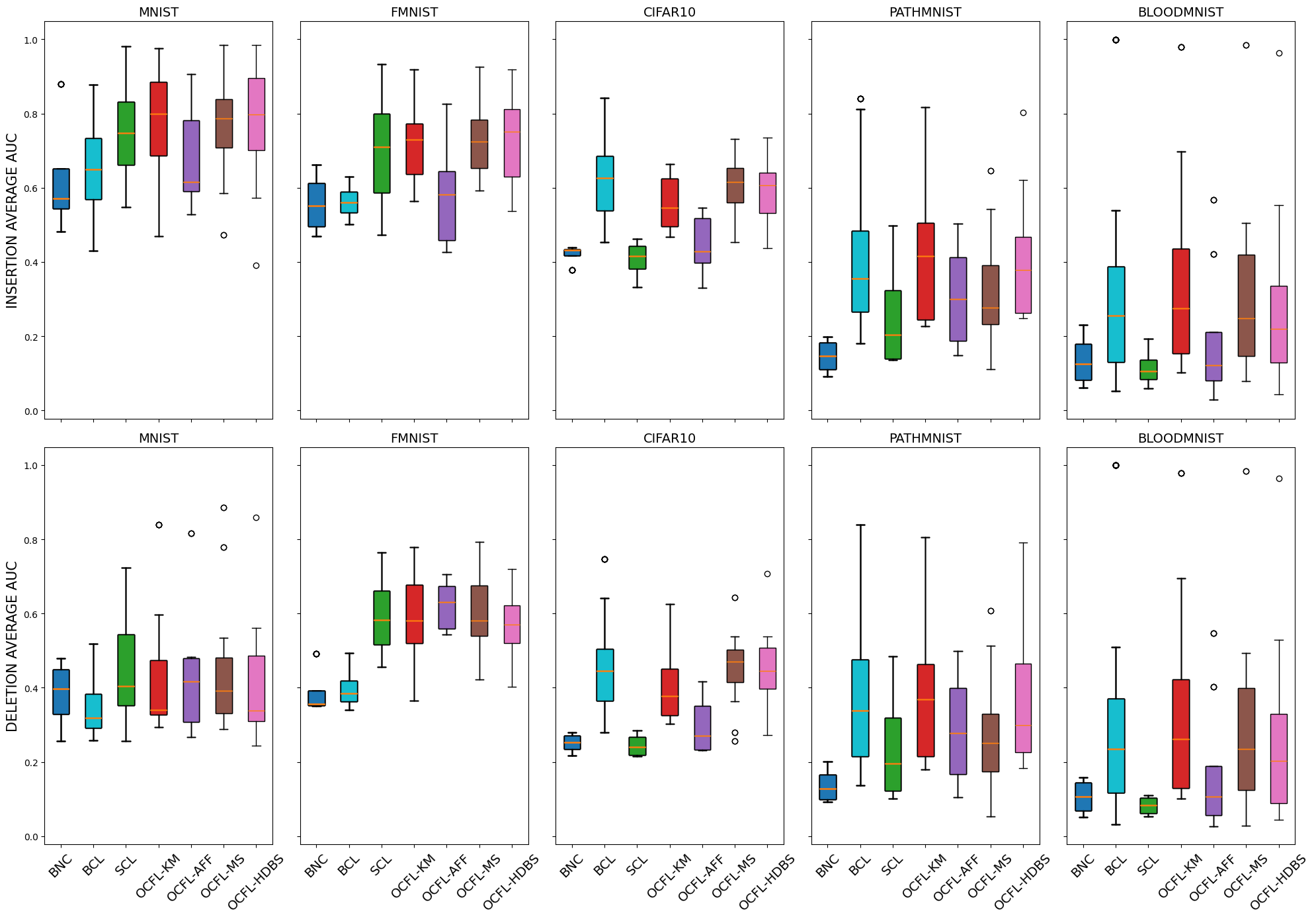}
    \Description{Grid of box plots of per-cluster mean insertion AUC (top row) and deletion AUC (bottom row) on in-cluster samples, one box per clustering method and the baseline, one column per dataset.}
    \caption{Boxplots for average insertion area under the curve (AUC) registered for particular clusters for \textbf{in-cluster distributions}. Each dataset is plotted as a separate column, and average inter-cluster insertion is placed in the first row, while average inter-cluster deletion is placed in the second row. One figure contains box plots of all six compared clustering algorithms and one additional plot for a baseline.}
    \label{fig:boxplot_1A}
\end{figure}

\begin{figure}
    \centering
    \includegraphics[width=1\linewidth]{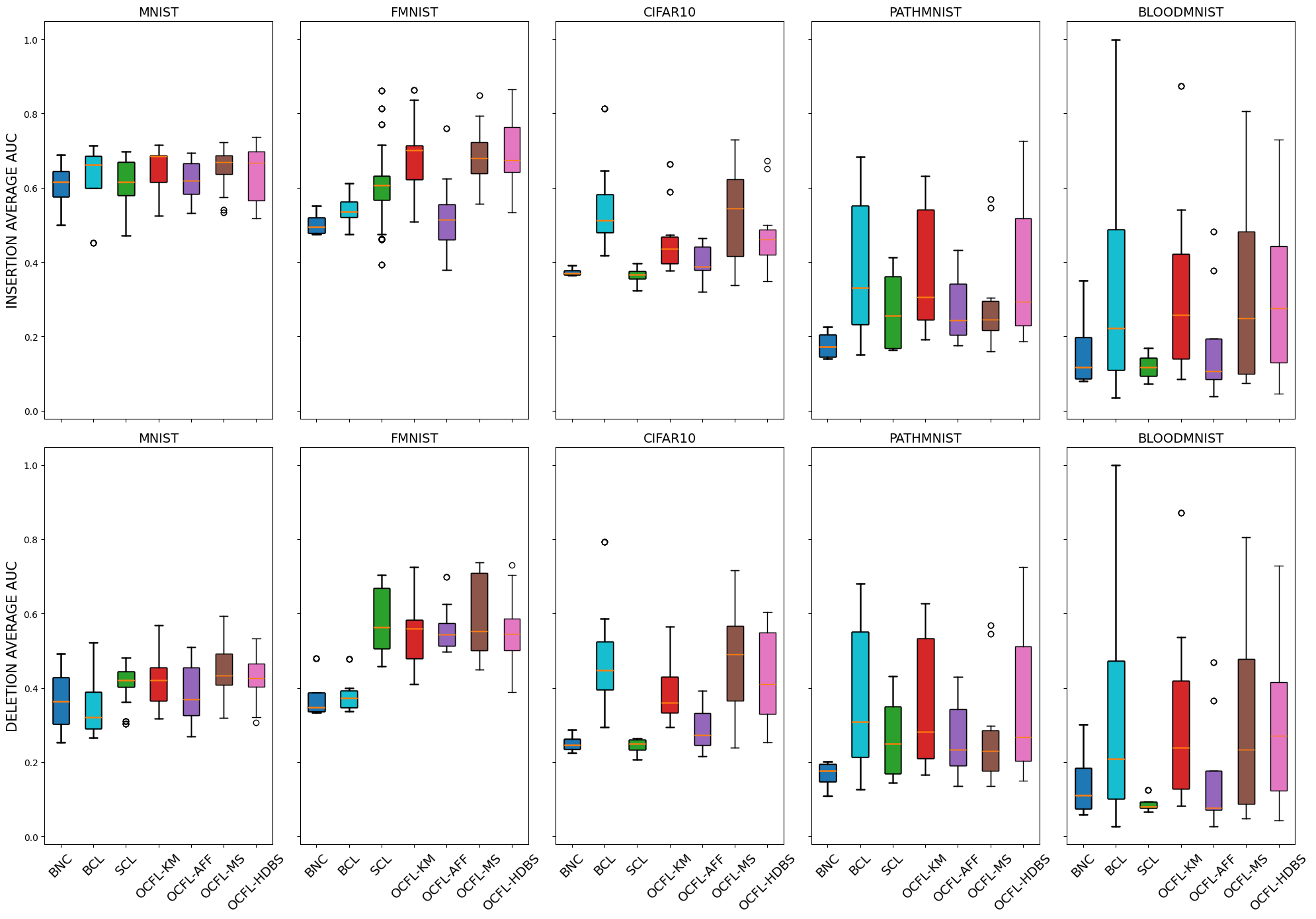}
    \Description{Grid of box plots of per-cluster mean insertion AUC (top row) and deletion AUC (bottom row) on orchestrator samples, one box per clustering method and the baseline, one column per dataset.}
    \caption{Boxplots for average insertion area under the curve (AUC) registered for particular clusters for \textbf{orchestrator distribution}. Each dataset is plotted as a separate column, and an average inter-cluster insertion is placed in the first row, while an average inter-cluster deletion is placed in the second row. One figure contains box plots of all six compared clustering algorithms and one additional plot for a baseline.}
    \label{fig:boxplot_1B}
\end{figure}

Algorithm \ref{algo:INDE} computes the average insertion and deletion scores for all clusters on either in-cluster, out-of-cluster or orchestrator distributions. Hence, for each particular dataset, data split, and algorithm, it is possible to obtain the average AUC for insertion-deletion and assess the variability of the AUC between the clusters. We perform those experiments in the following manner. Firstly, using Algorithm \ref{algo:INDE}, we calculate (in batches) the average AUC for insertion and deletion across the batch.\footnote{The same method was used by \cite{b44} to calculate average AUC across the batches of data to speed up the process. This method was adopted directly from the repository of that paper, available at the following link: \url{https://github.com/eclique/RISE/tree/master}} Afterwards, we average those scores, so that a single statistic represents each cluster. Given that we are interested in inter-cluster differences, we employ boxplots to assess the variability of those statistics across different splits within one dataset. The final results are plotted in Figure \ref{fig:boxplot_1A} for samples drawn from the in-cluster distribution and in Figure \ref{fig:boxplot_1B} for data points drawn from the orchestrator distribution.

\subsection{Critical Differences Analysis}
\label{xai:CD}

Critical Differences analysis (CD) is a more rigorous approach to exploring the quantitative differences in the quality of explanations generated by a particular hypothesis function. The method of CD diagrams was first introduced by \cite{b50} to compare outcomes of multiple different learning algorithms over multiple different datasets. The test is formally divided into two stages. Firstly, a Friedman test is performed to reject (or fail to reject) a hypothesis that there exist any significant differences in the tested methods. Then, multiple hypothesis testing is performed (for example, with adjusted Wilcoxon test or with Bonferroni's method) for each group of methods. Subsequently, the grouped methods are ranked and placed on the x-axis. The groups that could not be statistically deemed as different are linked by a horizontal crossbar (as explained by \cite{b50} or \cite{b51}).

In the case of this study, the CD method allows one to compare the insertion and deletion scores of various algorithms over multiple (repeated) test runs. In order to do that, an average across all clusters is calculated.\footnote{As an example, assume that INDE scores were calculated for each cluster $C_i \in C$ following Algorithm \ref{algo:INDE}. Afterwards, a mean value for each of those clusters is calculated, \ie $\hat{C_i} = \frac{1}{|C_i|}\sum_{e\in C_i}AUC(e)$. As the last step, those values are averaged across the clusters, so each run can be represented by a single statistic.} The results for the \textit{in-distribution} explanations are reported in Figure \ref{fig:INDE_CD_experiment_A}.
Figures for experiments performed on other distributions are reported in Appendix~\ref{app:cd_plots}.

\begin{figure}[h!]
\centering
\begin{subfigure}[b]{1\textwidth}
    \centering
    \includegraphics[width=0.7\linewidth]{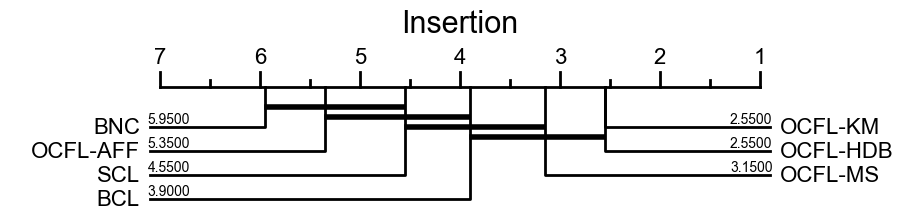}
    \Description{Critical difference diagram ranking the methods by insertion AUC on in-distribution samples.}
    \caption{Insertion}
    \label{fig:INDE_CD_insertion_A}
\end{subfigure}
\begin{subfigure}[b]{1\textwidth}
    \centering
    \includegraphics[width=0.7\linewidth]{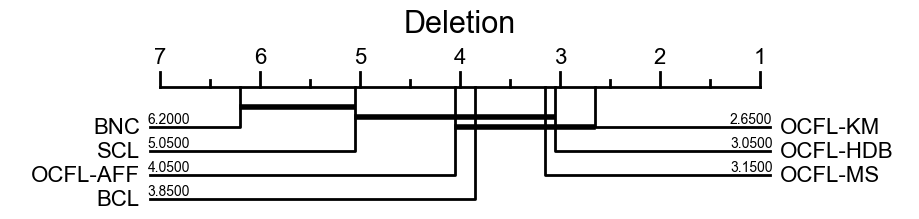}
    \Description{Critical difference diagram ranking the methods by deletion AUC on in-distribution samples.}
    \caption{Deletion}
    \label{fig:INDE_CD_deletion_A}
\end{subfigure}
\caption{Critical Difference Plots for insertion and deletion. Particular algorithms are ranked and plotted on the x-axis according to the mean AUC value of insertion (panel a) and deletion (panel b). The bold black bar indicates that no statistical differences were detected. The plots present results obtained on the in-distribution sample for the cluster hypothesis function as explained in Algorithm \ref{algo:INDE}.}
\label{fig:INDE_CD_experiment_A}
\end{figure}

\section{Scope and limitations}
\label{sec:considerations}

\paragraph{Types of Data Heterogeneity}
\label{limitations:heterogeneity}
The experiments vary only the label supports and label proportions across clusters, with a class-conditional feature distribution shared by all clients. A controlled evaluation of feature distribution skew and quality skew, as well as of concept drift and concept shift, is left for future work.

\paragraph{Privacy Issues}
\label{limitations:privacy}
\fl was proven to reduce the severity of unintended memorisation, thus reducing the potential data leakage from aggregated models, as proven by \cite{b29}. However, the baseline version of \fl is still vulnerable to privacy breaches, such as Membership Inference Attacks (MIA, see \cite{b30}), Reconstruction Attacks (RA, see \cite{b31}), and Cross-Silo Mitigation (CSM, see \cite{b41}). There are several strategies to mitigate those vulnerabilities, including Differential Privacy (DP), which is well studied in the context of the \fl (see studies conducted by \cite{b32}). It must be noted that the privacy-related issues were not a primary concern of this work, as we have focused our studies on the personalisation issues and explanation of the knowledge transfer with an additional usage of saliency maps. However, the subsequent studies should focus on that particular problem, addressing issues such as the impact of personalisation on the susceptibility to privacy risks and the cluster-level privacy budgets.

Undoubtedly, the baseline \fl offers stronger privacy guarantees by the very fact that the data is kept on the side of local devices. However, those guarantees may not be sufficient, given the nature of the data. It must be noted that our algorithm naturally extends to a majority of privacy-enhancing techniques such as DP, as we do not execute any additional process on the client side nor aggregate any information other than gradients. Although the performance is expected to drop, DP can be easily applied to our method.

On the final note, there is also a trade-off between the personalisation degree and the attack vulnerability. It was proven by \cite{b33} that well-generalised models are less vulnerable when it comes to certain types of privacy attacks. One of the advantages of proper clustering is the ability to create a correct clustering structure without creating a situation where some clients are singled out to a separate cluster.

\paragraph{Dynamic Client Environment}
\label{limitations:dynamic_client}
The \fl paradigm was designed to be dynamic, with clients joining and dropping out of the training at random intervals. In this paper, we have focused on a stable environment where all the clients are sampled each round. Since we use three different benchmark lines (baseline, \cite{b7} and \cite{b1}), we wanted to keep the environment stable, so as to receive comparable and unbiased results. However, our algorithm naturally extends to a dynamic client environment, where One-Shot Clustering is performed on only a subset of clients. Depending on the chosen clustering algorithm, the subsequent client attribution will differ, but in most cases, it will rely on attributing each client to the most appropriate cluster by comparing the received gradients with the given clustering structure.  Depending on the ratio of the sample size to the population size, One-Shot Clustering can be turned into Few-Shot Clustering, where the clustering is performed a few times, each time updating the centroids or other appropriate clustering structure.
Similarly to the works of \cite{b1} and \cite{b7} that we use as a comparison, we started from the stable federated scenario to briefly extend it into a dynamic environment. However, we wanted to mention the issue of dynamic clustering in this work and set it as a new direction for the research.

\paragraph{Combining Temperature Function with Moving Average}
\label{limitations:temperature_extension}
The presented Algorithm \ref{algo:ocfl} assumes performing clustering at the first sign of temperature rise (line~\ref{l:trigger} of Algorithm \ref{algo:ocfl}). Such a use of the Temperature Function was defined at the beginning of our experiments, and given the positive outcome of the experiments, we have decided to present it in this paper. However, a note must be made that it is possible to observe a moving average of the Temperature Function, possibly postponing the clustering in time until we record the divergence of federated gradients is observed throughout several rounds. In such a case, Algorithm \ref{algo:ocfl} will constitute a special case of the presented algorithm, with a time window equal to one. For testing the efficiency of this and other similar methods, we have specially implemented and attached a small prototypical library to this paper, which should allow the reader to experiment with different methods of detecting a suitable moment for performing clustering.

\section{Conclusions}
\label{sec:conclusions}
In this paper, we have presented a family of \ocfl clustering-agnostic algorithms that are suitable for performing the split of the population at an early stage in an automatic manner. Moreover, by an extensive empirical evaluation over five different datasets, each one in eight different configurations, we have evidenced the efficiency of the proposed solution, especially when combined with density-based clustering (i.e., OCFL-MS or OCFL-HDBS). Proposed algorithms are able to correctly identify the structure of data-generating distributions and attribute clients to an appropriate cluster as early as the first three rounds of the \fl process. This has a direct impact on the personalisation of local models, as evidenced by the empirical results. 

Furthermore, by using saliency maps to inspect local explanations generated by personalised clusters, we were not only able to empirically prove that our solution can deliver better personalisation, but also that the local hypothesis functions are thus able to generate more meaningful explanations on the provided data samples. The effect of the cluster assignment on local explanation quality in \cfl remains largely unexplored; building on earlier explainability-based and application-specific work \cite{b54, b55}, ours is -- to the best of our knowledge -- the first systematic, cluster-level evaluation of this relationship. Due to this, we have not only proven some additional empirical properties of the presented algorithm, but also proposed a few methodological frameworks which can be used to further develop the intersection of those two domains.

\section*{Disclosure of AI Assistance}
The authors of this work used AI-assisted tools for proof-reading, minor language correction, and formatting suggestions. Those tools were not used to generate scientific content, results, or claims. The code placed in the shared repository was written solely by the authors and reviewed accordingly. Whenever an AI system suggested a minor language or formatting change, the suggestion was checked beforehand.
\begin{acks}
This work was supported by the European Union in projects \textbf{LeADS} (GA no. 956562) and \textbf{CREXDATA} (GA no. 101092749). This work has also been supported by the PNRR-M4C2-Investimento 1.3, Partenariato Esteso PE00000013-“FAIR-Future Artificial Intelligence Research”-Spoke 1 “Human-centered AI” funded by the European Commission under the NextGeneration EU programme. This work was also funded by NextGenerationEU – National Recovery and Resilience Plan (Piano Nazionale di Ripresa e Resilienza, PNRR) – Project: “SoBigData.it – Strengthening the Italian RI for Social Mining and Big Data Analytics” – Prot. IR0000013 – Avviso n. 3264 del 28/12/2021. This work was also funded by the European Union under Grant Agreement no. 101120763 - TANGO. Additionally, Maciej Krzysztof Zuziak was supported by the Max Weber Fellowship awarded by the European University Institute (EUI). Views and opinions expressed are, however, those of the author(s) only and do not necessarily reflect those of the European Union or the European Health and Digital Executive Agency (HaDEA). Neither the European Union nor the granting authority can be held responsible for them.
\end{acks}

\bibliographystyle{ACM-Reference-Format}
\bibliography{sample-base}
\newpage
\input{appendix}

\end{document}

%% file: macro.tex
\usepackage{xspace}
\usepackage{soul}
\usepackage{acronym}

\newcommand{\ie}{\textit{i.e.},\xspace}

\acrodef{ml}[ML]{Machine Learning}
\newcommand{\ml}{\ac{ml}\xspace}

\acrodef{fl}[FL]{Federated Learning}
\newcommand{\fl}{\ac{fl}\xspace}

\acrodef{fedavg}[FedAvg]{Federated Averaging}
\newcommand{\fedavg}{\ac{fedavg}\xspace}

\acrodef{fedopt}[FedOpt]{Adaptive Federated Optimisation}
\newcommand{\fedopt}{\ac{fedopt}\xspace}

\acrodef{cfl}[CFL]{Clustered Federated Learning}
\newcommand{\cfl}{\ac{cfl}\xspace}

\acrodef{ocfl}[OCFL]{One-Shot Clustered Federated Learning}
\newcommand{\ocfl}{\ac{ocfl}\xspace}

\acrodef{iot}[IoT]{Internet-of-Things}
\newcommand{\iot}{\ac{iot}\xspace}

\acrodef{dp}[DP]{Differential Privacy}

\acrodef{he}[HE]{Homomorphic Encryption}

\acrodef{smpc}[SMPC]{Secure Multi-Party Computation}

%% file: appendix.tex
\appendix
\section{Appendix}

\subsection{Proof of the Perturbation Lemma}
\label{appendix:lemmaperturb}
Here we report the proof for the Perturbation Lemma \ref{lemma:perturb}.

\begin{proof}
Let $\varphi$ be the angle between $x$ and $a$. Then
\begin{equation}
\lVert x-a\rVert^2
= \bigl(\lVert x\rVert-\lVert a\rVert\bigr)^2
+ 2\lVert x\rVert\lVert a\rVert(1-\cos\varphi)
\geq 2\lVert x\rVert\lVert a\rVert(1-\cos\varphi).
\end{equation}

Moreover, $\lVert x\rVert \geq \lVert a\rVert-\varepsilon \geq \lVert a\rVert/2$. Therefore

\begin{equation}
2(1-\cos\varphi)
\leq \frac{\lVert x-a\rVert^2}{\lVert x\rVert\lVert a\rVert}
\leq 2\frac{\varepsilon^2}{\lVert a\rVert^2}.
\end{equation}

Since
\begin{equation}
\left\lVert
\frac{x}{\lVert x\rVert}
- \frac{a}{\lVert a\rVert}
\right\rVert^2
= 2(1-\cos\varphi),
\end{equation}

we obtain the stated bound, with a slightly conservative constant.
\end{proof}

\subsection{Proof of Theorem~\ref{th:cosine-distance}}
\label{appendix:cosine-distance}
Here we report the proof for Theorem~\ref{th:cosine-distance}.
\begin{proof}
Define the normalised observed and population directions

\begin{equation}
u_i = \frac{\Delta_i^t}{\lVert\Delta_i^t\rVert},
\qquad
v_c = \frac{\mu_c^t}{\lVert\mu_c^t\rVert}.
\end{equation}

By the perturbation lemma, $\lVert u_i-v_{z_i}\rVert \leq 2\rho_t. $ If $z_i=z_j=c$, then

\begin{equation}
\lVert u_i-u_j\rVert
\leq \lVert u_i-v_c\rVert + \lVert u_j-v_c\rVert
\leq 4\rho_t.
\end{equation}

For unit vectors $a,b$, $d_{\cos}(a,b) = 1-\langle a,b\rangle = \frac{1}{2}\lVert a-b\rVert^2.$ Hence
$ d_{\cos}(\Delta_i^t,\Delta_j^t) = \frac{1}{2}\lVert u_i-u_j\rVert^2 \leq 8\rho_t^2.$ Now suppose $z_i=c$ and $z_j=d$ with $c\neq d$. Since
\begin{equation}
\begin{aligned}
\bigl\lvert \langle u_i,u_j\rangle-\langle v_c,v_d\rangle\bigr\rvert
&\leq \bigl\lvert\langle u_i-v_c,u_j\rangle\bigr\rvert
+ \bigl\lvert\langle v_c,u_j-v_d\rangle\bigr\rvert\\
&\leq 4\rho_t,
\end{aligned}
\end{equation}

we have

\begin{equation}
\begin{aligned}
d_{\cos}(\Delta_i^t,\Delta_j^t)
&= 1-\langle u_i,u_j\rangle\\
&\geq 1-\langle v_c,v_d\rangle-4\rho_t\\
&= d_{\cos}(\mu_c^t,\mu_d^t)-4\rho_t\\
&\geq B_t-4\rho_t.
\end{aligned}
\end{equation}

If $B_t>4\rho_t+8\rho_t^2$, then the lower bound on between-cluster distances exceeds the upper bound on within-cluster distances. This is precisely block separation.
\end{proof}

\FloatBarrier
\subsection{Data splits}
\label{appendix:datasplits}
\begin{figure}
    \centering
    \includegraphics[width=0.7\linewidth]{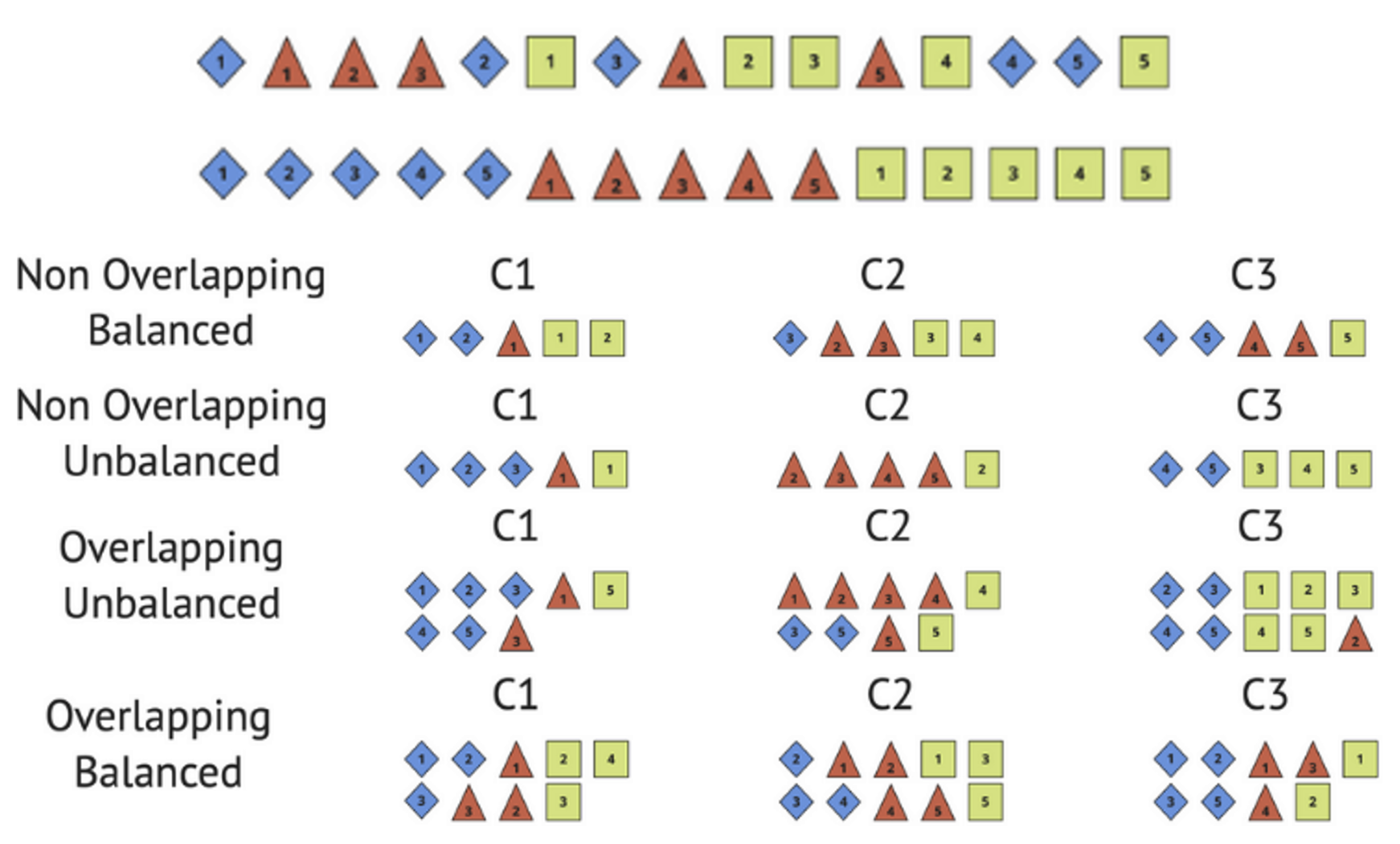}
    \Description{Schematic of the four data-split types; shapes denote classes and integers denote local identifiers, with one column per cluster.}
    \caption{Visualisation of different data distributions in \cfl. The different shapes represent different classes, while the integers imprinted into them serve as local identifiers. Four different splits are presented in each row, where each individual column is a separate cluster. As indicated by the individual numbers displayed on the shapes, some particular samples may be shared between the clusters.}
    \label{fig:data_splits_visualization}
\end{figure}

Here we report the details of how data is split in our experiments. The \textbf{non-overlapping balanced splits} resemble the most basic case, where the cluster has access to separate classes, the agents are uniformly distributed across the clusters (with the same number of agents belonging to each cluster), and the available labels are uniformly distributed across the clients. The \textbf{overlapping balanced split} implies similar circumstances but with a partial overlap of classes, i.e., the clients partially share some of the classes. Imbalance is simulated across two different levels simultaneously. Firstly, clients are unevenly distributed across clusters. For 15 clients, the first cluster holds $20\%$ of the clients, while the second and third hold approximately $47\%$ and $33\%$, respectively (3/7/5); for 30 clients, the proportions are approximately $23\%$, $47\%$ and $30\%$ (7/14/9). Secondly, the prior distribution across labels is not uniform but is sampled according to a weight vector $\sim Dir(\overline{v})$, where $\overline{v}$ is defined as before. \textbf{The non-overlapping imbalanced split} assumes that the data is distributed according to $Dir(\overline{v})$, but the set of classes is still disjoint across the clusters. A practical visualisation of this process is shown in Figure \ref{fig:data_splits_visualization}.

\FloatBarrier
\subsection{Hyperparameter Settings}
The Table \ref{tab:general_hyperparameters} presents a hyperparameter choice employed for solving classification tasks on particular datasets. In every case, the number of local epochs was always equal to $3$ and the general Learning Rate for the FedOpt algorithm was equal to $1.0$. Betas are defined as coefficients used for computing running averages of gradient and its square\footnote{PyTorch documentation available under the following link: \url{https://docs.pytorch.org/docs/stable/generated/torch.optim.Adam.html}} and are applicable only in the case of Adam optimiser. EPS is defined as the term added to the denominator to improve numerical stability\footnote{PyTorch documentation available under the following link: \url{https://docs.pytorch.org/docs/stable/generated/torch.optim.Adam.html}} and is also defined only for Adam.

\begin{table}[h!]
    \centering
    \resizebox{\columnwidth}{!}{%
    \begin{tabular}{|c|c|c|c|c|c|c|c|}
         \hline
         DATASET & OPTIMISER & BATCH SIZE & LR & WEIGHT DECAY & BETAS & EPS & GLOBAL ITERATIONS \\
         \hline
         MNIST & SGD & 32 & 1e-2 & 0 & NA & NA & 50 \\
         \hline
         FMNIST & SGD & 32 & 1e-2 & 0 & NA & NA & 50 \\
         \hline
         CIFAR10 & SGD & 64 & 1e-2 & 0 & NA & NA & 80 \\
         \hline
         PATHMNIST & ADAM & 128 & 1e-7 & 1e-2 & (0.9, 0.999) & 1e-8 & 75 \\
         \hline
         BLOODMNIST & ADAM & 128 & 1e-6 & 1e-2 & (0.9, 0.999) & 1e-8 & 75 \\
         \hline
    \end{tabular}}
    \caption{Table presenting the hyperparameters used for solving classification tasks on particular datasets shared for all simulations presented in the paper.}
    \label{tab:general_hyperparameters}
\end{table}

\FloatBarrier
\subsection{Model-Agnostic Distributed Multitask Optimisation Fine-Tuning}
\label{app:sattler}
The following subsection overviews fine-tuning performed in relation to the Clustered Federated Learning: Model-Agnostic Distributed Multitask Optimisation Under Privacy Constraints presented in the paper of \cite{b7}. 

Since the authors' implementation included in the GitHub repository contained an additional hyperparameter - a minimal number of rounds after which clustering can be performed - we have also transferred that hyperparameter into our implementation. Given that the algorithm of \cite{b7} requires three hyperparameters (including the cool-down period that is incorporated in the code presented on GitHub), the original authors' recommendations were employed to select a set of the best hyperparameters for a fair comparison. 

As indicated in Subsection \ref{experiments:set-up}, one problem connected to this is that the authors of \cite{b7} recommend setting the values of epsilons as $\epsilon_1 = \max_t\lVert\Delta\theta^t_c\rVert / 10$ and $\epsilon_2 \in [\epsilon_1, 10\epsilon_1]$ where $\Delta\theta^t_i = SGD(\theta^{t-1}, D_i) - \theta^{t-1}$. This would require running all the simulation splits, registering the magnitude of gradients, observing the maximal possible magnitude and then choosing that value as a hyperparameter for another run, while also adjusting the cool-down round that was mentioned before. Given the number of simulations performed for the presented studies, this would yield a highly impractical workflow. Moreover, we have noticed that in the original implementation placed in the GitHub repository, the authors also do not construct their experiments in such a way, providing specified $\epsilon_1$ and $\epsilon_2$ values from the beginning. 

\begin{figure}[t]
    \centering
    \includegraphics[width=0.95\linewidth]{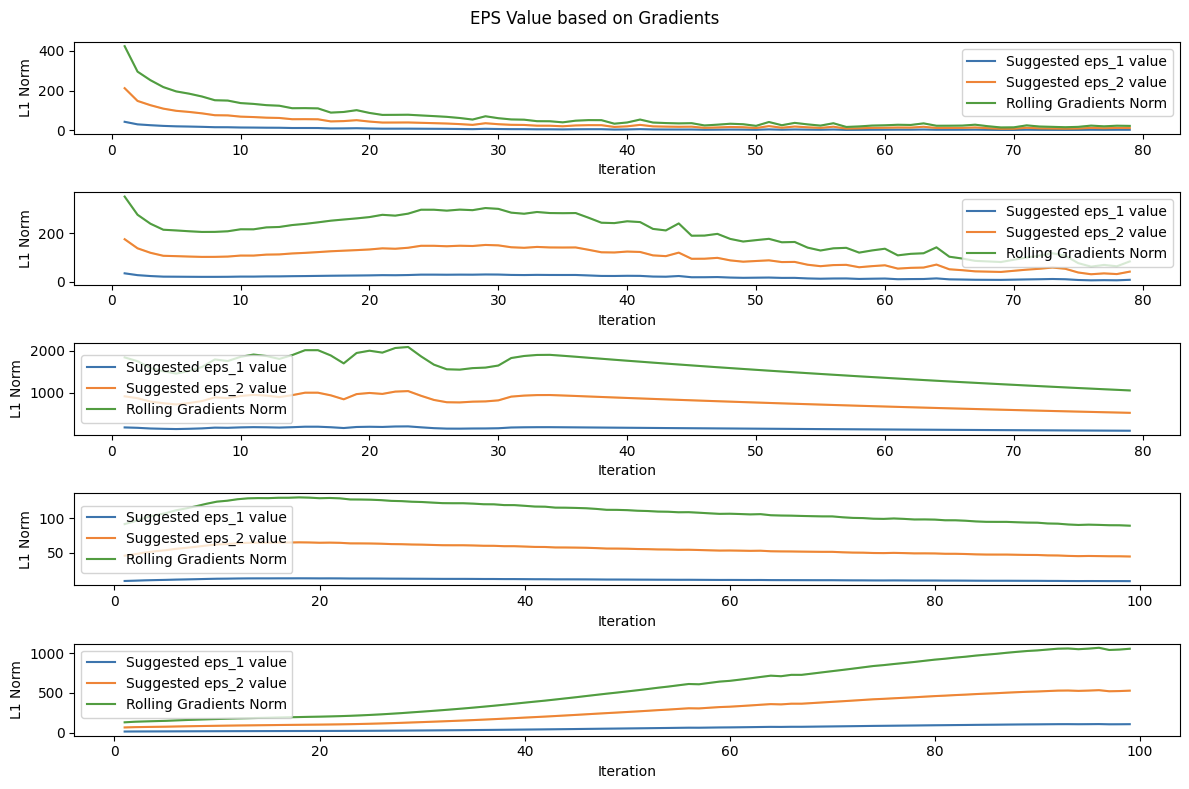}
    \Description{Line plots of the proxy update magnitude over centralised training rounds, used to choose the epsilon values and the cool-down round.}
    \caption{Presentation of the proxy value $\Delta\hat{\theta}^t = |SGD(\theta^t, D) - SGD(\theta^{t-1}, D)|$ that is used in exchange to approximate the possible $\theta$ value used in \cite{b7} as a variable for establishing the approximate values of $\epsilon_1$ and $\epsilon_2$ - two hyperparameters that are necessary for a correct clustering performed by the algorithm.}
    \label{fig:epsilon_rolling_gradient}
\end{figure}

Because of that, an approximation method was constructed for both values. We have performed a centralised training on an original dataset and defined a proxy value $\Delta\hat{\theta}^t = |SGD(\theta^t, D) - SGD(\theta^{t-1}, D)|$ that is used in exchange for the original $\theta$ value. Afterwards, the round of convergence is used to select the value for a cool-down period. The recommendations from \cite{b7} are used to select values of $\epsilon_1$ and $\epsilon_2$. The registered $\epsilon_1$ and $\epsilon_2$ values on centralised models are presented in Figure \ref{fig:epsilon_rolling_gradient}. Since no recommendations on the clustering round $c$ were provided in the original paper, we deduced that it must be somehow related to the convergence of the gradients. Hence, we used that heuristic to select an appropriate clustering round $c$.

If that failed, we turned to a heuristic search based on the code made available by \cite{b7}. In that way, it was established that one workable combination for MNIST and FMNIST datasets is $\epsilon_1 = 0.35$ and $\epsilon_2 = 1.00$. No such combination was found for CIFAR10 and PathMNIST or BloodMNIST datasets. The set of hyperparameters used specifically for each dataset is reported in Table \ref{tab:sattler_clustering_hyperparams}

\begin{table}[t]
    \centering
    \footnotesize
    \begin{tabular}{|c|c|c|c|}
         \hline
         DATASET & $\epsilon_1$ & $\epsilon_2$ & $c$ \\
         \hline
         MNIST & 0.35 &  1.00 & 30 \\
         \hline
         FMNIST & 0.35 &  1.00 & 40 \\
         \hline
         CIFAR10 & 185.00 & 925.00 & 20 \\
         \hline
         PATHMNIST & 12 & 60 & 40 \\
         \hline
         BLOODMNIST & 38 & 188 & 40 \\
         \hline
    \end{tabular}
    \caption{Set of hyperparameters used for Model-Agnostic Distributed Multitask Optimisation presented in \cite{b7}. 
    Sattler et al.'s published values: $\epsilon_1 = 6.00$ and $\epsilon_2 = 30.00$ for MNIST and $\epsilon_1 = 25.00$ and $\epsilon_2 = 125.00$ for FMNIST.}
    \label{tab:sattler_clustering_hyperparams}
\end{table}

\FloatBarrier
\clearpage
\subsection{Recorded Value of Temperature Function per Dataset and Data Split}
\label{app:temperature_individual}
\FloatBarrier

\begin{figure}[h!]
    \centering
    \includegraphics[width=0.8\linewidth]{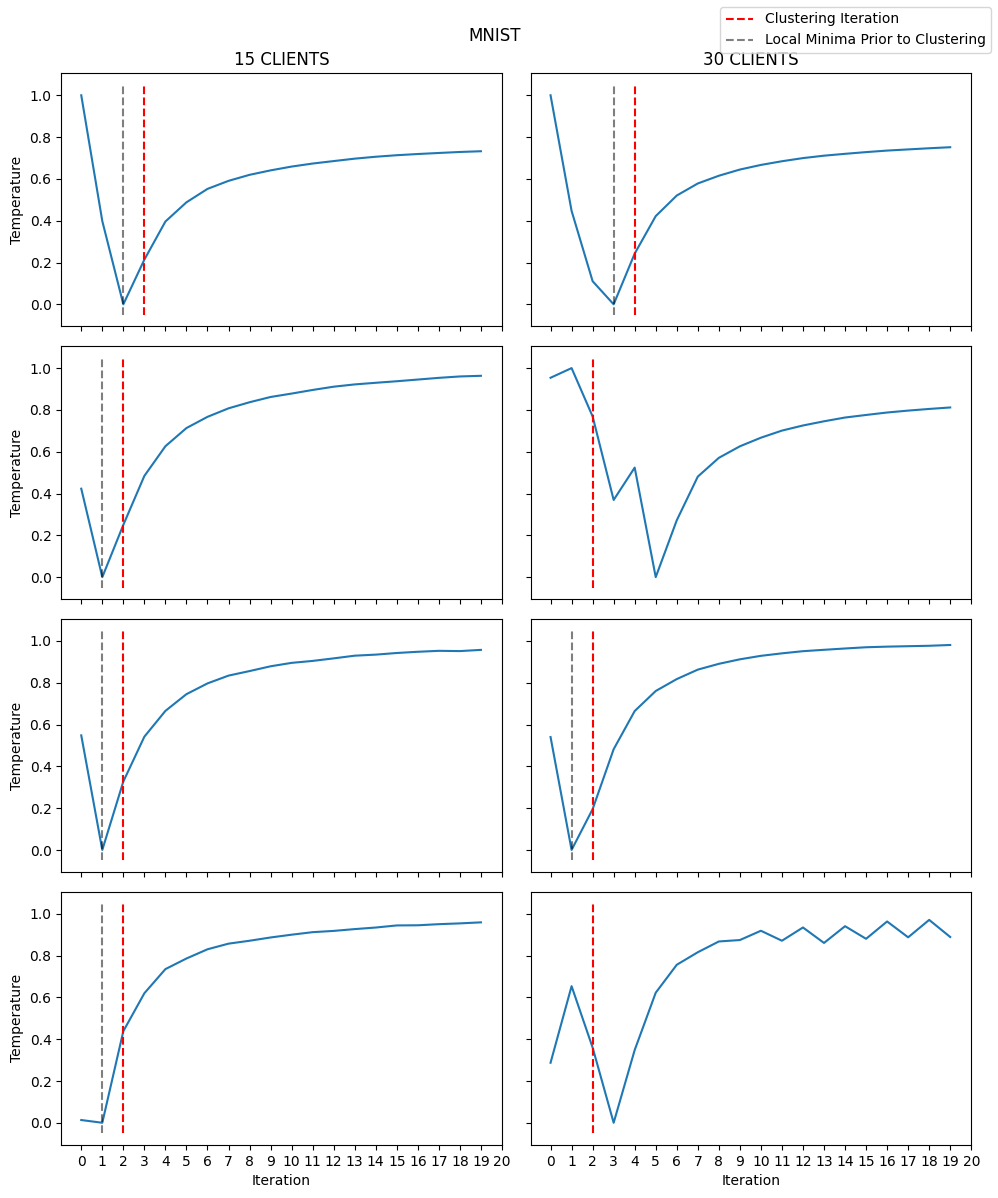}
    \Description{Grid of line plots of the normalised Temperature Function per round for each data split and client count on MNIST; a red dotted line marks the clustering round and a grey dotted line the preceding local minimum, when present.}
    \caption{Plots of the (normalised) Temperature Function defined in Definition \ref{def:Temperature_Function} for the MNIST dataset. Iterations are placed on the x-axis, while the value of the monitored function is placed on the y-axis. Columns correspond to 15 and 30 clients, and rows to the four data splits. Clustering iteration is indicated by a red dotted line, and a local minimum before clustering is indicated by a grey dotted line. Sometimes, the local minimum before clustering does not occur. In that case, only a clustering iteration is plotted on the graph.}
\end{figure}

\begin{figure}[h!]
    \centering
    \includegraphics[width=0.8\linewidth]{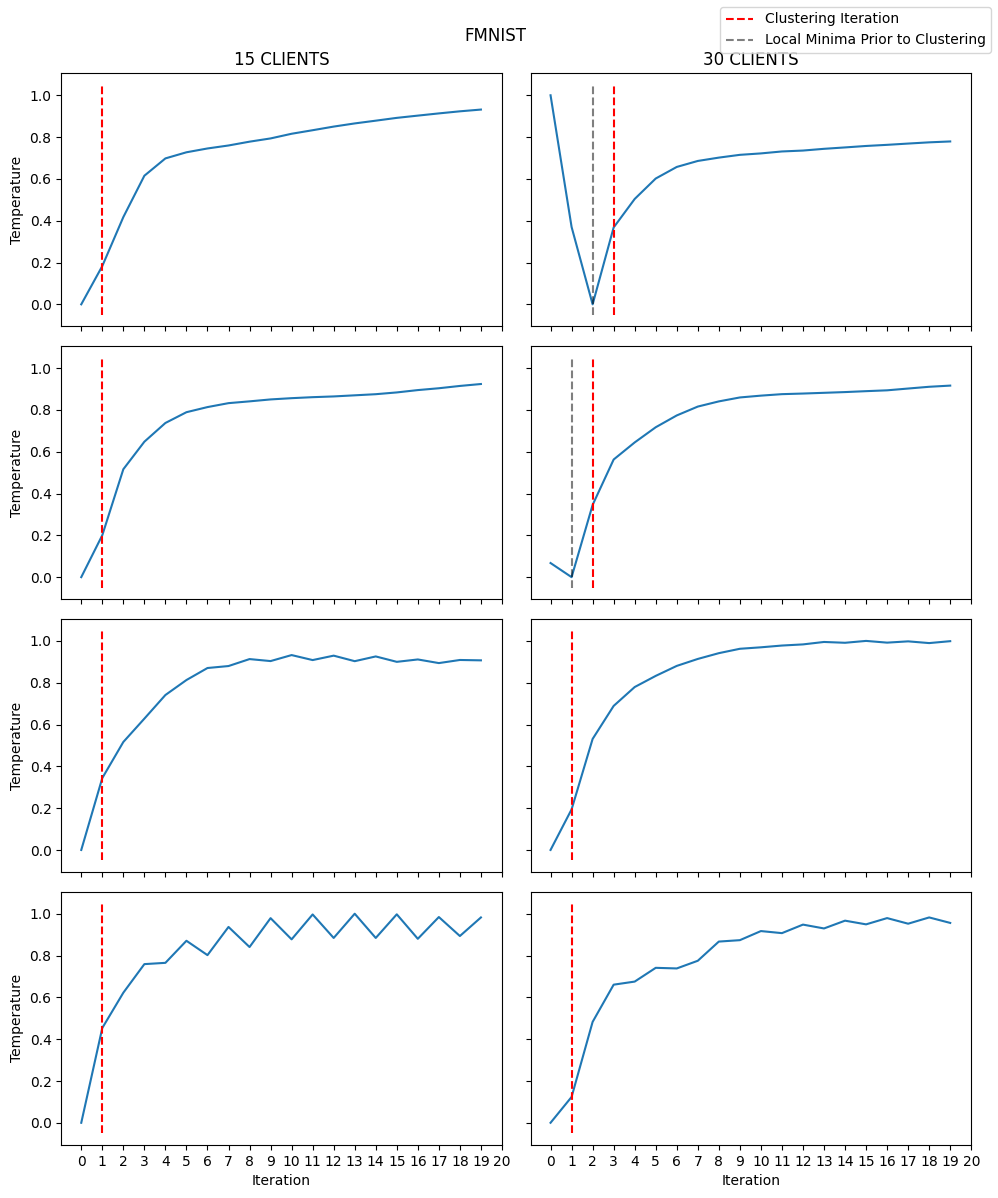}
    \Description{Grid of line plots of the normalised Temperature Function per round for each data split and client count on FMNIST; a red dotted line marks the clustering round and a grey dotted line the preceding local minimum, when present.}
    \caption{Plots of the (normalised) Temperature Function defined in Definition \ref{def:Temperature_Function} for the FMNIST dataset. Iterations are placed on the x-axis, while the value of the monitored function is placed on the y-axis. Columns correspond to 15 and 30 clients, and rows to the four data splits. Clustering iteration is indicated by a red dotted line, and a local minimum before clustering is indicated by a grey dotted line. Sometimes, the local minimum before clustering does not occur. In that case, only a clustering iteration is plotted on the graph.}
\end{figure}

\begin{figure}[h!]
    \centering
    \includegraphics[width=0.8\linewidth]{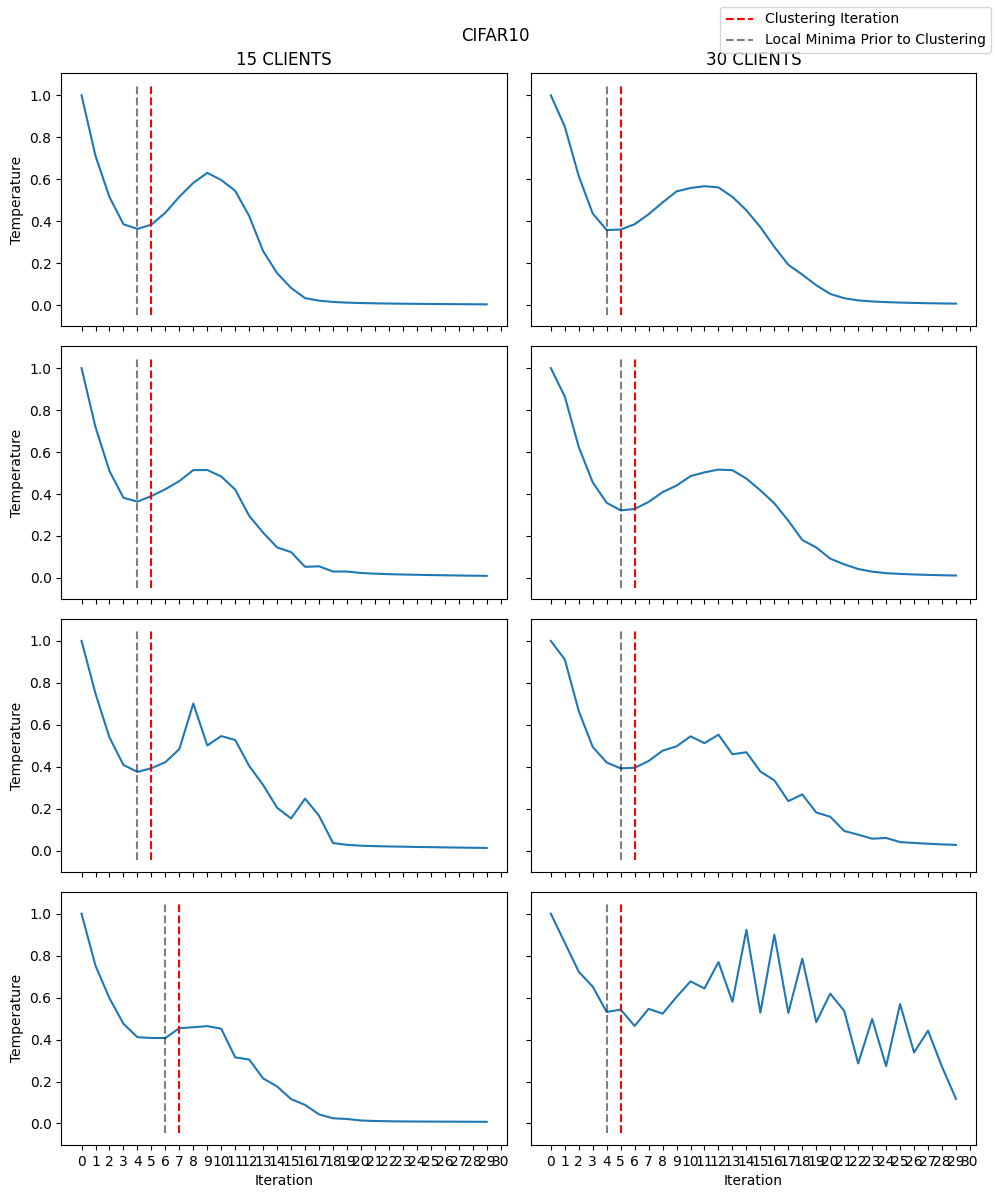}
    \Description{Grid of line plots of the normalised Temperature Function per round for each data split and client count on CIFAR10; a red dotted line marks the clustering round and a grey dotted line the preceding local minimum, when present.}
    \caption{Plots of the (normalised) Temperature Function defined in Definition \ref{def:Temperature_Function} for the CIFAR10 dataset. Iterations are placed on the x-axis, while the value of the monitored function is placed on the y-axis. Columns correspond to 15 and 30 clients, and rows to the four data splits. Clustering iteration is indicated by a red dotted line, and a local minimum before clustering is indicated by a grey dotted line. Sometimes, the local minimum before clustering does not occur. In that case, only a clustering iteration is plotted on the graph.}
\end{figure}

\begin{figure}[h!]
    \centering
    \includegraphics[width=0.8\linewidth]{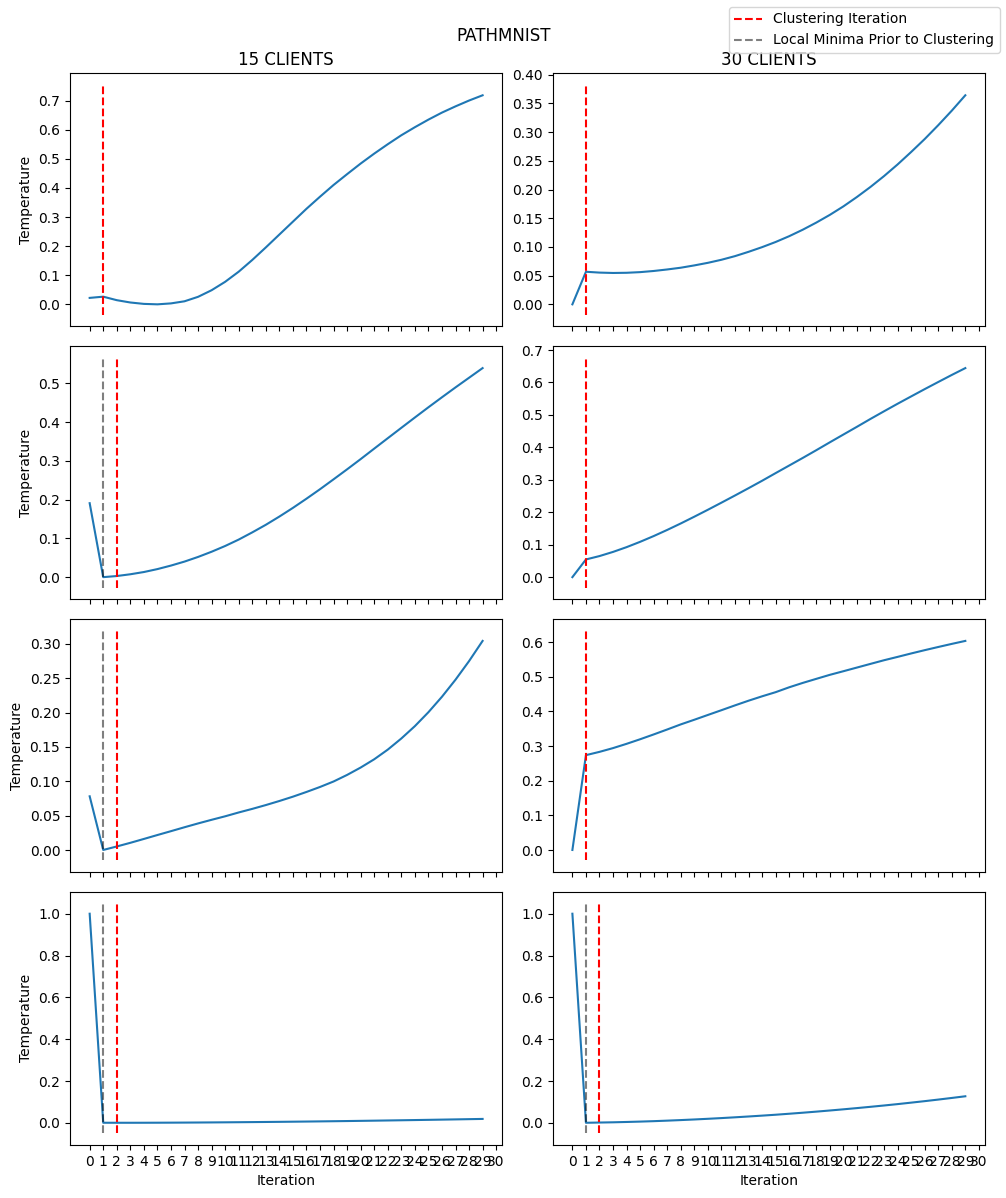}
    \Description{Grid of line plots of the normalised Temperature Function per round for each data split and client count on PathMNIST; a red dotted line marks the clustering round and a grey dotted line the preceding local minimum, when present.}
    \caption{Plots of the (normalised) Temperature Function defined in Definition \ref{def:Temperature_Function} for the PathMNIST dataset. Iterations are placed on the x-axis, while the value of the monitored function is placed on the y-axis. Columns correspond to 15 and 30 clients, and rows to the four data splits. Clustering iteration is indicated by a red dotted line, and a local minimum before clustering is indicated by a grey dotted line. Sometimes, the local minimum before clustering does not occur. In that case, only a clustering iteration is plotted on the graph.}
\end{figure}

\begin{figure}[h!]
    \centering
    \includegraphics[width=0.8\linewidth]{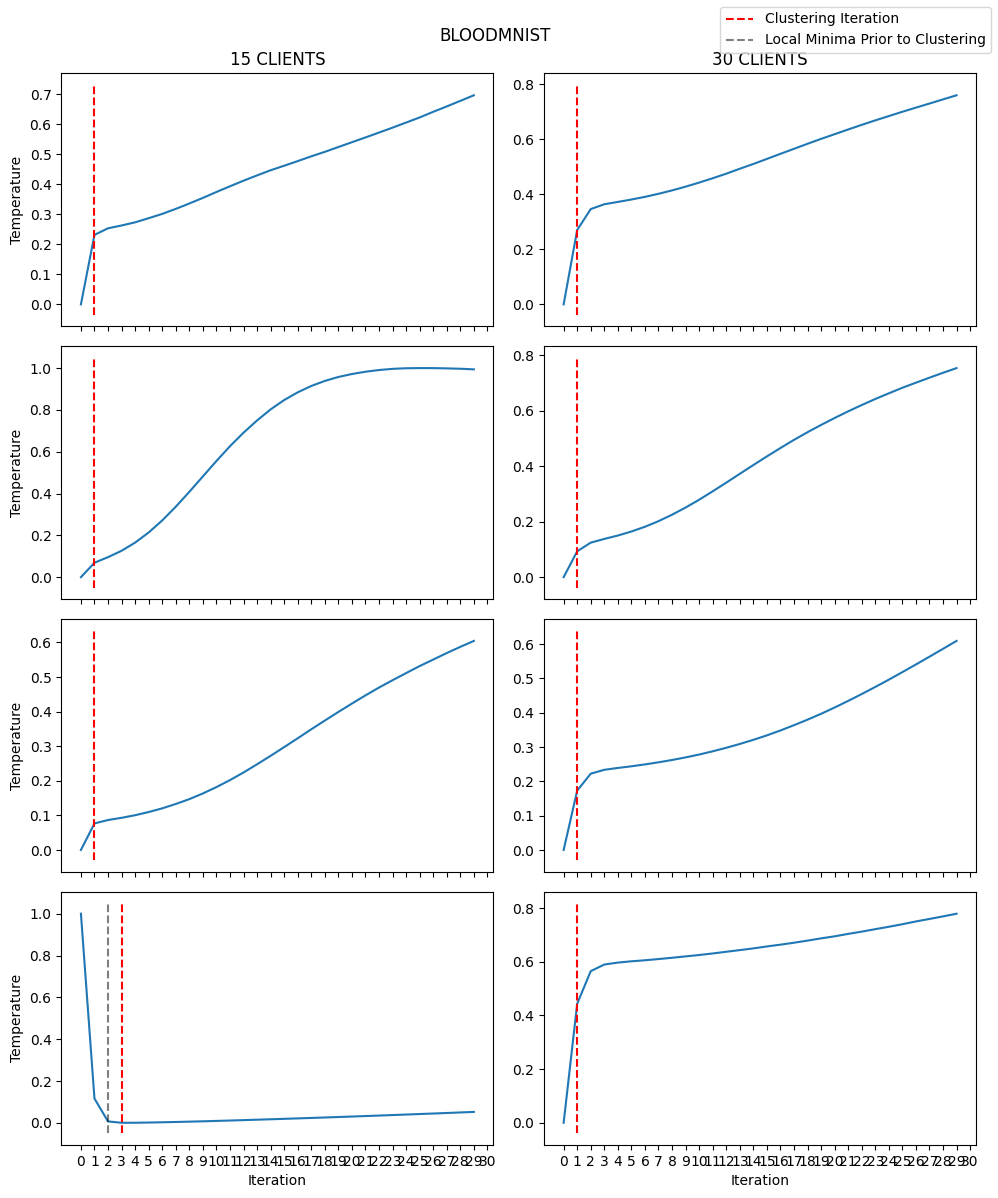}
    \Description{Grid of line plots of the normalised Temperature Function per round for each data split and client count on BloodMNIST; a red dotted line marks the clustering round and a grey dotted line the preceding local minimum, when present.}
    \caption{Plots of the (normalised) Temperature Function defined in Definition \ref{def:Temperature_Function} for the BloodMNIST dataset. Iterations are placed on the x-axis, while the value of the monitored function is placed on the y-axis. Columns correspond to 15 and 30 clients, and rows to the four data splits. Clustering iteration is indicated by a red dotted line, and a local minimum before clustering is indicated by a grey dotted line. Sometimes, the local minimum before clustering does not occur. In that case, only a clustering iteration is plotted on the graph.}
\end{figure}
\clearpage
\FloatBarrier
\subsection{Additional Temperature Function Recordings}
\label{app:temperature_function_additional}

\begin{figure}[h!]
    \centering
    \includegraphics[width=1\linewidth]{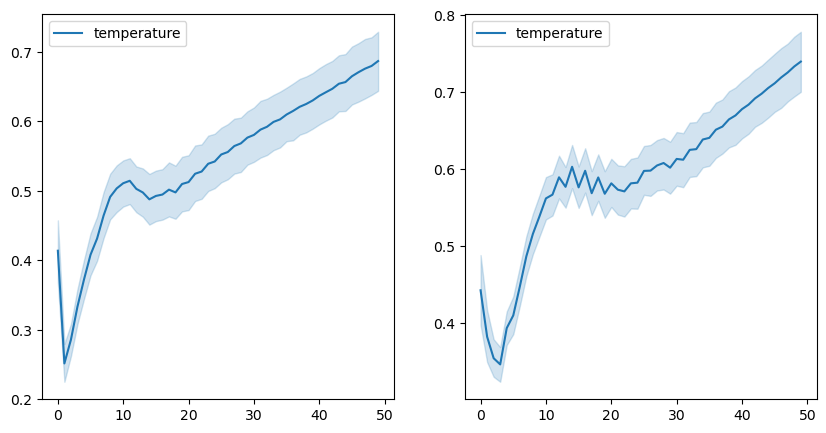}
    \Description{Two line plots of the normalised Temperature Function averaged across all datasets and splits, for 15 clients (left) and 30 clients (right), with confidence bands.}
    \caption{Behaviour of the (normalised) Temperature Function as defined in the Definition \ref{def:Temperature_Function} averaged across all the data runs and data splits. A solid blue line symbolises the mean across all the scenarios (overlapping and non-overlapping with balanced and imbalanced settings). Blue shadow represents confidence intervals established based on the variations between different scenarios and datasets. The average for 15 clients is placed on the left, while the average for 30 clients is placed on the right.}
\end{figure}

\subsection{Additional Block-Separation Margin Results}
\label{app:margin_section}
\begin{table}[t]
  \centering
  \caption{Per-scenario components of the ground-truth separation margin at
  $\tau_T$: between-DGP separation $B^\star$, within-DGP diameter $W^\star$, and margin
  $M^\star=B^\star-W^\star$, and the pair-order agreement $A^\star$, for every dataset, overlap
  setting, balance setting, and client count $|P|$. Positive values in bold.}
  \label{app:margin}
  \small
  \begin{tabular}{ll c cccc}
    \toprule
    Dataset & Split & $|P|$ & $B^\star_{\tau_T}$ & $W^\star_{\tau_T}$ & $M^\star_{\tau_T}$ & $A^\star_{\tau_T}$ \\
    \midrule
\multirow{8}{*}{MNIST} & Non-overlapping, Balanced & 15 & $8.7\!\times\!10^{-1}$ & $2.8\!\times\!10^{-1}$ & {\boldmath $+6.0\!\times\!10^{-1}$} & 1.000 \\
 & Non-overlapping, Balanced & 30 & $7.8\!\times\!10^{-1}$ & $1.6\!\times\!10^{-1}$ & {\boldmath $+6.2\!\times\!10^{-1}$} & 1.000 \\
 & Non-overlapping, Imbalanced & 15 & $8.6\!\times\!10^{-1}$ & $1.1\!\times\!10^{-1}$ & {\boldmath $+7.4\!\times\!10^{-1}$} & 1.000 \\
 & Non-overlapping, Imbalanced & 30 & $8.4\!\times\!10^{-1}$ & $4.0\!\times\!10^{-1}$ & {\boldmath $+4.4\!\times\!10^{-1}$} & 1.000 \\
 & Overlapping, Balanced & 15 & $4.7\!\times\!10^{-1}$ & $2.4\!\times\!10^{-1}$ & {\boldmath $+2.3\!\times\!10^{-1}$} & 1.000 \\
 & Overlapping, Balanced & 30 & $5.4\!\times\!10^{-1}$ & $2.8\!\times\!10^{-1}$ & {\boldmath $+2.5\!\times\!10^{-1}$} & 1.000 \\
 & Overlapping, Imbalanced & 15 & $6.2\!\times\!10^{-1}$ & $5.0\!\times\!10^{-1}$ & {\boldmath $+1.2\!\times\!10^{-1}$} & 1.000 \\
 & Overlapping, Imbalanced & 30 & $4.7\!\times\!10^{-1}$ & $2.5\!\times\!10^{-1}$ & {\boldmath $+2.2\!\times\!10^{-1}$} & 1.000 \\
\midrule
\multirow{8}{*}{FMNIST} & Non-overlapping, Balanced & 15 & $9.1\!\times\!10^{-1}$ & $2.0\!\times\!10^{-1}$ & {\boldmath $+7.1\!\times\!10^{-1}$} & 1.000 \\
 & Non-overlapping, Balanced & 30 & $9.5\!\times\!10^{-1}$ & $2.8\!\times\!10^{-1}$ & {\boldmath $+6.7\!\times\!10^{-1}$} & 1.000 \\
 & Non-overlapping, Imbalanced & 15 & $9.1\!\times\!10^{-1}$ & $5.0\!\times\!10^{-1}$ & {\boldmath $+4.1\!\times\!10^{-1}$} & 1.000 \\
 & Non-overlapping, Imbalanced & 30 & $9.3\!\times\!10^{-1}$ & $3.9\!\times\!10^{-1}$ & {\boldmath $+5.4\!\times\!10^{-1}$} & 1.000 \\
 & Overlapping, Balanced & 15 & $6.2\!\times\!10^{-1}$ & $3.6\!\times\!10^{-1}$ & {\boldmath $+2.6\!\times\!10^{-1}$} & 1.000 \\
 & Overlapping, Balanced & 30 & $4.7\!\times\!10^{-1}$ & $2.5\!\times\!10^{-1}$ & {\boldmath $+2.2\!\times\!10^{-1}$} & 1.000 \\
 & Overlapping, Imbalanced & 15 & $5.9\!\times\!10^{-1}$ & $2.3\!\times\!10^{-1}$ & {\boldmath $+3.6\!\times\!10^{-1}$} & 1.000 \\
 & Overlapping, Imbalanced & 30 & $5.4\!\times\!10^{-1}$ & $3.5\!\times\!10^{-1}$ & {\boldmath $+1.9\!\times\!10^{-1}$} & 1.000 \\
\midrule
\multirow{8}{*}{CIFAR10} & Non-overlapping, Balanced & 15 & $5.1\!\times\!10^{-4}$ & $8.3\!\times\!10^{-4}$ & $-3.2\!\times\!10^{-4}$ & 0.668 \\
 & Non-overlapping, Balanced & 30 & $1.2\!\times\!10^{-3}$ & $2.0\!\times\!10^{-3}$ & $-8.7\!\times\!10^{-4}$ & 0.667 \\
 & Non-overlapping, Imbalanced & 15 & $5.1\!\times\!10^{-4}$ & $8.4\!\times\!10^{-4}$ & $-3.3\!\times\!10^{-4}$ & 0.446 \\
 & Non-overlapping, Imbalanced & 30 & $1.1\!\times\!10^{-3}$ & $2.5\!\times\!10^{-3}$ & $-1.4\!\times\!10^{-3}$ & 0.398 \\
 & Overlapping, Balanced & 15 & $6.5\!\times\!10^{-4}$ & $7.1\!\times\!10^{-4}$ & $-5.8\!\times\!10^{-5}$ & 0.845 \\
 & Overlapping, Balanced & 30 & $1.7\!\times\!10^{-3}$ & $2.0\!\times\!10^{-3}$ & $-3.0\!\times\!10^{-4}$ & 0.780 \\
 & Overlapping, Imbalanced & 15 & $5.8\!\times\!10^{-4}$ & $6.6\!\times\!10^{-4}$ & $-7.9\!\times\!10^{-5}$ & 0.925 \\
 & Overlapping, Imbalanced & 30 & $1.7\!\times\!10^{-3}$ & $1.9\!\times\!10^{-3}$ & $-2.8\!\times\!10^{-4}$ & 0.797 \\
\midrule
\multirow{8}{*}{PATHMNIST} & Non-overlapping, Balanced & 15 & $1.4\!\times\!10^{-6}$ & $8.3\!\times\!10^{-6}$ & $-6.9\!\times\!10^{-6}$ & 0.945 \\
 & Non-overlapping, Balanced & 30 & $2.5\!\times\!10^{-6}$ & $4.2\!\times\!10^{-6}$ & $-1.7\!\times\!10^{-6}$ & 0.980 \\
 & Non-overlapping, Imbalanced & 15 & $6.7\!\times\!10^{-6}$ & $3.6\!\times\!10^{-6}$ & {\boldmath $+3.1\!\times\!10^{-6}$} & 1.000 \\
 & Non-overlapping, Imbalanced & 30 & $5.4\!\times\!10^{-6}$ & $9.8\!\times\!10^{-6}$ & $-4.5\!\times\!10^{-6}$ & 0.998 \\
 & Overlapping, Balanced & 15 & $5.7\!\times\!10^{-7}$ & $1.7\!\times\!10^{-6}$ & $-1.1\!\times\!10^{-6}$ & 0.968 \\
 & Overlapping, Balanced & 30 & $1.5\!\times\!10^{-6}$ & $9.7\!\times\!10^{-6}$ & $-8.2\!\times\!10^{-6}$ & 0.938 \\
 & Overlapping, Imbalanced & 15 & $5.9\!\times\!10^{-6}$ & $8.3\!\times\!10^{-7}$ & {\boldmath $+5.1\!\times\!10^{-6}$} & 1.000 \\
 & Overlapping, Imbalanced & 30 & $1.2\!\times\!10^{-5}$ & $7.4\!\times\!10^{-6}$ & {\boldmath $+5.0\!\times\!10^{-6}$} & 1.000 \\
\midrule
\multirow{8}{*}{BLOODMNIST} & Non-overlapping, Balanced & 15 & $7.6\!\times\!10^{-5}$ & $1.9\!\times\!10^{-5}$ & {\boldmath $+5.6\!\times\!10^{-5}$} & 1.000 \\
 & Non-overlapping, Balanced & 30 & $1.2\!\times\!10^{-4}$ & $6.3\!\times\!10^{-5}$ & {\boldmath $+5.5\!\times\!10^{-5}$} & 1.000 \\
 & Non-overlapping, Imbalanced & 15 & $7.2\!\times\!10^{-5}$ & $2.4\!\times\!10^{-5}$ & {\boldmath $+4.7\!\times\!10^{-5}$} & 1.000 \\
 & Non-overlapping, Imbalanced & 30 & $1.2\!\times\!10^{-4}$ & $7.1\!\times\!10^{-5}$ & {\boldmath $+4.9\!\times\!10^{-5}$} & 1.000 \\
 & Overlapping, Balanced & 15 & $4.6\!\times\!10^{-5}$ & $2.3\!\times\!10^{-5}$ & {\boldmath $+2.4\!\times\!10^{-5}$} & 1.000 \\
 & Overlapping, Balanced & 30 & $7.1\!\times\!10^{-5}$ & $8.8\!\times\!10^{-5}$ & $-1.7\!\times\!10^{-5}$ & 0.998 \\
 & Overlapping, Imbalanced & 15 & $5.0\!\times\!10^{-5}$ & $2.6\!\times\!10^{-5}$ & {\boldmath $+2.4\!\times\!10^{-5}$} & 1.000 \\
 & Overlapping, Imbalanced & 30 & $4.8\!\times\!10^{-5}$ & $7.4\!\times\!10^{-4}$ & $-7.0\!\times\!10^{-4}$ & 0.845 \\

    \bottomrule
  \end{tabular}
\end{table}

\begin{table}[t]
  \centering
  \caption{Own-partition separation margin of \textsc{HDBSCAN} ($M^{\mathrm{HDBSCAN}}_{\tau_T}$)
  versus the ground-truth margin $M^\star_{\tau_T}$ at the triggered round. Positive (block-separated)
  values are in bold.}
  \label{app:margin-hdbscan}
  \small
  \begin{tabular}{ll c cc}
    \toprule
    Dataset & Split & $|P|$ & $M^{\mathrm{HDBSCAN}}_{\tau_T}$ & $M^\star_{\tau_T}$ \\
    \midrule
\multirow{8}{*}{MNIST} & Non-overlapping, Balanced & 15 & {\boldmath $+6.0\!\times\!10^{-1}$} & {\boldmath $+6.0\!\times\!10^{-1}$} \\
 & Non-overlapping, Balanced & 30 & {\boldmath $+6.2\!\times\!10^{-1}$} & {\boldmath $+6.2\!\times\!10^{-1}$} \\
 & Non-overlapping, Imbalanced & 15 & {\boldmath $+7.4\!\times\!10^{-1}$} & {\boldmath $+7.4\!\times\!10^{-1}$} \\
 & Non-overlapping, Imbalanced & 30 & {\boldmath $+4.4\!\times\!10^{-1}$} & {\boldmath $+4.4\!\times\!10^{-1}$} \\
 & Overlapping, Balanced & 15 & {\boldmath $+2.3\!\times\!10^{-1}$} & {\boldmath $+2.3\!\times\!10^{-1}$} \\
 & Overlapping, Balanced & 30 & {\boldmath $+2.5\!\times\!10^{-1}$} & {\boldmath $+2.5\!\times\!10^{-1}$} \\
 & Overlapping, Imbalanced & 15 & {\boldmath $+1.2\!\times\!10^{-1}$} & {\boldmath $+1.2\!\times\!10^{-1}$} \\
 & Overlapping, Imbalanced & 30 & {\boldmath $+2.2\!\times\!10^{-1}$} & {\boldmath $+2.2\!\times\!10^{-1}$} \\
\midrule
\multirow{8}{*}{FMNIST} & Non-overlapping, Balanced & 15 & {\boldmath $+7.1\!\times\!10^{-1}$} & {\boldmath $+7.1\!\times\!10^{-1}$} \\
 & Non-overlapping, Balanced & 30 & {\boldmath $+6.7\!\times\!10^{-1}$} & {\boldmath $+6.7\!\times\!10^{-1}$} \\
 & Non-overlapping, Imbalanced & 15 & {\boldmath $+4.1\!\times\!10^{-1}$} & {\boldmath $+4.1\!\times\!10^{-1}$} \\
 & Non-overlapping, Imbalanced & 30 & {\boldmath $+5.4\!\times\!10^{-1}$} & {\boldmath $+5.4\!\times\!10^{-1}$} \\
 & Overlapping, Balanced & 15 & {\boldmath $+2.6\!\times\!10^{-1}$} & {\boldmath $+2.6\!\times\!10^{-1}$} \\
 & Overlapping, Balanced & 30 & {\boldmath $+2.2\!\times\!10^{-1}$} & {\boldmath $+2.2\!\times\!10^{-1}$} \\
 & Overlapping, Imbalanced & 15 & {\boldmath $+3.6\!\times\!10^{-1}$} & {\boldmath $+3.6\!\times\!10^{-1}$} \\
 & Overlapping, Imbalanced & 30 & {\boldmath $+1.9\!\times\!10^{-1}$} & {\boldmath $+1.9\!\times\!10^{-1}$} \\
\midrule
\multirow{8}{*}{CIFAR10} & Non-overlapping, Balanced & 15 & -- & $-3.2\!\times\!10^{-4}$ \\
 & Non-overlapping, Balanced & 30 & -- & $-8.7\!\times\!10^{-4}$ \\
 & Non-overlapping, Imbalanced & 15 & $-3.3\!\times\!10^{-4}$ & $-3.3\!\times\!10^{-4}$ \\
 & Non-overlapping, Imbalanced & 30 & -- & $-1.4\!\times\!10^{-3}$ \\
 & Overlapping, Balanced & 15 & $-1.0\!\times\!10^{-4}$ & $-5.8\!\times\!10^{-5}$ \\
 & Overlapping, Balanced & 30 & -- & $-3.0\!\times\!10^{-4}$ \\
 & Overlapping, Imbalanced & 15 & $-8.6\!\times\!10^{-5}$ & $-7.9\!\times\!10^{-5}$ \\
 & Overlapping, Imbalanced & 30 & -- & $-2.8\!\times\!10^{-4}$ \\
\midrule
\multirow{8}{*}{PATHMNIST} & Non-overlapping, Balanced & 15 & $-6.9\!\times\!10^{-6}$ & $-6.9\!\times\!10^{-6}$ \\
 & Non-overlapping, Balanced & 30 & {\boldmath $+2.6\!\times\!10^{-5}$} & $-1.7\!\times\!10^{-6}$ \\
 & Non-overlapping, Imbalanced & 15 & {\boldmath $+3.1\!\times\!10^{-6}$} & {\boldmath $+3.1\!\times\!10^{-6}$} \\
 & Non-overlapping, Imbalanced & 30 & $-4.5\!\times\!10^{-6}$ & $-4.5\!\times\!10^{-6}$ \\
 & Overlapping, Balanced & 15 & {\boldmath $+5.4\!\times\!10^{-7}$} & $-1.1\!\times\!10^{-6}$ \\
 & Overlapping, Balanced & 30 & $-4.6\!\times\!10^{-7}$ & $-8.2\!\times\!10^{-6}$ \\
 & Overlapping, Imbalanced & 15 & {\boldmath $+5.1\!\times\!10^{-6}$} & {\boldmath $+5.1\!\times\!10^{-6}$} \\
 & Overlapping, Imbalanced & 30 & {\boldmath $+5.0\!\times\!10^{-6}$} & {\boldmath $+5.0\!\times\!10^{-6}$} \\
\midrule
\multirow{8}{*}{BLOODMNIST} & Non-overlapping, Balanced & 15 & {\boldmath $+5.6\!\times\!10^{-5}$} & {\boldmath $+5.6\!\times\!10^{-5}$} \\
 & Non-overlapping, Balanced & 30 & {\boldmath $+5.5\!\times\!10^{-5}$} & {\boldmath $+5.5\!\times\!10^{-5}$} \\
 & Non-overlapping, Imbalanced & 15 & {\boldmath $+4.7\!\times\!10^{-5}$} & {\boldmath $+4.7\!\times\!10^{-5}$} \\
 & Non-overlapping, Imbalanced & 30 & {\boldmath $+4.9\!\times\!10^{-5}$} & {\boldmath $+4.9\!\times\!10^{-5}$} \\
 & Overlapping, Balanced & 15 & {\boldmath $+2.4\!\times\!10^{-5}$} & {\boldmath $+2.4\!\times\!10^{-5}$} \\
 & Overlapping, Balanced & 30 & $-3.9\!\times\!10^{-5}$ & $-1.7\!\times\!10^{-5}$ \\
 & Overlapping, Imbalanced & 15 & {\boldmath $+2.4\!\times\!10^{-5}$} & {\boldmath $+2.4\!\times\!10^{-5}$} \\
 & Overlapping, Imbalanced & 30 & {\boldmath $+1.8\!\times\!10^{-6}$} & $-7.0\!\times\!10^{-4}$ \\

    \bottomrule
  \end{tabular}
\end{table}

\begin{table}[t]
  \centering
  \caption{Own-partition separation margin of \textsc{K-Means} ($M^{\mathrm{KMeans}}_{\tau_T}$)
  versus the ground-truth margin $M^\star_{\tau_T}$ at the triggered round. Positive (block-separated)
  values are in bold.}
  \label{app:margin-kmeans}
  \small
  \begin{tabular}{ll c cc}
    \toprule
    Dataset & Split & $|P|$ & $M^{\mathrm{KMeans}}_{\tau_T}$ & $M^\star_{\tau_T}$ \\
    \midrule
\multirow{8}{*}{MNIST} & Non-overlapping, Balanced & 15 & {\boldmath $+6.4\!\times\!10^{-1}$} & {\boldmath $+6.0\!\times\!10^{-1}$} \\
 & Non-overlapping, Balanced & 30 & {\boldmath $+6.5\!\times\!10^{-1}$} & {\boldmath $+6.2\!\times\!10^{-1}$} \\
 & Non-overlapping, Imbalanced & 15 & {\boldmath $+7.4\!\times\!10^{-1}$} & {\boldmath $+7.4\!\times\!10^{-1}$} \\
 & Non-overlapping, Imbalanced & 30 & {\boldmath $+4.4\!\times\!10^{-1}$} & {\boldmath $+4.4\!\times\!10^{-1}$} \\
 & Overlapping, Balanced & 15 & {\boldmath $+2.0\!\times\!10^{-1}$} & {\boldmath $+2.3\!\times\!10^{-1}$} \\
 & Overlapping, Balanced & 30 & {\boldmath $+2.5\!\times\!10^{-1}$} & {\boldmath $+2.5\!\times\!10^{-1}$} \\
 & Overlapping, Imbalanced & 15 & {\boldmath $+1.5\!\times\!10^{-1}$} & {\boldmath $+1.2\!\times\!10^{-1}$} \\
 & Overlapping, Imbalanced & 30 & {\boldmath $+2.3\!\times\!10^{-1}$} & {\boldmath $+2.2\!\times\!10^{-1}$} \\
\midrule
\multirow{8}{*}{FMNIST} & Non-overlapping, Balanced & 15 & {\boldmath $+7.1\!\times\!10^{-1}$} & {\boldmath $+7.1\!\times\!10^{-1}$} \\
 & Non-overlapping, Balanced & 30 & {\boldmath $+6.7\!\times\!10^{-1}$} & {\boldmath $+6.7\!\times\!10^{-1}$} \\
 & Non-overlapping, Imbalanced & 15 & {\boldmath $+4.9\!\times\!10^{-1}$} & {\boldmath $+4.1\!\times\!10^{-1}$} \\
 & Non-overlapping, Imbalanced & 30 & {\boldmath $+5.3\!\times\!10^{-1}$} & {\boldmath $+5.4\!\times\!10^{-1}$} \\
 & Overlapping, Balanced & 15 & {\boldmath $+2.6\!\times\!10^{-1}$} & {\boldmath $+2.6\!\times\!10^{-1}$} \\
 & Overlapping, Balanced & 30 & {\boldmath $+2.2\!\times\!10^{-1}$} & {\boldmath $+2.2\!\times\!10^{-1}$} \\
 & Overlapping, Imbalanced & 15 & {\boldmath $+3.6\!\times\!10^{-1}$} & {\boldmath $+3.6\!\times\!10^{-1}$} \\
 & Overlapping, Imbalanced & 30 & {\boldmath $+2.4\!\times\!10^{-1}$} & {\boldmath $+1.9\!\times\!10^{-1}$} \\
\midrule
\multirow{8}{*}{CIFAR10} & Non-overlapping, Balanced & 15 & $-1.9\!\times\!10^{-4}$ & $-3.2\!\times\!10^{-4}$ \\
 & Non-overlapping, Balanced & 30 & $-8.8\!\times\!10^{-4}$ & $-8.7\!\times\!10^{-4}$ \\
 & Non-overlapping, Imbalanced & 15 & $-3.4\!\times\!10^{-4}$ & $-3.3\!\times\!10^{-4}$ \\
 & Non-overlapping, Imbalanced & 30 & $-7.9\!\times\!10^{-4}$ & $-1.4\!\times\!10^{-3}$ \\
 & Overlapping, Balanced & 15 & $-8.6\!\times\!10^{-5}$ & $-5.8\!\times\!10^{-5}$ \\
 & Overlapping, Balanced & 30 & $-3.5\!\times\!10^{-4}$ & $-3.0\!\times\!10^{-4}$ \\
 & Overlapping, Imbalanced & 15 & $-7.9\!\times\!10^{-5}$ & $-7.9\!\times\!10^{-5}$ \\
 & Overlapping, Imbalanced & 30 & $-3.4\!\times\!10^{-4}$ & $-2.8\!\times\!10^{-4}$ \\
\midrule
\multirow{8}{*}{PATHMNIST} & Non-overlapping, Balanced & 15 & {\boldmath $+9.2\!\times\!10^{-7}$} & $-6.9\!\times\!10^{-6}$ \\
 & Non-overlapping, Balanced & 30 & $-2.9\!\times\!10^{-6}$ & $-1.7\!\times\!10^{-6}$ \\
 & Non-overlapping, Imbalanced & 15 & {\boldmath $+3.1\!\times\!10^{-6}$} & {\boldmath $+3.1\!\times\!10^{-6}$} \\
 & Non-overlapping, Imbalanced & 30 & $-4.5\!\times\!10^{-6}$ & $-4.5\!\times\!10^{-6}$ \\
 & Overlapping, Balanced & 15 & $-7.5\!\times\!10^{-7}$ & $-1.1\!\times\!10^{-6}$ \\
 & Overlapping, Balanced & 30 & $-1.9\!\times\!10^{-6}$ & $-8.2\!\times\!10^{-6}$ \\
 & Overlapping, Imbalanced & 15 & {\boldmath $+5.1\!\times\!10^{-6}$} & {\boldmath $+5.1\!\times\!10^{-6}$} \\
 & Overlapping, Imbalanced & 30 & {\boldmath $+5.0\!\times\!10^{-6}$} & {\boldmath $+5.0\!\times\!10^{-6}$} \\
\midrule
\multirow{8}{*}{BLOODMNIST} & Non-overlapping, Balanced & 15 & {\boldmath $+7.0\!\times\!10^{-5}$} & {\boldmath $+5.6\!\times\!10^{-5}$} \\
 & Non-overlapping, Balanced & 30 & {\boldmath $+5.1\!\times\!10^{-5}$} & {\boldmath $+5.5\!\times\!10^{-5}$} \\
 & Non-overlapping, Imbalanced & 15 & {\boldmath $+3.2\!\times\!10^{-5}$} & {\boldmath $+4.7\!\times\!10^{-5}$} \\
 & Non-overlapping, Imbalanced & 30 & {\boldmath $+7.0\!\times\!10^{-5}$} & {\boldmath $+4.9\!\times\!10^{-5}$} \\
 & Overlapping, Balanced & 15 & {\boldmath $+9.6\!\times\!10^{-6}$} & {\boldmath $+2.4\!\times\!10^{-5}$} \\
 & Overlapping, Balanced & 30 & $-1.3\!\times\!10^{-6}$ & $-1.7\!\times\!10^{-5}$ \\
 & Overlapping, Imbalanced & 15 & {\boldmath $+2.4\!\times\!10^{-5}$} & {\boldmath $+2.4\!\times\!10^{-5}$} \\
 & Overlapping, Imbalanced & 30 & $-5.3\!\times\!10^{-5}$ & $-7.0\!\times\!10^{-4}$ \\

    \bottomrule
  \end{tabular}
\end{table}

\begin{table}[t]
  \centering
  \caption{Own-partition separation margin of \textsc{Agglomerative} ($M^{\mathrm{Agglomerative}}_{\tau_T}$)
  versus the ground-truth margin $M^\star_{\tau_T}$ at the triggered round. Positive (block-separated)
  values are in bold.}
  \label{app:margin-agglomerative}
  \small
  \begin{tabular}{ll c cc}
    \toprule
    Dataset & Split & $|P|$ & $M^{\mathrm{Agglomerative}}_{\tau_T}$ & $M^\star_{\tau_T}$ \\
    \midrule
\multirow{8}{*}{MNIST} & Non-overlapping, Balanced & 15 & {\boldmath $+6.0\!\times\!10^{-1}$} & {\boldmath $+6.0\!\times\!10^{-1}$} \\
 & Non-overlapping, Balanced & 30 & {\boldmath $+6.2\!\times\!10^{-1}$} & {\boldmath $+6.2\!\times\!10^{-1}$} \\
 & Non-overlapping, Imbalanced & 15 & {\boldmath $+7.4\!\times\!10^{-1}$} & {\boldmath $+7.4\!\times\!10^{-1}$} \\
 & Non-overlapping, Imbalanced & 30 & {\boldmath $+4.4\!\times\!10^{-1}$} & {\boldmath $+4.4\!\times\!10^{-1}$} \\
 & Overlapping, Balanced & 15 & {\boldmath $+2.3\!\times\!10^{-1}$} & {\boldmath $+2.3\!\times\!10^{-1}$} \\
 & Overlapping, Balanced & 30 & {\boldmath $+2.5\!\times\!10^{-1}$} & {\boldmath $+2.5\!\times\!10^{-1}$} \\
 & Overlapping, Imbalanced & 15 & {\boldmath $+1.2\!\times\!10^{-1}$} & {\boldmath $+1.2\!\times\!10^{-1}$} \\
 & Overlapping, Imbalanced & 30 & {\boldmath $+2.2\!\times\!10^{-1}$} & {\boldmath $+2.2\!\times\!10^{-1}$} \\
\midrule
\multirow{8}{*}{FMNIST} & Non-overlapping, Balanced & 15 & {\boldmath $+7.1\!\times\!10^{-1}$} & {\boldmath $+7.1\!\times\!10^{-1}$} \\
 & Non-overlapping, Balanced & 30 & {\boldmath $+6.7\!\times\!10^{-1}$} & {\boldmath $+6.7\!\times\!10^{-1}$} \\
 & Non-overlapping, Imbalanced & 15 & {\boldmath $+4.1\!\times\!10^{-1}$} & {\boldmath $+4.1\!\times\!10^{-1}$} \\
 & Non-overlapping, Imbalanced & 30 & {\boldmath $+5.4\!\times\!10^{-1}$} & {\boldmath $+5.4\!\times\!10^{-1}$} \\
 & Overlapping, Balanced & 15 & {\boldmath $+2.6\!\times\!10^{-1}$} & {\boldmath $+2.6\!\times\!10^{-1}$} \\
 & Overlapping, Balanced & 30 & {\boldmath $+2.2\!\times\!10^{-1}$} & {\boldmath $+2.2\!\times\!10^{-1}$} \\
 & Overlapping, Imbalanced & 15 & {\boldmath $+3.6\!\times\!10^{-1}$} & {\boldmath $+3.6\!\times\!10^{-1}$} \\
 & Overlapping, Imbalanced & 30 & {\boldmath $+1.9\!\times\!10^{-1}$} & {\boldmath $+1.9\!\times\!10^{-1}$} \\
\midrule
\multirow{8}{*}{CIFAR10} & Non-overlapping, Balanced & 15 & $-1.7\!\times\!10^{-4}$ & $-3.2\!\times\!10^{-4}$ \\
 & Non-overlapping, Balanced & 30 & $-5.9\!\times\!10^{-4}$ & $-8.7\!\times\!10^{-4}$ \\
 & Non-overlapping, Imbalanced & 15 & $-1.1\!\times\!10^{-4}$ & $-3.3\!\times\!10^{-4}$ \\
 & Non-overlapping, Imbalanced & 30 & $-4.4\!\times\!10^{-4}$ & $-1.4\!\times\!10^{-3}$ \\
 & Overlapping, Balanced & 15 & $-5.5\!\times\!10^{-5}$ & $-5.8\!\times\!10^{-5}$ \\
 & Overlapping, Balanced & 30 & $-3.0\!\times\!10^{-4}$ & $-3.0\!\times\!10^{-4}$ \\
 & Overlapping, Imbalanced & 15 & $-7.9\!\times\!10^{-5}$ & $-7.9\!\times\!10^{-5}$ \\
 & Overlapping, Imbalanced & 30 & $-2.7\!\times\!10^{-4}$ & $-2.8\!\times\!10^{-4}$ \\
\midrule
\multirow{8}{*}{PATHMNIST} & Non-overlapping, Balanced & 15 & {\boldmath $+9.1\!\times\!10^{-7}$} & $-6.9\!\times\!10^{-6}$ \\
 & Non-overlapping, Balanced & 30 & $-3.0\!\times\!10^{-6}$ & $-1.7\!\times\!10^{-6}$ \\
 & Non-overlapping, Imbalanced & 15 & {\boldmath $+3.1\!\times\!10^{-6}$} & {\boldmath $+3.1\!\times\!10^{-6}$} \\
 & Non-overlapping, Imbalanced & 30 & $-4.5\!\times\!10^{-6}$ & $-4.5\!\times\!10^{-6}$ \\
 & Overlapping, Balanced & 15 & $-7.7\!\times\!10^{-7}$ & $-1.1\!\times\!10^{-6}$ \\
 & Overlapping, Balanced & 30 & $-2.6\!\times\!10^{-6}$ & $-8.2\!\times\!10^{-6}$ \\
 & Overlapping, Imbalanced & 15 & {\boldmath $+5.1\!\times\!10^{-6}$} & {\boldmath $+5.1\!\times\!10^{-6}$} \\
 & Overlapping, Imbalanced & 30 & {\boldmath $+5.0\!\times\!10^{-6}$} & {\boldmath $+5.0\!\times\!10^{-6}$} \\
\midrule
\multirow{8}{*}{BLOODMNIST} & Non-overlapping, Balanced & 15 & {\boldmath $+5.6\!\times\!10^{-5}$} & {\boldmath $+5.6\!\times\!10^{-5}$} \\
 & Non-overlapping, Balanced & 30 & {\boldmath $+5.5\!\times\!10^{-5}$} & {\boldmath $+5.5\!\times\!10^{-5}$} \\
 & Non-overlapping, Imbalanced & 15 & {\boldmath $+4.7\!\times\!10^{-5}$} & {\boldmath $+4.7\!\times\!10^{-5}$} \\
 & Non-overlapping, Imbalanced & 30 & {\boldmath $+4.9\!\times\!10^{-5}$} & {\boldmath $+4.9\!\times\!10^{-5}$} \\
 & Overlapping, Balanced & 15 & {\boldmath $+2.4\!\times\!10^{-5}$} & {\boldmath $+2.4\!\times\!10^{-5}$} \\
 & Overlapping, Balanced & 30 & $-1.7\!\times\!10^{-5}$ & $-1.7\!\times\!10^{-5}$ \\
 & Overlapping, Imbalanced & 15 & {\boldmath $+2.4\!\times\!10^{-5}$} & {\boldmath $+2.4\!\times\!10^{-5}$} \\
 & Overlapping, Imbalanced & 30 & {\boldmath $+1.8\!\times\!10^{-6}$} & $-7.0\!\times\!10^{-4}$ \\

    \bottomrule
  \end{tabular}
\end{table}

\clearpage
\FloatBarrier
\subsection{Additional Clustering Results}
\label{app:clustering_results}

\begin{table}[h!]
  \centering
  \resizebox{0.55\columnwidth}{!}{%
    \tiny
    \begin{tabular}{cc|cc|cc}
      \toprule
      Dataset      & Algorithm & \multicolumn{2}{c|}{Non Overlapping} & \multicolumn{2}{c}{Overlapping} \\
                   &           & Balanced   & Imbalanced      & Balanced      & Imbalanced     \\
      \midrule
      \multirow{6}{*}{MNIST\,15}
                   & SCL       & 31 & 23 & 25 & 24 \\
                   & BCL       & 21 & 21 & 21 & 21 \\
                   & OCFL-KM   & 3 & 3 & 3 & 3 \\
                   & OCFL-AFF  & 3 & 3 & 3 & 4 \\
                   & OCFL-MS   & 3 & 3 & 3 & 3 \\
                   & OCFL-HDBS  & 3 & 3 & 3 & 3 \\
      \bottomrule
      \multirow{6}{*}{MNIST\,30}
                   & SCL       & 44 & 23 & 35 & 34 \\
                   & BCL       & 21 & 21 & 21 & 21 \\
                   & OCFL-KM   & 3 & 3 & 6 & 4 \\
                   & OCFL-AFF  & 3 & 4 & 5 & 4 \\
                   & OCFL-MS   & 4 & 4 & 5 & 4 \\
                   & OCFL-HDBS  & 5 & 2 & 4 & 4 \\
      \bottomrule
      \multirow{6}{*}{FMNIST\,15}
                   & SCL       & 46 & 24 & 34 & 35 \\
                   & BCL       & 21 & 21 & 21 & 21 \\
                   & OCFL-KM   & 2 & 2 & 2 & 2 \\
                   & OCFL-AFF  & 2 & 3 & 3 & 3 \\
                   & OCFL-MS   & 2 & 2 & 2 & 2 \\
                   & OCFL-HDBS  & 3 & 2 & 3 & 2 \\
      \bottomrule
      \multirow{6}{*}{FMNIST\,30}
                   & SCL       & 23 & 23 & 50 & 50 \\
                   & BCL       & 21 & 21 & 21 & 21 \\
                   & OCFL-KM   & 3 & 2 & 3 & 3 \\
                   & OCFL-AFF  & 2 & 3 & 4 & 3 \\
                   & OCFL-MS   & 2 & 3 & 4 & 2 \\
                   & OCFL-HDBS  & 2 & 2 & 4 & 3 \\
      \bottomrule
      \multirow{6}{*}{CIFAR10\,15}
                   & SCL       & -- & -- & -- & -- \\
                   & BCL       & 21 & 21 & 21 & 21 \\
                   & OCFL-KM   & 8 & 6 & 6 & 6 \\
                   & OCFL-AFF  & 6 & 6 & 6 & 7 \\
                   & OCFL-MS   & 5 & 7 & 6 & 7 \\
                   & OCFL-HDBS  & 5 & 9 & 5 & 6 \\
      \bottomrule
      \multirow{6}{*}{CIFAR10\,30}
                   & SCL       & -- & -- & -- & -- \\
                   & BCL       & 21 & 21 & 21 & 21 \\
                   & OCFL-KM   & 7 & 8 & 6 & 7 \\
                   & OCFL-AFF  & 7 & 8 & 6 & 7 \\
                   & OCFL-MS   & 8 & 7 & 6 & 7 \\
                   & OCFL-HDBS  & 8 & 7 & 7 & 6 \\
      \bottomrule
      \multirow{6}{*}{PATHMNIST\,15}
                   & SCL       & -- & -- & -- & -- \\
                   & BCL       & 21 & 21 & 21 & 21 \\
                   & OCFL-KM   & 3 & 3 & 3 & 3 \\
                   & OCFL-AFF  & 2 & 4 & 2 & 3 \\
                   & OCFL-MS   & 2 & 3 & 4 & 3 \\
                   & OCFL-HDBS  & 4 & 3 & 2 & 3 \\
      \bottomrule
      \multirow{6}{*}{PATHMNIST\,30}
                   & SCL       & -- & -- & -- & -- \\
                   & BCL       & 21 & 21 & 21 & 21 \\
                   & OCFL-KM   & 2 & 3 & 2 & 2 \\
                   & OCFL-AFF  & 2 & 4 & 2 & 2 \\
                   & OCFL-MS   & 2 & 3 & 2 & 2 \\
                   & OCFL-HDBS  & 2 & 3 & 2 & 2 \\
      \bottomrule
      \multirow{6}{*}{BLOODMNIST\,15}
                   & SCL       & -- & -- & -- & -- \\
                   & BCL       & 21 & 21 & 21 & 21 \\
                   & OCFL-KM   & 2 & 5 & 2 & 2 \\
                   & OCFL-AFF  & 2 & 5 & 2 & 2 \\
                   & OCFL-MS   & 2 & 5 & 2 & 2 \\
                   & OCFL-HDBS  & 2 & 5 & 2 & 2 \\
      \bottomrule
      \multirow{6}{*}{BLOODMNIST\,30}
                   & SCL       & -- & -- & -- & -- \\
                   & BCL       & 21 & 21 & 21 & 21 \\
                   & OCFL-KM   & 2 & 2 & 2 & 2 \\
                   & OCFL-AFF  & 2 & 2 & 2 & 2 \\
                   & OCFL-MS   & 2 & 2 & 2 & 2 \\
                   & OCFL-HDBS  & 2 & 2 & 2 & 2 \\
      \bottomrule
    \end{tabular}%
  }
  \caption{Table presenting the clustering round for all datasets (with a number of clients equal to 15 or 30), data splits, and across six different clustering algorithms: \cite{b7} (SCL), \cite{b1} (BCL), OCFL with K-Means Algorithm (OCFL-KM), OCFL with Affinity Propagation (OCFL-AFF), OCFL with Mean-Shift algorithm (OCFL-MS) and OCFL with HDBSCAN algorithm (OCFL-HDBS). The double hyphen (--) is placed if the algorithm has failed to perform any clustering in the given scenario. In the case of the \cite{b7} algorithm, the last clustering round is displayed.}
\end{table}

\clearpage
\FloatBarrier

\subsection{Computation of the separation margin and pair-order agreement}
\label{app:margin-protocol}

The following Section describes the exact procedure used for computation of the pair-order agreement $A^\star_{\tau_T}$ as displayed in Tables \ref{tab:margin} and \ref{app:margin}. It also explains the per-algorithm margin calculation displayed in Tables~\ref{app:margin-hdbscan} - \ref{app:margin-agglomerative}.

\paragraph{Inputs.} (i) The divergence matrix $\boldsymbol{\Gamma}_{\tau_T}\in\mathbb{R}^{n\times n}$
of pairwise cosine dissimilarities between client updates (Definition~\ref{def:Divergence_Matrix}).
We use the matrix of the temperature-triggered round $\tau_T$,
identified as the round at which Algorithm~\ref{algo:ocfl} performed its one-shot clustering.
(ii) The true DGP assignment $z\in\{1,2,3\}^{n}$, known by construction of the simulated splits:
clients are allocated to the three DGPs in contiguous blocks of sizes $5/5/5$ ($|P|=15$) and
$10/10/10$ ($|P|=30$) in the balanced setting, and $3/7/5$ and $7/14/9$ in the imbalanced setting,
identically across datasets and overlap settings.

\paragraph{Pair sets.} Let
$\mathcal{W}=\{\{i,j\}: z_i=z_j,\ i\neq j\}$ and
$\mathcal{B}=\{\{i,j\}: z_i\neq z_j\}$
denote the unordered within-DGP and between-DGP client pairs, with
$|\mathcal{W}|+|\mathcal{B}|=\binom{n}{2}$.

\paragraph{Separation margin.} We report
$W^\star_{\tau_T}=\max_{\{i,j\}\in\mathcal{W}}\boldsymbol{\Gamma}_{\tau_T}(i,j)$,
$B^\star_{\tau_T}=\min_{\{i,j\}\in\mathcal{B}}\boldsymbol{\Gamma}_{\tau_T}(i,j)$, and
$M^\star_{\tau_T}=B^\star_{\tau_T}-W^\star_{\tau_T}$
(Equation~\eqref{eq:patch-true-margin}). $M^\star_{\tau_T}$ is a worst-case scenario statistic: its value is
determined by exactly one between-DGP pair and one within-DGP pair.

\paragraph{Pair-order agreement.} We additionally report
\begin{equation}
A^\star_{\tau_T}
=\frac{1}{|\mathcal{B}|\,|\mathcal{W}|}
\sum_{b\in\mathcal{B}}\sum_{w\in\mathcal{W}}
\mathbb{1}\!\left[\boldsymbol{\Gamma}_{\tau_T}(b)>\boldsymbol{\Gamma}_{\tau_T}(w)\right],
\label{eq:pair-order}
\end{equation}
the fraction of (between, within) pair couples in the correct order - equivalently, the
Mann-Whitney statistic (ROC AUC) of the entries of $\boldsymbol{\Gamma}_{\tau_T}$ as a separator
of between- versus within-DGP pairs. Ties count as violations, which is a conservative experiment design.
$A^\star_{\tau_T}\in[0,1]$; $A^\star_{\tau_T}=\tfrac12$ corresponds to chance ordering, and
$M^\star_{\tau_T}>0$ holds if and only if $A^\star_{\tau_T}=1$. Unlike $M^\star_{\tau_T}$, a single
boundary pair changes $A^\star_{\tau_T}$ by at most $\max(|\mathcal{W}|,|\mathcal{B}|)^{-1}$
of its range, and the statistic depends only on the ordering of the entries of
$\boldsymbol{\Gamma}_{\tau_T}$, making it invariant to monotone rescaling and hence comparable
across datasets whose absolute dissimilarity scales differ by orders of magnitude.

\paragraph{Algorithm margins.} The per-algorithm margins (Tables~\ref{app:margin-hdbscan} - \ref{app:margin-agglomerative})
replace the true assignment $z$ by the partition returned by the respective clustering algorithm at
$\tau_T$, keeping the matrix fixed; when a clustering algorithm returns a single cluster, $B$ is undefined over
an empty $\mathcal{B}$ and no margin is reported (``--'').

\clearpage
\FloatBarrier
\subsection{Demonstration of Individual Saliency Maps}
\label{app:saliency_maps}

\begin{figure}[h!]
    \centering
    \includegraphics[width=0.5\linewidth]{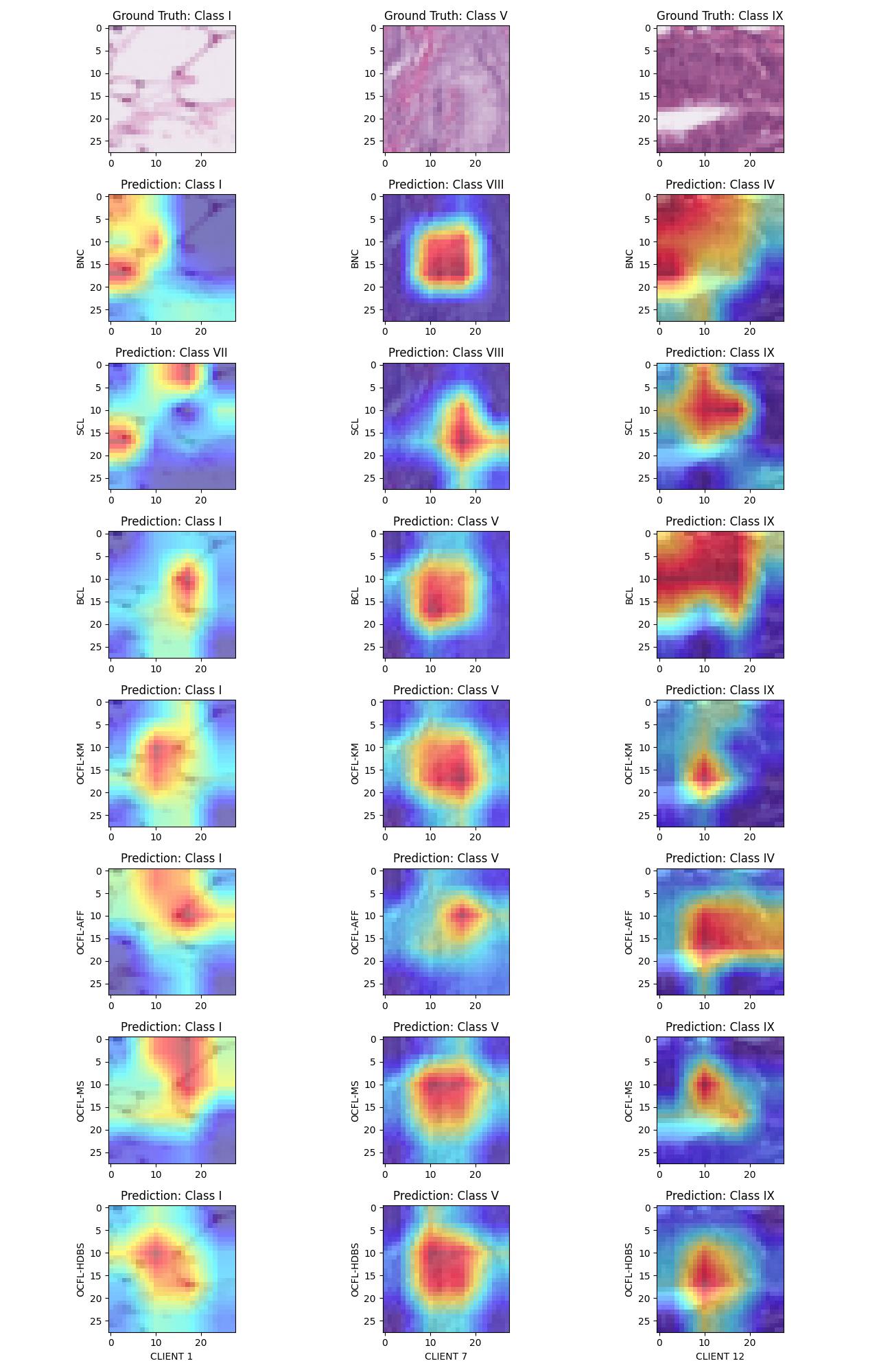}
    \Description{Grid of saliency maps on PathMNIST; columns are clients, rows are clustering methods, with true and predicted classes above each image.}
    \caption{Showcase of various saliency maps generated using seven different local models trained by various clustering algorithms on a PathMNIST dataset: Baseline (BNC), \cite{b7} (SCL), \cite{b1} (BCL), OCFL with K-Means (OCFL-KM), OCFL with Affinity Propagation (OCFL-AFF), OCFL with Mean-Shift (OCFL-MS) and OCFL with HDBSCAN (OCFL-HDBS). Each column represents a different client, while each row represents a different clustering algorithm. Additionally, the true and predicted classes are placed above the images.}
\end{figure}

\begin{figure}[h!]
    \centering
    \includegraphics[width=0.5\linewidth]{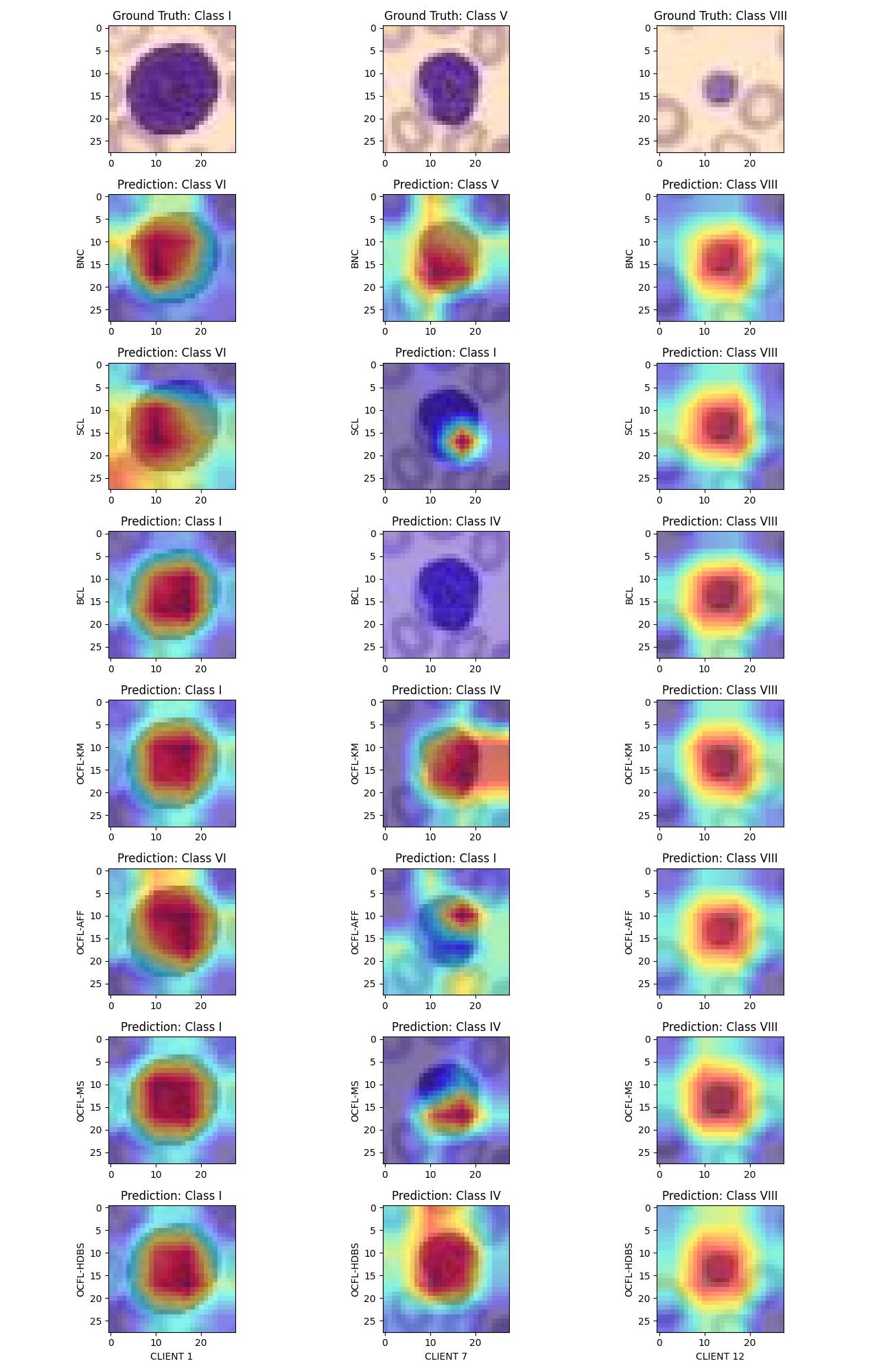}
    \Description{Grid of saliency maps on BloodMNIST; columns are clients, rows are clustering methods, with true and predicted classes above each image.}
    \caption{Showcase of various saliency maps generated using seven different local models trained by various clustering algorithms on a BloodMNIST dataset: Baseline (BNC), \cite{b7} (SCL), \cite{b1} (BCL), OCFL with K-Means (OCFL-KM), OCFL with Affinity Propagation (OCFL-AFF), OCFL with Mean-Shift (OCFL-MS) and OCFL with HDBSCAN (OCFL-HDBS). Each column represents a different client, while each row represents a different clustering algorithm. Additionally, the true and predicted classes are placed above the images.}
\end{figure}
\clearpage
\FloatBarrier

\clearpage
\FloatBarrier
\subsection{CD Plots for Insertion-Deletion}
\label{app:cd_plots}

\begin{figure}[h!]
\centering
\begin{subfigure}[b]{1\textwidth}
    \centering
    \includegraphics[width=0.7\linewidth]{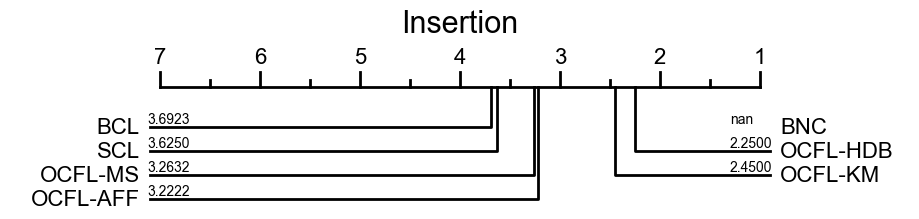}
    \Description{Critical difference diagram ranking the methods by insertion AUC on out-of-cluster samples.}
\end{subfigure}
\begin{subfigure}[b]{1\textwidth}
    \centering
    \includegraphics[width=0.7\linewidth]{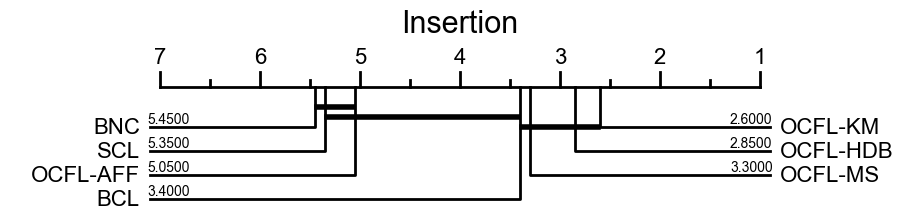}
    \Description{Critical difference diagram ranking the methods by insertion AUC on orchestrator samples.}
\end{subfigure}
\begin{subfigure}[b]{1\textwidth}
    \centering
    \includegraphics[width=0.7\linewidth]{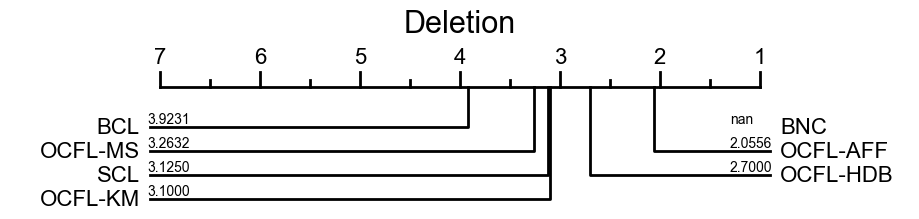}
    \Description{Critical difference diagram ranking the methods by deletion AUC on out-of-cluster samples.}
\end{subfigure}
\begin{subfigure}[b]{1\textwidth}
    \centering
    \includegraphics[width=0.7\linewidth]{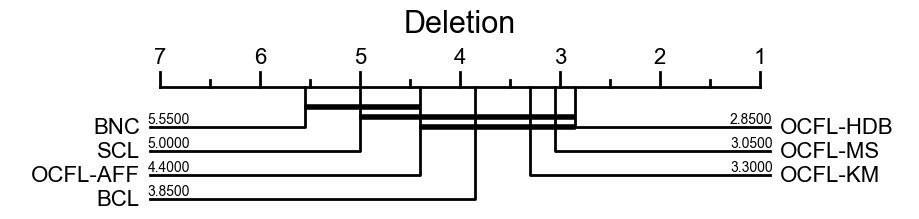}
    \Description{Critical difference diagram ranking the methods by deletion AUC on orchestrator samples.}
\end{subfigure}
    \caption{Critical Difference Plots for insertion and deletion. Particular algorithms are ranked and plotted on the x-axis according to their mean rank in insertion AUC (first two panels) and deletion AUC (last two panels); rank 1 is best. The panels show, from top to bottom, insertion on out-of-cluster samples, insertion on orchestrator samples, deletion on out-of-cluster samples and deletion on orchestrator samples. The bold black bar indicates that no statistical differences were detected. The plots present results obtained on the out-of-cluster and orchestrator samples for the cluster hypothesis function as explained in Algorithm \ref{algo:INDE}.}
\end{figure}

\FloatBarrier
\section{Online Resources}
As part of this paper, we share an extensive online resource that may help researchers assess the presented One-Shot Clustered Federated Learning (OCFL) algorithm or advance research on the topic of Clustered Federated Learning (CFL) itself.

There are two separate GitHub repositories associated with this paper. The first one contains results obtained during original runs, and the code for generating all the figures. The second one contains code for recreating all the experiments as well as source code for experimenting with modified versions of the presented algorithms. We justify the decision to split the code into two distinct repositories by the fact that the second one contains adjustable code for designing custom experiments. It may be the case that some readers are interested in only one of those repositories; in such a case, it is a better choice to keep them separate.

The repository for all the numerical results 
is fully available on GitHub.\footnote{The repository can be accessed under the following link: \url{https://github.com/MKZuziak/OCFL_Archive}} It contains all the numerical results obtained during all the different $280$ simulation runs, including a complete set of metrics registered on each client. It also contains a source code for analysing the results and generating graphs and tables presented in this paper. The second repository\footnote{The repository can be accessed under the following link: \url{https://github.com/MKZuziak/OCFLSuite}} contains the scripts necessary to reproduce all the simulations. It is also structured in such a way that it is entirely possible to either adjust the hyperparameters of the run (including the clustering algorithm) or the source code itself, making it a convenient starting point for further experimentation with CFL.